\pdfoutput=1
\documentclass[11pt]{article}
\usepackage[margin=1in]{geometry}
\usepackage[T1]{fontenc}
\usepackage{lmodern}
\usepackage{microtype}
\usepackage{graphicx}
\usepackage{booktabs}
\usepackage{array}
\usepackage{adjustbox}
\usepackage{float}
\usepackage{longtable}
\usepackage{amsmath,amssymb}
\usepackage{tikz}
\usetikzlibrary{arrows.meta,positioning,fit,decorations.pathreplacing}
\usepackage[round]{natbib}
\setcitestyle{authoryear,round,citesep={;},aysep={,},yysep={;}}
\renewcommand{\cite}[1]{\citep{#1}}
\usepackage{authblk}
\usepackage{url}
\usepackage{hyperref}
\definecolor{linkblue}{rgb}{0,0.08,0.45}
\hypersetup{colorlinks=true,linkcolor=linkblue,citecolor=linkblue,urlcolor=linkblue,hypertexnames=false,
  pdftitle={Late but Valuable: Why Deadlines Change the Measured Value of LLM Reasoning},
  pdfauthor={Sina Barazandeh, Benjamin E. Lundell, Jose Lugo-Martinez},
  pdfkeywords={agentic systems, LLM serving, inference effort, deadlines, fallback, laboratory automation}}
\newcommand{\hd}[1]{\begin{tabular}[b]{@{}c@{}}#1\end{tabular}}
\newcommand{\doi}[1]{doi: \href{https://doi.org/#1}{\nolinkurl{#1}}}
\extrafloats{100}

\makeatletter
\g@addto@macro\UrlBreaks{\do\a\do\b\do\c\do\d\do\e\do\f\do\g\do\h\do\i\do\j\do\k\do\l\do\m\do\n\do\o\do\p\do\q\do\r\do\s\do\t\do\u\do\v\do\w\do\x\do\y\do\z\do\A\do\B\do\C\do\D\do\E\do\F\do\G\do\H\do\I\do\J\do\K\do\L\do\M\do\N\do\O\do\P\do\Q\do\R\do\S\do\T\do\U\do\V\do\W\do\X\do\Y\do\Z\do\0\do\1\do\2\do\3\do\4\do\5\do\6\do\7\do\8\do\9}
\makeatother

\title{Late but Valuable: Why Deadlines Change the\\ Measured Value of LLM Reasoning}
\author[1]{Sina Barazandeh}
\author[2]{Benjamin E.\ Lundell}
\author[1]{Jose Lugo-Martinez\thanks{Corresponding author: \texttt{jlugomar@andrew.cmu.edu}}}
\affil[1]{Ray and Stephanie Lane Computational Biology Department, Kenneth C.\ Griffin School of Computer Science, Carnegie Mellon University, Pittsburgh, PA, USA}
\affil[2]{Arm, Inc.}
\date{}

\begin{document}
\maketitle

\begin{abstract}
Large language models (LLMs) increasingly plan and run experiments in automated laboratories. These experiments keep running while the model reasons and run on schedules with hard deadlines. If an LLM-generated answer arrives late, the system executes a safe fallback action instead, such as rejecting a sample. This creates a systems problem that standard serving metrics do not capture. An answer that matches the fallback adds no value, and answers that differ from the fallback may require more reasoning time. Reporting answer quality and latency separately can therefore overstate the value of reasoning. We test this on synthetic sample decisions, where the best choice is known exactly and a runtime enforces the deadline.
We compare one LLM's fast mode, which answers directly, with its thinking mode, which reasons first. In a prespecified replay of 1{,}200 new decisions, thinking answers that missed a deadline were worth about five times as much as on-time ones.
Multiplying the on-time rate by the mean answer value predicted about twice the value that thinking delivered.
This prediction overstated thinking's advantage over fast mode by 0.0241 per decision, and by 0.0191 in a repeat on 300 new decisions. Both 95\% confidence intervals excluded zero.
In a single-GPU Qwen3-8B experiment using simultaneous request bursts, raising the number of requests from 8 to 56 reduced the share of thinking answers that met a 45~s deadline from 79.2\% to 1.4\%, although total output tripled.
Evaluations of LLM agents for laboratories should therefore measure each answer's value and arrival time together and score the action that actually runs.
\end{abstract}

\raggedbottom

\section{Introduction}
\label{sec:intro}

Automated laboratories have become a research priority, with public investment in robotic and AI-programmable cloud laboratories that users will access remotely~\cite{aiactionplan2025,genesis2025,nsf2026cloudlabs}. An OECD report argues that faster research could be the most economically and socially valuable use of AI~\cite{oecd2023aiscience}. Such laboratories increasingly use large language models (LLMs) to plan and run experiments~\cite{unilabos,panapitiya2026autolabs}. These experiments run on schedules, and a correct answer adds no value if it arrives after its step is due. A laboratory process continues while the model reasons. Its steps have hard timing limits, from a reaction that must be stopped (quenched) after 35~s~\cite{bateni2024smartdope} to strict intervals between operations involving live cells~\cite{itoh2021scheduling}. When a step has a hard deadline, a late answer is unusable and the system runs a preset action. An agent supervising optical-fiber production must decide after each quality measurement whether to accept the sample, reject it or keep tuning. The next process step is due at a fixed time. If the answer arrives late, the system executes a preset \emph{fallback} action, which here rejects the sample. Robotic platforms also fail, so laboratory software already needs such preset actions. In one autonomous materials laboratory, up to 22\% of tasks of one type raised an exception~\cite{alabos}. A mobile robotic chemist discarded 11 of 688 results because of workflow errors or a faulty vial seal~\cite{burger2020mobile}.

Conservative fallback actions raise a systems problem that standard serving metrics do not capture. Traditional serving benchmarks and schedulers aim to complete a specified fraction of requests within a latency target~\cite{reddi2020mlperf,zhong2024distserve}, and slower answers are still delivered. In a laboratory, the fallback replaces a late answer, so an answer that agrees with it adds nothing, however fast it arrives. More reasoning can improve an answer but takes time~\cite{snell2025scaling,ma2025nothinking}. Laboratory time limits, from seconds to minutes, can overlap with model reasoning times, so the amount of reasoning can decide which action runs. Our central question is how to measure the value of extra reasoning under hard deadlines.

A natural approach measures two numbers for each mode, the share of valid, on-time answers and the mean gain of a valid answer over the fallback. We call these \emph{separate summaries}. Their product predicts the gain the mode delivers. When some valid answers arrive on time and others late, the prediction is exact only if the two groups have the same mean gain.
Suppose all answers are valid, with a mean gain of 0.2 over the fallback, and half arrive in time. The prediction is then 0.1. On-time answers might gain 0.05 and late ones 0.35 if the model is slowest on the cases where its answer helps most. The deadline then discards the more valuable half, so the delivered gain over fallback is 0.025, a quarter of the prediction.
The error arises because the product of two averages need not equal the average of their product. Let $\pi$ be the share of valid answers. Among valid answers, let $T$ be 1 for an answer that arrives in time and 0 otherwise, and let $v$ be its gain over the fallback. The on-time rate is $\pi\,\mathbb{E}[T]$ and the delivered gain is $\pi\,\mathbb{E}[Tv]$, so the overprediction is
\begin{align}
B&=\pi\,\mathbb{E}[T]\,\mathbb{E}[v]-\pi\,\mathbb{E}[Tv]=-\pi\,\mathrm{Cov}(T,v)\nonumber\\
&=\pi\,\bar t\,(1-\bar t)\,(\bar v_{\mathrm{late}}-\bar v_{\mathrm{on}}).\label{eq:bias}
\end{align}
Here $\bar t=\mathbb{E}[T]$ is the on-time share of valid answers, and $\bar v_{\mathrm{on}}$ and $\bar v_{\mathrm{late}}$ are the mean gains of on-time and late answers. The last step uses that $T$ is 0 or 1 and that both groups exist. In the example, $0.5\times0.5\times0.3=0.075$, the gap between 0.1 and 0.025. The error is zero when all valid answers arrive in time or none do, and it grows with the gap in value between late and on-time answers. Separate summaries measure $\bar t$ and the mean of $v$ but not how they vary together, so they cannot detect the error.

We test this on synthetic decisions about laboratory samples. The processes, priors and payoffs are fully specified, so the best decision and the value of every action can be computed exactly. We use these tasks to study how answer value relates to latency, not to argue that an LLM should replace an exact calculation. A runtime sends each decision to a model, records when the answer arrives, enforces the deadline and executes the model's chosen action or the fallback in a process simulator. Our main studies use Claude Haiku 5.5. Thinking improved its decisions when time was ignored (\S\ref{sec:effort}). In separate accept-or-reject studies, short deadlines nearly eliminated thinking's advantage over fast mode (\S\ref{sec:slack}). A replay of recorded answers then measured how much separate summaries misjudge the value that thinking delivers (\S\ref{sec:dependence}). We also compare delivered value across four hosted models (\S\ref{sec:cost}). For Claude Sonnet 5.5, whose fast mode already made good decisions, we detected neither the quality gain nor the deadline effect.

The replay, our central result, shows why separate summaries overstate thinking's value. At a 2.1~s deadline, Haiku's late thinking answers gained about 0.25 on average over rejecting, against about 0.05 for on-time ones.  Answers with more than 400 thinking tokens gained 0.53 on average, about ten times the gain of thinking answers with no thinking tokens. None of them arrived within 2.1~s. Separate summaries predicted that thinking would do about as well as fast mode, but thinking delivered less than half of fast mode's value. The overestimate of thinking's advantage over fast mode was 0.0241 per decision (95\% confidence interval (CI) $[0.0165,0.0317]$), and it persisted within groups matched by task, number of measurements and collection batch (Figure~\ref{fig:main}c). A repeat on 300 new instances reproduced the overestimate but not the ranking of the two modes.

In the replay, the 2.1~s deadline discarded thinking's more valuable answers. Serving load can make this worse, because a busy server leaves each answer less time before its deadline. 
Total throughput can hide this effect, because it counts tokens produced, not when each answer arrives or what it is worth. We served Qwen3-8B on one GPU and raised the number of simultaneous requests from 8 to 56. The server produced three times as many tokens per second, but the mean generation speed of each thinking answer roughly halved.  Lower speeds and later starts leave less time for long answers. With the observed mean timings, a 45~s deadline left time for roughly 600 tokens, down from 1{,}500, and the share of thinking answers that met it fell from 79.2\% to 1.4\% (\S\ref{sec:serving}).

We contribute an evaluation that pairs each answer's value with its arrival time and scores the action that executes, supported by controlled model studies and a shared GPU experiment. Our results show that maximizing inference throughput or answer quality independently does not necessarily maximize useful decisions executed before their deadlines. 

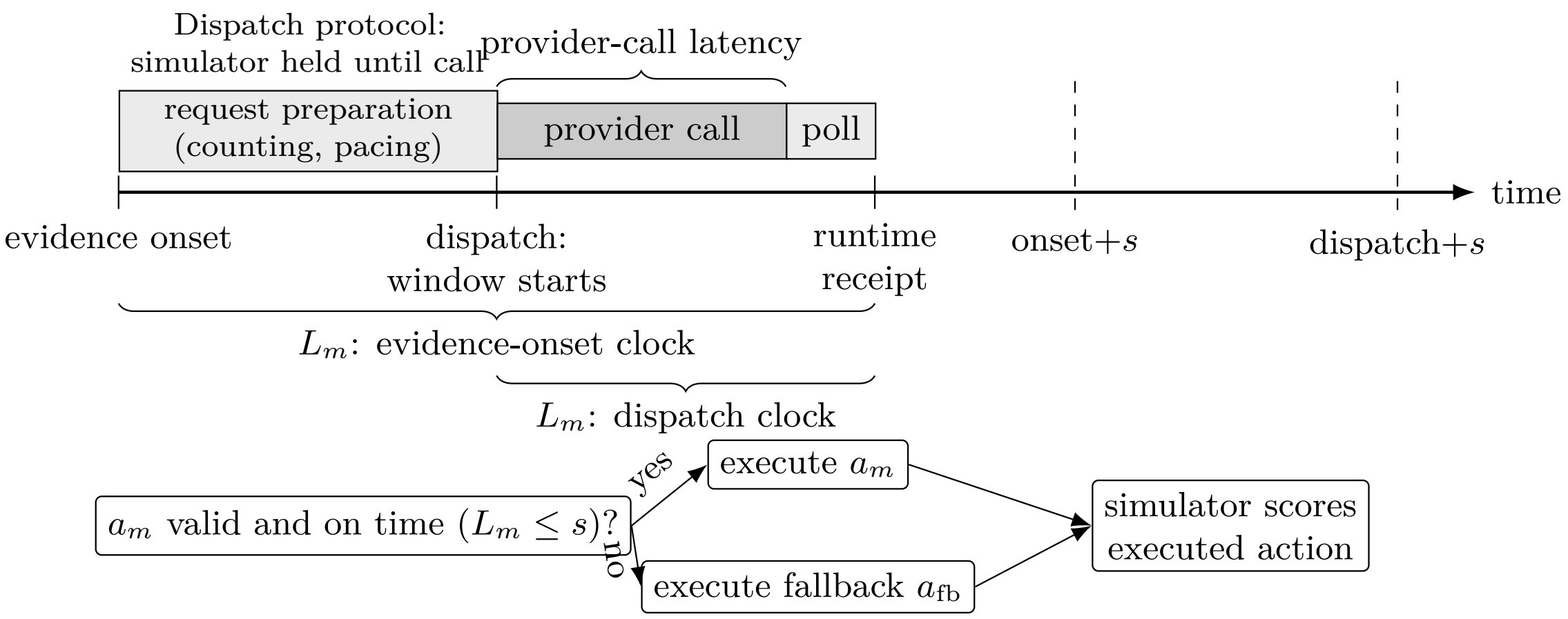
\begin{figure*}[t]
\centering
\resizebox{0.92\textwidth}{!}{%
\begin{tikzpicture}[font=\footnotesize,
  arr/.style={-{Latex[length=1.8mm]}, thin},
  seg/.style={draw, minimum height=5mm, inner sep=2pt, align=center}]
\draw[thick,-{Latex[length=2mm]}] (0,0) -- (12.2,0) node[right]{time};
\node[seg, fill=black!8, minimum width=3.4cm, text width=3.1cm, font=\scriptsize, anchor=west]
 at (0,0.55) {request preparation (counting, pacing)};
\node[seg, fill=black!20, minimum width=2.6cm, anchor=west] at (3.4,0.55) {provider call};
\node[seg, fill=black!8, minimum width=0.8cm, anchor=west] at (6,0.55) {poll};
\node[align=center, text width=3.4cm, font=\scriptsize] at (1.7,1.35)
 {Dispatch protocol:\\simulator held until call};
\draw (0,0.15) -- (0,-0.15) node[below]{evidence onset};
\draw (3.4,0.15) -- (3.4,-0.15) node[below,align=center]{dispatch:\\window starts};
\draw (6.8,0.15) -- (6.8,-0.15) node[below,align=center]{runtime\\receipt};
\draw[dashed] (8.6,1) -- (8.6,-0.2) node[below]{onset$+s$};
\draw[dashed] (11.5,1) -- (11.5,-0.2) node[below]{dispatch$+s$};
\draw[decorate,decoration={brace,amplitude=4pt,mirror}] (0,-1) -- (6.8,-1)
 node[midway,below=3pt]{$L_m$: evidence-onset clock};
\draw[decorate,decoration={brace,amplitude=4pt}] (3.4,0.95) -- (6,0.95)
 node[midway,above=3pt]{provider-call latency};
\draw[decorate,decoration={brace,amplitude=4pt,mirror}] (3.4,-1.65) -- (6.8,-1.65)
 node[midway,below=3pt]{$L_m$: dispatch clock};
\node[draw, rounded corners=1.5pt, align=center, inner sep=3pt] (q) at (2.2,-3) {$a_m$ valid and on time ($L_m\le s$)?};
\node[draw, rounded corners=1.5pt, align=center, inner sep=3pt] (ex) at (6.2,-2.45) {execute $a_m$};
\node[draw, rounded corners=1.5pt, align=center, inner sep=3pt] (fb) at (6.2,-3.55) {execute fallback $a_{\text{fb}}$};
\node[draw, rounded corners=1.5pt, align=center, inner sep=3pt] (sim) at (10,-3) {simulator scores\\executed action};
\draw[arr] (q.east) -- node[above,sloped]{yes} (ex.west);
\draw[arr] (q.east) -- node[below,sloped]{no} (fb.west);
\draw[arr] (ex.east) -- (sim.west);
\draw[arr] (fb.east) -- (sim.west);
\end{tikzpicture}}
\caption{Two timing protocols and runtime execution. Evidence-onset timing includes request preparation
(counting, pacing and admission), the provider call and polling. Dispatch timing and the simulated process window start together at the provider call, and the simulator is paused until then. Each protocol allows deadline $s$ plus 10~s for removal. Provider-call latency includes service overhead. $a_m$ is mode $m$'s answer, $a_{\text{fb}}$ the fallback and $L_m$ the delay to receipt. The runtime executes valid, on-time answers or the fallback.}
\label{fig:harness}
\end{figure*}

\section{Simulated laboratory tasks and measurement}
\label{sec:setting}
\label{sec:background}

In rational metareasoning, an agent keeps thinking while expected decision improvement outweighs the time cost~\cite{russell1991metareasoning,lin2015metareasoning}. Here, missing the deadline incurs the cost of executing the fallback.

\paragraph{Sample-stopping decisions.} Each instance asks whether to stop work on a sample. Instances come from two \emph{task families}, synthetic optical-fiber drawing and quantum-dot synthesis (quantum dots are semiconductor nanoparticles). Both use a fully specified sample-quality model with finitely many states, so final decisions can be scored exactly. The families differ in measurement noise and action durations. The model receives the measurement history, the available actions and their durations, and a cutoff after which no new work may start. Every decision ends with \emph{removal}, 10~s of work to take the sample out of the instrument. Histories of 1, 3 or 5 measurements define the first, middle and late \emph{observation stages}, respectively. The model chooses whether to accept
the sample, reject it or continue by re-measuring or tuning. In the decision-quality study, a \emph{terminal} observation leaves too little time for further measurement or tuning after allowing for the decision and removal. Continuing receives the rejection value minus the cost of work started (at most 0.0031). In the deadline and replay studies, only accept and reject are available, so every decision is final regardless of the remaining time.
Our deadlines range from 2.1 to 60~s, comparable to reported laboratory time limits such as the 35~s reaction stop above. These tasks are simpler than open-ended laboratory planning, and we do not estimate deployment gains from them.

\paragraph{The sample model.} The hidden sample model has 42 equally likely states, one for each combination of optimum $z\in\{-1,-0.9,\ldots,1\}$ and sharpness $r\in\{1.5,3\}$. At process setting $x$, quality is $h(x;z,r)=0.98\exp[-r(x-z)^2]$, and the sample is good when $h\ge0.8$. Each measurement reports the quality and the distance $z-x$ to the optimum, with independent zero-mean Gaussian noise. Noise standard deviations for quality and distance are 0.05 and 0.08 for optical-fiber measurements, 0.10 and 0.25 for ordinary quantum-dot measurements and 0.02 and 0.04 for high-resolution quantum-dot measurements. Normalizing the likelihoods over the 42 states gives the exact posterior probability
$p$ that the sample is good. Accepting is worth $Q_{\text{acc}}=2p-1-10^{-4}$ and rejecting $Q_{\text{rej}}=-10^{-4}$,
where $10^{-4}$ is the charge for 10~s of removal work.  Prompts give the task, prior, costs, cutoff and history, but no reference answer.
Appendix~A.1 gives the full specification and example prompts.

\paragraph{Exact values and regret.} For these terminal decisions, the \emph{exact Bayes controller} chooses the action with the highest posterior expected value $Q(a)$. It needs no LLM, is optimal by construction and serves only as the scoring reference. \emph{Decision regret} $\mathbb{E}[\max_a Q(a)-Q(a_m)]$ measures expected loss from mode $m$'s answer $a_m$ relative to the best action, regardless of arrival time. Here and below, $\mathbb{E}$ averages over a study's instances.

\paragraph{An example decision.} One optical-fiber observation (Appendix~A.1) has an exact posterior probability of about 0.9998 for a good sample, so accepting is optimal. When offered accept or reject, fast mode rejected the sample and low-effort thinking accepted it. With 2.1~s allowed from request dispatch, the thinking answer arrived too late and rejection executed. With 60~s, a separately generated acceptance arrived in time. Because the sample was good, accepting it gave a realized utility of $1-10^{-4}=0.9999$. This example shows how a deadline can prevent a better answer from being executed.

\paragraph{Deadlines, fallback and clocks.} The deadline $s$ is the time allowed to receive an answer, which we also call slack. The measured delay $L_m$ runs from the start of timing to runtime receipt, including polling. We start timing in two ways (Figure~\ref{fig:harness}). The \emph{dispatch clock} starts when the request is sent,
with the simulated process paused until then. The \emph{evidence-onset clock} starts when the observation becomes available and includes request preparation and pacing. An answer is \emph{valid} if it is well-formed and names an action available at that observation. Failed, refused, truncated or malformed responses are invalid. Replacing late or invalid answers with the fallback gives the executed action and its expected utility
\begin{gather*}
A_m(s)=\begin{cases} a_m & \text{if } a_m \text{ is valid and } L_m\leq s,\\ a_{\text{fb}} & \text{otherwise,}\end{cases}\\
J_m(s)=\mathbb{E}\big[U_{\text{exec}}(A_m(s))\big],
\end{gather*}
where the fallback $a_{\text{fb}}$ is reject. In the deadline experiments, $U_{\text{exec}}$ measures \emph{realized utility} from the process simulator. Accepting a good sample earns $+1$, a false accept $-1$ and rejection $0$, each minus $10^{-5}$ per second of removal work. The replay scores the executed action by its expected value given the evidence (\emph{posterior expected utility}), without simulating an outcome. Both measures use the same payoff scale. Because realized utility depends on each sample's simulated outcome, values from the deadline studies and the replay should not be compared directly. The advantage of extra effort in executed utility is $\Delta J(s)=J_{\text{think}}(s)-J_{\text{fast}}(s)$, and economic value also depends on inference cost (\S\ref{sec:cost}).

\paragraph{Separate and joint estimates.} For the replay we score each valid answer by its gain over the fallback,
$v=Q(a)-Q(a_{\mathrm{fb}})$, with $v=0$ for a failure. Let $q_m(s)$ be the probability that mode $m$'s answer is valid and
received by the deadline. The \emph{separate-summary} prediction multiplies the on-time rate by the mean value of all valid answers, and the \emph{joint} estimate pairs each answer's value with its arrival time,\begin{align*} G_{\mathrm{sep},m}(s)&=q_m(s)\,\mathbb{E}[v\mid\mathrm{valid}],\\ G_{\mathrm{joint},m}(s)&=\mathbb{E}[v\,\mathbb{I}\{\mathrm{valid},L_m\le s\}].
\end{align*}
Here $\mathbb{I}$ indicates a valid answer received by the deadline. The joint estimate is therefore the posterior expected gain delivered after deadline enforcement. Their gap $B_m(s)=G_{\mathrm{sep},m}(s)-G_{\mathrm{joint},m}(s)$
is the overprediction of separate summaries. By Equation~(\ref{eq:bias}), $B_m(s)=-\pi\,\mathrm{Cov}(T,v)$, with the covariance taken over mode $m$'s valid answers at deadline $s$. Separate summaries overpredict when late answers are worth more than on-time ones and underpredict when they are worth less. 

\begin{table*}[t]
\centering
\caption{The core studies use distinct new instances and protocols. Regret scores the returned answer and
ignores timing. Realized and posterior expected utility score the executed action, including fallback. Details are in Appendices~A.2 and C.1.}
\label{tab:studies}
\footnotesize
\setlength{\tabcolsep}{4pt}
\renewcommand{\arraystretch}{1.0}
\begin{tabular}{@{}>{\raggedright\arraybackslash}p{0.15\textwidth}>{\raggedright\arraybackslash}p{0.2\textwidth}>{\raggedright\arraybackslash}p{0.3\textwidth}>{\raggedright\arraybackslash}p{0.27\textwidth}@{}}
\toprule
study & instances and actions & what is measured & evidence status \\
\midrule
decision quality (\S\ref{sec:effort}) & 1{,}200 terminal instances; continuation offered & terminal decision regret by effort level & prespecified tests; regret breakdown descriptive \\
\addlinespace[4pt]
deadline execution (\S\ref{sec:slack}) & 600 instances per clock; accept or reject only; a new response at each deadline & how thinking's advantage in realized utility changes between 2.1 and 60~s & one prespecified test per clock; other deadlines descriptive \\
\addlinespace[4pt]
prospective replay (\S\ref{sec:dependence}) & 1{,}200 instances; accept or reject only; one 60~s-prompt answer per mode & how much separate summaries overstate thinking's delivered posterior expected advantage & one prespecified test; pooled ranking secondary \\
\addlinespace[4pt]
Sonnet check (\S\ref{sec:effort}, \S\ref{sec:slack}) & 1{,}200 and 600 instances & regret contrast and dispatch-clock interaction & two prespecified tests \\
\bottomrule
\end{tabular}
\end{table*}

\section{Experimental design}
\label{sec:methods}

We ran three hosted studies of Claude Haiku 5.5 and a check on Claude Sonnet 5.5, which tests whether the results carry over to a second model (Table~\ref{tab:studies}). Section~\ref{sec:serving} adds a study of serving load on one local GPU, and Section~\ref{sec:cost} compares four hosted models. The Haiku and Sonnet studies used new instances generated without selection. We fixed each study's inputs, schedules, failure handling and analysis before collecting scored data.

\paragraph{Inference modes.} For Haiku~5.5, \emph{fast mode} disables thinking and uses the low effort setting. \emph{Thinking mode} enables the provider's adaptive thinking at low, medium or high effort, so the model chooses how much to reason. Unless stated otherwise, ``thinking'' means low effort. These are request settings, so a thinking-mode response may still use few or no thinking tokens. Sonnet~5.5 does not allow thinking to be disabled. Its fast mode uses low effort with the provider's most restrictive setting, which permits no thinking before the answer. The hosted studies use a 16{,}000-token cap and a JavaScript Object Notation (JSON) output schema. Failed, refused, truncated or malformed responses count as the fallback. Costs come from recorded token usage at list prices.

\paragraph{Protocols.} The decision-quality study offers continuation, so the model must infer from the durations and cutoff that no further work fits. The deadline and replay studies offer only accept and reject. In the deadline studies, every combination of instance, deadline and mode gets its own call, with the cutoff in the prompt. The decision-quality and deadline studies sent requests one at a time, without sampling parameters or automatic retries, and balanced instances over task family and observation stage. 

\paragraph{Uncertainty.} We call a result \emph{confirmatory} only when it comes from a test fixed in a study plan. We treat all other results, including those in Section~\ref{sec:cost}, as descriptive. Within each plan, we apply a Bonferroni correction to its tests. The largest plan has four tests, so its intervals are 98.75\%. They cover terminal decision regret, the two deadline tests and a fault-diagnosis test on 128 fixed cases (Appendix~B.1). The two higher-effort comparisons form one family of two tests and the Sonnet check another, and the replay and the load study each have a single test. We use percentile bootstrap intervals over instances. The load study instead uses a paired $t$ interval over 12 GPU allocations, its independent units. Appendices~A.3 and A.6 give the procedures.

\paragraph{Study sequence and checks.} Earlier findings shaped the questions and settings of later studies. The replay was planned after an exploratory look at dispatch-clock records. Our error control applies within each study, not to these choices across studies (Appendix~C.1). Separate code reproduced the estimates from raw records and reran recorded actions in the simulator. These checks confirm the computations, not the physical realism of the simulator (Appendix~C.4).

\begin{figure*}[t]
\centering
\includegraphics[width=\textwidth]{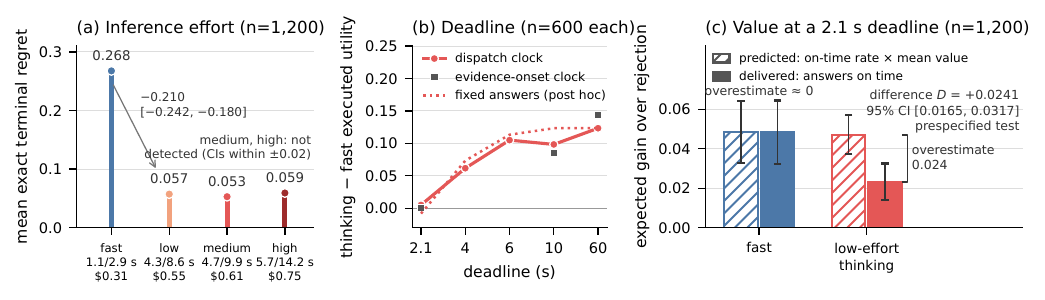}
\caption{Inference effort and executed outcomes. (a) Exact terminal decision regret on 1{,}200 synthetic stopping instances for Claude Haiku 5.5. Labels give median/99th-percentile provider-call latency and cost per thousand decisions.
(b) Realized utility, low-effort thinking minus fast mode, timed from provider dispatch (line) or from evidence onset (gray squares). The dotted curve is a post hoc, descriptive sweep of fixed 60~s-prompt answers, and deadlines are equally spaced categories.
(c) Posterior expected gain over rejection at a 2.1~s deadline on 1{,}200 new instances answered under 60~s prompts. Gains are averaged within groups of comparable instances and are either predicted from separate summaries (on-time rate times mean answer value) or delivered by on-time answers. Bars and 95\% CIs are descriptive. Only the difference in overprediction, $D=+0.0241$ (thinking minus fast mode), is confirmatory. In (a), the low-minus-fast interval is 98.75\%, and the medium-minus-low and high-minus-medium intervals are 97.5\%. In (b), points are descriptive, and only the advantage at 60~s minus that at 2.1~s is tested.}
\label{fig:main}
\end{figure*}
\section{Decision quality without deadlines}
\label{sec:effort}

The first study asks whether thinking returns better decisions than fast mode when time is ignored.

\begin{table}[t]
\centering
\caption{Prespecified Haiku~5.5 comparisons of terminal decision regret on the same 1,200 new instances (first mode
minus second; lower is better). Fast mode disables thinking, and low, medium and high are thinking effort levels. Bonferroni-adjusted $p$ values are compared with .05. Percentile bootstrap CIs are 98.75\% for the first row (four-test family) and 97.5\% for the others (two-test family). $^\dagger$Medium-effort responses were collected the same day, about 13~h after low-effort responses.}
\label{tab:confirm}
\small
\setlength{\tabcolsep}{2pt}
\begin{tabular}{@{}lrrrl@{}}
\toprule
contrast & effect & CI & adj.\ $p$ & result \\
\midrule
low $-$ fast & $-0.210$ & $[-0.242,-0.180]$ & .002 & detected \\
medium $-$ low$^\dagger$ & $-0.004$ & $[-0.017, 0.008]$ & .88 & not detected \\
high $-$ medium & $+0.006$ & $[-0.007, 0.020]$ & .60 & not detected \\
\bottomrule
\end{tabular}
\end{table}

\paragraph{Thinking improved Haiku's decisions.} On 1{,}200 new terminal instances, every response in both modes was valid. Thinking reduced Haiku's decision regret from 0.268 in fast mode to 0.057, about 79\% lower. The prespecified difference was $-0.2104$ (98.75\% CI $[-0.2417,-0.1798]$; Table~\ref{tab:confirm}, Figure~\ref{fig:main}a). Mean cost per decision was about \$0.00031 in fast mode and \$0.00055 with thinking. An earlier, smaller study detected no improvement over fast mode (Appendix~A.4).

\paragraph{Much of the gain came from avoiding unnecessary continuation.} Most of fast mode's regret came from asking to continue when accepting was optimal (0.189 of 0.268). Thinking cut this part to 0.043 (post hoc; Appendix~A.4). The prompt did not label the decision as final, so the model had to infer from the durations and cutoff that no further work fit. Part of thinking's gain therefore comes from recognizing that no time was left.

\paragraph{Higher effort added latency without a detected gain.} Medium and high effort did not reduce regret further (Table~\ref{tab:confirm}). The intervals extend at most about 0.02 from zero, under a tenth of the gain from thinking, although no equivalence threshold was set in advance. Medium-effort responses were collected about 13~h after low-effort ones, so that comparison cannot separate effort from changes at the provider. High effort raised 99th-percentile latency from 8.6 to 14.2~s and cost per decision from \$0.00055 to \$0.00075.

\paragraph{Sonnet.} On 1{,}200 new instances, Sonnet~5.5's regret was 0.062 in fast mode and 0.047 with thinking. The prespecified difference was not detected ($-0.0143$, 97.5\% CI $[-0.0301,0.0012]$; raw $p=.041$ against $\alpha=.025$). Sonnet's fast mode already had regret close to Haiku's with thinking (on different instances). This may leave less room for improvement, although we did not test that explanation. The interval includes effects of a few hundredths, so the two modes are not shown to be equivalent.

\section{Deadline effects on executed value}
\label{sec:slack}

We next ask whether thinking still does better than fast mode once a deadline decides which action runs. These studies use new instances and allow only accept or reject, so the model cannot make the first study's continuation errors.

\paragraph{Timing from request dispatch.} Both modes answered 600 new instances at deadlines of 2.1, 4, 6, 10 and 60~s. The clock and the simulated process start together when the request is sent. Every combination of instance, deadline and mode received a new response, so changes in the answer cannot be separated from changes in timing. The 2.1~s deadline is close to thinking's median latency (2.06~s, against 0.88~s for fast mode), so about half of thinking answers met the deadline.

\paragraph{Thinking's advantage depended on the deadline.} The advantage over fast mode in realized utility was $+0.005$ at 2.1~s and $+0.123$ at 60~s (Figure~\ref{fig:main}b). At 2.1~s, 98.8\% of fast answers and 54.5\% of thinking answers arrived in time. The prespecified test compared thinking's advantage at 60~s with its advantage at 2.1~s. The difference was 0.118 (98.75\% CI $[0.065,0.172]$; Bonferroni-adjusted $p=.0004$). Results at 4, 6 and 10~s are descriptive (Appendix~A.5). A post hoc replay of fixed 60~s answers showed a similar pattern.

\paragraph{Timing from evidence onset.} In a laboratory, the deadline starts when the measurement becomes available, not when the request is sent. The evidence-onset clock therefore also counts request preparation and the client's pacing between calls, which had a median of 4.6~s. On 600 more new instances, thinking's advantage was zero at 2.1~s and 0.143 at 60~s (prespecified difference 0.143, 98.75\% CI $[0.103,0.183]$). At 2.1~s, preparation and pacing alone made at least 96.5\% of answers in both modes miss the deadline. With this pacing, request handling can matter as much as the model's own latency (Appendix~A.5).

\paragraph{Sonnet.} For Sonnet~5.5 on 600 new dispatch-clock instances, thinking's advantage grew by $+0.0150$ from 2.1 to 60~s. This prespecified difference was not detected (97.5\% CI $[-0.0133,0.0450]$; raw $p=.27$ against $\alpha=.025$).

\section{Prediction error from separate summaries}
\label{sec:dependence}

The deadline studies drew a new answer at each deadline, so they cannot tell whether a deadline discarded better answers or the model gave different answers. The replay holds the answers fixed, so it can compare late and on-time answers and measure the error of separate summaries.

\paragraph{Design.} An exploratory look at recorded 60~s answers from the dispatch-clock study showed overprediction, so we planned a confirmation in advance. It used 1{,}200 new instances collected in two batches of 600, with 100 instances per task family and observation stage in each batch. Each instance received one fast and one thinking answer to the same accept-or-reject prompt, which allowed 60~s. We requested the two modes back to back in random order. We replayed each answer's recorded arrival time against 2.1, 4, 6 and 10~s deadlines and replaced late answers with rejection. All 2{,}400 responses were valid, and no instances were added after outcomes were seen. A later study of Haiku thinking and Opus compared prompts that stated the actual deadline with the 60~s prompt. Estimated changes in delivered gain were at most 0.02, but the 95\% intervals reach 0.048, so larger effects remain possible (Appendix~B.5).

\paragraph{Estimates.} Each answer is scored by its posterior expected gain over rejecting, and \S\ref{sec:setting} defines the separate and joint estimates. We computed both within 12 groups that share task family, observation stage and batch, and then averaged them. The primary contrast is $D=B_{\mathrm{think}}(2.1)-B_{\mathrm{fast}}(2.1)$, thinking's overprediction minus fast mode's. A positive $D$ means separate summaries overstate thinking's advantage.

\begin{table}[t]
\centering
\caption{Replay at the 2.1~s deadline (1{,}200 new instances; posterior expected gain over rejecting). Here $q$ is the pooled share of answers that were valid and on time, and $\overline v$ is the mean gain for all valid answers and for on-time and late answers. $G_{\mathrm{sep}}$, $G_{\mathrm{joint}}$ and $B$ are averaged equally over the 12 groups. Entries are
descriptive. Only $D=B_{\mathrm{think}}-B_{\mathrm{fast}}=+0.0241$ (95\% CI $[0.0165,0.0317]$) is confirmatory.}
\label{tab:replay}
\small
\setlength{\tabcolsep}{3pt}
\begin{tabular}{@{}lrrrr@{}}
\toprule
\multicolumn{5}{@{}l}{Pooled answer diagnostics} \\
mode & $q$ & $\overline v$ & $\overline v$ on time & $\overline v$ late \\
\midrule
fast & 0.9625 & 0.0501 & 0.0504 & 0.0444 \\
low & 0.4525 & 0.1605 & 0.0509 & 0.2511 \\
\bottomrule
\end{tabular}
\par\smallskip
\begin{tabular}{@{}lrrr@{}}
\toprule
\multicolumn{4}{@{}l}{Estimates averaged equally over groups} \\
mode & $G_{\mathrm{sep}}$ & $G_{\mathrm{joint}}$ & $B$ \\
\midrule
fast & 0.0483 & 0.0485 & $-0.0001$ \\
low & 0.0470 & 0.0230 & $+0.0240$ \\
\bottomrule
\end{tabular}
\end{table}

\paragraph{Late thinking answers were worth more.} At 2.1~s, 45\% of thinking answers arrived in time. On-time thinking answers gained 0.0509 on average and late ones 0.2511 (Table~\ref{tab:replay}). By Equation~(\ref{eq:bias}), this gap is what makes separate summaries overpredict. Pooled over all instances, where every answer was valid, it gives $0.4525\times0.5475\times(0.2511-0.0509)=0.0496$, the pooled prediction 0.0726 minus the delivered 0.0230. Of the 543 on-time thinking answers, 96.1\% used no thinking tokens, while late answers used a median of 225.  Answers with more than 400 thinking tokens gained 0.526 on average, against 0.049 for thinking answers with no thinking tokens, and the repeat on new instances showed the same pattern (Appendix~A.7). 

\paragraph{Separate summaries overstated thinking's value.} Within groups, separate summaries predicted that thinking would deliver 0.0470 per decision, about the same as fast mode. Thinking delivered 0.0230 (Figure~\ref{fig:main}c). Thinking's overprediction exceeded fast mode's by $D=0.0241$ (95\% CI $[0.0165,0.0317]$), below the 0.03 effect used to plan the study. Grouping comparable instances reduced the overprediction but did not remove it. Without grouping, the separate-summary prediction for thinking was 0.0726. Overprediction was also positive at 4 and 6~s (Appendix~A.6). Because the link between answer value and arrival time is observational, we cannot separate the effect of thinking from that of task difficulty.

\paragraph{Separate summaries can change the ranking of the modes.} Without grouping, separate summaries ranked thinking above fast mode, and within groups they rated the two modes nearly equal. The joint estimate favored fast mode by 0.025 per decision in a descriptive comparison (95\% CI $[0.008,0.043]$).

\paragraph{Overprediction replicated, but the ranking did not.} After the main analysis, we repeated the replay on 300 new instances, 50 for each task family and observation stage. The prompts, modes and analysis were unchanged, and we fixed the design before collecting any data. Requests were still sent one at a time, with 3~s between them instead of 6~s. Thinking's overprediction again exceeded fast mode's, with $D=0.0191$ (95\% CI $[0.0039,0.0341]$). Late thinking answers were again worth more than on-time ones, 0.206 against 0.066. The ranking of the two modes did not replicate. Fast mode gained little over rejection on these instances, and the joint estimate favored thinking by 0.025, with an interval that includes zero (Appendix~A.7).

\paragraph{The fallback affected the direction of the error.} We also scored the same answers again as if a late answer led to accepting the sample. The prompt never stated what follows a late answer, so this choice could not affect the answers. With this unsafe fallback, separate summaries understated thinking's advantage at 4 and 6~s in both runs, and at 2.1~s the two runs disagreed. The unsafe fallback also made late answers far more costly. This check was not planned in advance (Appendix~A.7).

\section{Serving load and concurrent requests}
\label{sec:systems}

The replay shows that valuable answers can arrive too late. In a laboratory, many instruments and agents may share one inference server, so the delay of an answer also depends on the other requests.

\subsection{Serving load}
\label{sec:serving}

We varied the number of requests on one A6000 GPU serving Qwen3-8B with vLLM~\cite{vllm}. The thinking configuration also changes the sampling settings and token cap relative to fast mode (Appendix~B.2). We scored the executed actions on 36 fixed instances by posterior expected utility. Answers received 45~s or more after the burst started triggered the fallback.

A prespecified capacity scan found that up to 56 simulated laboratories running at once, without thinking, did not fully load the GPU, so we sent simultaneous bursts instead. Simultaneous requests may delay one another differently from requests that arrive during ordinary laboratory work. Each burst held six scored requests from one mode plus background thinking requests, for 8 requests in total at low load and 56 at high load. With 8 requests, 79.2\% of thinking answers were valid and on time (mean arrival 39.4~s). With 56, only 1.4\% were valid and on time (mean arrival 87.4~s). All fast answers were on time at both loads, arriving after 2.6 and 14.2~s on average. Raising the load widened thinking's deficit against fast mode by 0.1970 in posterior expected utility (prespecified test over 12 independent GPU allocations, 95\% $t$ CI $[0.1387,0.2553]$). 

\begin{figure}[t]
\centering
\includegraphics[width=0.62\textwidth]{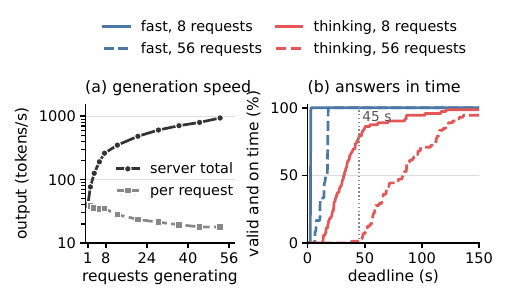}
\caption{Serving load on one GPU (Qwen3-8B). (a) Server output and estimated output per request by the number of requests generating at once, over all bursts, in intervals without prompt processing or waiting requests. (b) Share of the 72 scored answers per condition that were valid and received by each deadline after the burst started. The study used 45~s (dotted). Descriptive, made after the study.}
\label{fig:serving}
\end{figure}

\paragraph{Where the extra time went.} The client and the server recorded when each request was sent, when its first token appeared and when it finished. Table~\ref{tab:serving} uses these records to break down the delay of a thinking answer. Of the 48~s added at high load, 0.6~s passed before the request was sent. Another 7.4~s passed between sending the request and its first token, including waiting to be scheduled and processing the prompt. The other 40.1~s came while the answer was generated. Thinking answers were only slightly longer on average at high load, but their mean generation speed roughly halved. Across all bursts, server output grew from 39 tokens per second with one request generating to 938 with 49 to 56. Over the same range, the estimated speed of each request fell from 39 to 18 tokens per second (Figure~\ref{fig:serving}a). An answer of $n$ tokens, generated at $r$ tokens per second after its first token at time $t_0$, meets a deadline $s$ only if roughly $n\le r(s-t_0)$. With observed mean timings and speeds, this limit fell from about 1{,}500 to 600 tokens at 45~s, while the median thinking answer at high load used about 1{,}260 tokens. This limit is a rough guide. It predicts 11\% of thinking answers on time at high load, against 1.4\% observed (Appendix~B.3.2).  Fast answers had only seven tokens, so most of their delay came before the first token. Figure~\ref{fig:serving}b shows the share in time at every deadline. This breakdown was made after the study and covers one model on one GPU.

\paragraph{Background thinking requests barely delayed fast answers.} A separate prespecified test kept 56 requests per burst and switched 42 of the 50 background requests from the fast to the thinking configuration. Six monitored fast requests slowed by only 0.15~s on average (approximate 95\% interval $[0.08,0.23]$, exact randomization test $p=.0024$), far below the 5~s threshold set in advance, and all stayed valid within 20~s.

\begin{figure*}[t]
\centering
\includegraphics[width=0.82\textwidth]{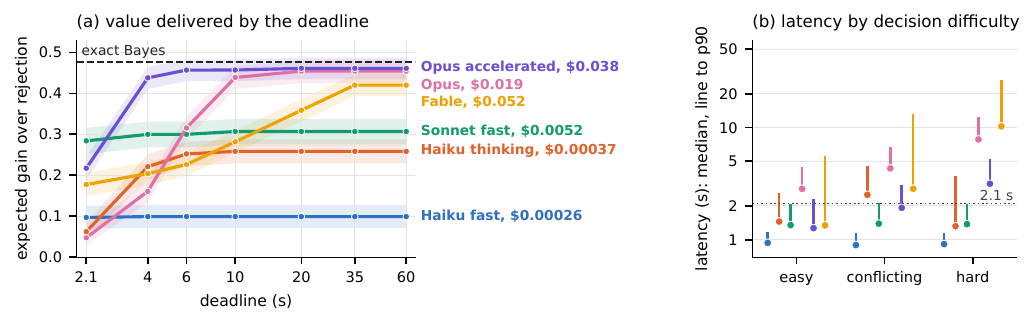}
\caption{Hosted configurations on 480 decisions, 160 per difficulty level and weighted equally. (a) Posterior expected
gain over rejection delivered by each deadline, with pointwise 95\% bootstrap bands and the exact Bayes gain (dashed).
(b) Median arrival time (dot) and 90th percentile (line end) by difficulty, with 2.1~s dotted. Time axes are
logarithmic, and line labels give mean list-price cost per call, with the same colors in (b).}
\label{fig:cost}
\end{figure*}

\begin{table}[t]
\centering
\caption{Where the time of a scored thinking answer went in the serving-load study. Means over 72 answers per load. The first three rows add up to the time from the start of the burst to receipt. Server output is tokens per second over a whole burst. Descriptive, made after the study.}
\label{tab:serving}
\small
\begin{tabular}{@{}lrr@{}}
\toprule
& 8 requests & 56 requests \\
\midrule
burst start to dispatch (s) & 0.13 & 0.71 \\
dispatch to first token (s) & 1.95 & 9.33 \\
generating the answer (s) & 37.29 & 77.37 \\
total time to receipt (s) & 39.37 & 87.41 \\
\addlinespace[2pt]
output tokens per answer & 1{,}359 & 1{,}430 \\
speed per answer (tokens/s) & 35.8 & 17.4 \\
server output (tokens/s) & 140 & 438 \\
valid and on time at 45~s & 79.2\% & 1.4\% \\
\bottomrule
\end{tabular}
\end{table}

\subsection{Requesting both modes}
\label{sec:hybrid}

If thinking answers are valuable but often late, a runtime can request both modes at once. It uses the thinking answer if it is valid and on time, otherwise the fast answer if that is valid and on time, and otherwise the fallback. Like hedged requests~\cite{deanbarroso2013tail}, this sends a backup request, but it prefers the thinking answer whenever that answer is on time. We replayed the rule on the dispatch-clock study's separately recorded answers (600 instances per deadline). Averaged equally over five deadlines, the rule's realized utility exceeded that of the better single mode at each deadline by 0.0077, with the better mode chosen afterward (prespecified, approximate 95\% CI $[0.0013,0.0123]$). Gains were largest at 2.1 and 4~s, where thinking often missed (Figure~\ref{fig:tiered}). The rule cost $1.78\times$ as much as thinking alone. In a live cohort of 600 new instances at 2.1 and 4~s, the gain over the better mode replayed from the same answers was inconclusive ($+0.0125$, 95\% CI $[-0.0008,0.0233]$, Appendix~B.4).

\section{Delivered value across models}
\label{sec:cost}

A laboratory must also choose among models with different costs and latencies. A one-hour study of four hosted models, planned before collection, asks how deadlines change the value each configuration delivers and whether separate summaries mislead that choice. All results here are descriptive.

\paragraph{Design.} We selected 160 new decisions at each of three difficulty levels. We defined the difficulty levels before any model call with two simple rules, one that accepts when the latest quality reading is at least 0.8 and one that applies the same threshold to the mean reading. Both rules choose the best action on \emph{easy} decisions, one does on \emph{conflicting} decisions and neither does on \emph{hard} decisions. These levels make up 79.6\%, 17.9\% and 2.5\% of unselected instances. We weight them equally, so values here are not directly comparable with the replay's (Appendix~B.5 gives population-weighted values). Six configurations answered every decision under the replay's 60~s prompt. Claude Haiku~5.5 used both modes, Sonnet~5.5 used fast mode, Fable~5.1 used thinking, and Opus~5.5 used thinking with standard or paid accelerated output. As in the replay, deadlines were applied to recorded arrival times. Across the main run and the additional deadline-prompt, live-pair and load checks, we made 6{,}724 calls. All returned valid answers, and the study cost \$84 (Appendix~B.5).

\paragraph{Harder decisions slowed the stronger models.} On hard decisions, Opus chose the best action 88\% of the time (91\% with accelerated output) and Fable 60\%, against 17 to 28\% for Haiku and Sonnet without thinking. Standard Opus and Fable also slowed down the most. From easy to hard decisions, their median arrival times rose from 2.8 to 7.8~s and from 1.3 to 10.2~s, while configurations without thinking stayed near 1~s (Figure~\ref{fig:cost}b). By 2.1~s, no configuration delivered more than 0.044 on hard decisions, against 0.422 for the exact Bayes controller.

\paragraph{The best tested configuration depended on the deadline.} By 2.1~s, Sonnet in fast mode delivered the most, 0.284 per decision against an exact Bayes value of 0.476. Standard Opus delivered only 0.047 by 2.1~s because most of its answers were late, although they were worth 0.454 when all were used (Figure~\ref{fig:cost}a). Opus with accelerated output, at twice the token price, delivered the most from 4~s, and standard Opus came within 0.02 of it from 10~s. If accepting a good sample gains $V$ dollars and accepting a bad one loses $V$, a configuration's net value is $V$ times its delivered gain minus its cost per call. Appendix~B.5 gives the resulting break-even values and evaluates combined calls.

\paragraph{Grouping by difficulty changed the error.} At 2.1~s, Haiku thinking's overprediction exceeded fast mode's by 0.073 (95\% CI $[0.056,0.089]$). Grouping by difficulty cut Haiku thinking's overprediction from 0.074 to 0.038, so differences between levels explain about half of it. For Fable, the error was negative when averaged over task family and stage ($-0.026$). Adding difficulty level to these groups brought it to near zero (0.0006, 95\% CI $[-0.011,0.009]$), so the sign of Fable's estimated error depended on grouping. We also used each estimate to choose a configuration on half the decisions and scored that choice on the other half. The two estimates led to different net values in only 4 of 42 combinations of deadline and sample value, and every difference interval included zero. A biased estimate therefore need not change which configuration is chosen, but it misstates how much value that configuration delivers.

\section{Related work}
\label{sec:related}

\paragraph{Reasoning under deadlines.} Gaia2 reports lower performance at higher versus lower effort in time-sensitive scenarios under a simulated clock~\cite{gaia2}. Real-Time Reasoning Gym advances games at a fixed rate with fallback actions at deadlines~\cite{rtgym}.
Timely Machine treats test time as wall-clock time~\cite{ma2026timely}. Under time pressure, reasoning models can stop reasoning and answer incorrectly~\cite{wu2026interruptible}.
\citet{carlon2026shorter} attribute most of lower effort's advantage over higher effort at a 512-token limit to earlier completion.
Other work links longer responses to errors~\cite{hochlehnert2025sober}. Further studies examine insufficient reasoning~\cite{su2025underthinking} and model accuracy and reasoning length jointly~\cite{xu2025lart}.
A position paper on streaming inference formulates expected on-time quality~\cite{zhang2026timeconstraints}. These studies address reasoning quality, response length or time limits. We estimate the value a reasoning mode delivers when late answers are discarded.

\paragraph{Serving agents with deadlines.} TimelyLLM schedules segmented generation around robot execution time and was evaluated with drone and robot-arm workloads on one GPU~\cite{timelyllm}. Kairos serves robot fleets and includes physical-robot validation~\cite{kairos}.
Firm-Deadline Tool Control selects tool-rich or tool-restricted modes using predicted deadline feasibility~\cite{ozturk2026firm}.
Agentix minimizes program latency~\cite{autellix}, while HiveMind schedules concurrent API use~\cite{hivemind}. LLM schedulers target time to first token or differing service-level objectives~\cite{cho2026pascal,zhang2025jitserve}.

\paragraph{Agents in laboratories.} Laboratory software schedules experiments~\cite{alabos}, and LLMs plan and run experiments~\cite{unilabos,panapitiya2026autolabs}. Other work specifies an agent-to-instrument protocol~\cite{lap}, uses LLM agents for computational research~\cite{schmidgall2025agentlaboratory} or repairs and delays conflicting agent actions~\cite{coagent,atomix}. These systems leave open how long the model should reason before a step is due. Laboratories also differ from the robot and game settings above in what a missed deadline costs. A reaction or cell culture cannot wait for the model, and a late answer can waste a sample, reagents and instrument time.

\paragraph{Inference strategies and output constraints.} Hedged requests reduce long response times~\cite{deanbarroso2013tail}, and cascades trade quality against computation~\cite{narasimhan2025cascades}. Anytime algorithms return useful results before a deadline~\cite{zilberstein1996anytime}.
Structured-output constraints can trade validity against correctness~\cite{constrainttax}, as Appendix~B.4 also illustrates.

\section{Limitations and conclusion}
\label{sec:implications}
\label{sec:conclusion}

The exact Bayes controller is optimal for these synthetic decisions, so the LLMs can at best match its expected value. Our results do not show that an LLM should replace such a controller where one exists. The effort comparisons concern Haiku~5.5's provider-defined effort levels, and all hosted models come from one provider. The four-model study also used one collection window, selected decisions and assumed dollar values. The studies use different instances, protocols and outcome measures, and the replay's association between answer value and arrival time is observational. The core studies score one terminal decision per simulated history, so they do not test multistep behavior. The comparisons between adjacent effort levels cannot rule out changes in regret of about 0.02, and Sonnet~5.5's thinking-minus-fast intervals include effects of a few hundredths. Future work includes per-request reasoning policies and multistep tests.

\paragraph{Conclusion.} LLMs increasingly make decisions inside automated laboratories, where steps have hard deadlines and failures make safe fallback actions standard. When an answer misses its deadline, the system executes the fallback, so the value of reasoning depends on both answer quality and arrival time. In our replay, separate summaries predicted about twice the value that thinking delivered. With an unsafe fallback, the same summaries understated thinking's advantage at some deadlines. For Qwen3-8B on one shared GPU, total output tripled under load while the mean generation speed of each thinking answer roughly halved. Evaluations of LLM agents for laboratories should keep each answer's value, arrival time and executed action together. They should compare modes within groups of comparable requests and measure arrival times under realistic serving load. System designers should evaluate net value under their intended fallback and serving load.

\section*{Acknowledgments}
This work was supported in part by the generosity of SoftBank Group Corp. This work was partially supported by National Institutes of Health (NIH) grant 1U54AG075931 and National Science Foundation (NSF) award 2134998 (to J.L.-M.).

\bibliography{references}

\begin{thebibliography}{41}
\providecommand{\natexlab}[1]{#1}
\providecommand{\url}[1]{\texttt{#1}}
\expandafter\ifx\csname urlstyle\endcsname\relax
  \providecommand{\doi}[1]{doi: #1}\else
  \providecommand{\doi}{doi: \begingroup \urlstyle{rm}\Url}\fi

\bibitem[Agyemang et~al.(2026)Agyemang, Kponyo, Somuah, Amponsah, Boakye, and
  Agyekum]{hivemind}
Justice~Owusu Agyemang, Jerry~John Kponyo, Obed~Kwasi Somuah, Elliot Amponsah,
  Godfred Manu~Addo Boakye, and Kwame Opuni-Boachie~Obour Agyekum.
\newblock {HiveMind}: {OS}-inspired scheduling for concurrent {LLM} agent
  workloads.
\newblock arXiv:2604.17111, 2026.

\bibitem[Bateni et~al.(2024)Bateni, Sadeghi, Orouji, Bennett, Punati, Stark,
  Wang, Rosko, Chen, Castellano, Reyes, and Abolhasani]{bateni2024smartdope}
Fazel Bateni, Sina Sadeghi, Negin Orouji, Jeffrey~A. Bennett, Venkat~S. Punati,
  Christine Stark, Junyu Wang, Michael~C. Rosko, Ou~Chen, Felix~N. Castellano,
  Kristofer~G. Reyes, and Milad Abolhasani.
\newblock Smart dope: A self-driving fluidic lab for accelerated development of
  doped perovskite quantum dots.
\newblock \emph{Advanced Energy Materials}, 14\penalty0 (1):\penalty0 2302303,
  2024.
\newblock \doi{10.1002/aenm.202302303}.

\bibitem[Burger et~al.(2020)Burger, Maffettone, Gusev, Aitchison, Bai, Wang,
  Li, Alston, Li, Clowes, Rankin, Harris, Sprick, and Cooper]{burger2020mobile}
Benjamin Burger, Phillip~M. Maffettone, Vladimir~V. Gusev, Catherine~M.
  Aitchison, Yang Bai, Xiaoyan Wang, Xiaobo Li, Ben~M. Alston, Buyi Li, Rob
  Clowes, Nicola Rankin, Brandon Harris, Reiner~Sebastian Sprick, and Andrew~I.
  Cooper.
\newblock A mobile robotic chemist.
\newblock \emph{Nature}, 583\penalty0 (7815):\penalty0 237--241, 2020.
\newblock \doi{10.1038/s41586-020-2442-2}.

\bibitem[Carlon et~al.(2026)Carlon, Ginis, and Algaba]{carlon2026shorter}
Francesca Carlon, Vincent Ginis, and Andres Algaba.
\newblock Shorter reasoning, earlier answers? an evaluation of reasoning
  interfaces, 2026.
\newblock arXiv:2608.03401.

\bibitem[Cho et~al.(2026)Cho, Bang, Hwang, and Rhu]{cho2026pascal}
Eunyeong Cho, Jehyeon Bang, Ranggi Hwang, and Minsoo Rhu.
\newblock {PASCAL}: A phase-aware scheduling algorithm for serving
  reasoning-based large language models.
\newblock In \emph{{IEEE} International Symposium on High Performance Computer
  Architecture ({HPCA})}, pages 1--16, 2026.
\newblock \doi{10.1109/HPCA68181.2026.11408493}.

\bibitem[Dai et~al.(2026)Dai, Ananthanarayanan, Cox, Foukas, Radunovic, and
  Netravali]{kairos}
Yinwei Dai, Ganesh Ananthanarayanan, Landon Cox, Xenofon Foukas, Bozidar
  Radunovic, and Ravi Netravali.
\newblock Kairos: A scalable serving system for physical {AI}, 2026.
\newblock arXiv:2605.11381.

\bibitem[Dean and Barroso(2013)]{deanbarroso2013tail}
Jeffrey Dean and Luiz~Andr{\'e} Barroso.
\newblock The tail at scale.
\newblock \emph{Communications of the {ACM}}, 56\penalty0 (2):\penalty0 74--80,
  2013.
\newblock \doi{10.1145/2408776.2408794}.

\bibitem[{Executive Office of the President}(2025)]{genesis2025}
{Executive Office of the President}.
\newblock Executive order 14363: Launching the {Genesis Mission}.
\newblock
  \url{https://www.federalregister.gov/documents/2025/11/28/2025-21665/launching-the-genesis-mission},
  November 2025.
\newblock Federal Register 90:55035--55039.

\bibitem[Fei et~al.(2024)Fei, Rendy, Kumar, Dartsi, Sahasrabuddhe, McDermott,
  Wang, Szymanski, Walters, Milsted, Zeng, Jain, and Ceder]{alabos}
Yuxing Fei, Bernardus Rendy, Rishi Kumar, Olympia Dartsi, Hrushikesh~P.
  Sahasrabuddhe, Matthew~J. McDermott, Zheren Wang, Nathan~J. Szymanski,
  Lauren~N. Walters, David Milsted, Yan Zeng, Anubhav Jain, and Gerbrand Ceder.
\newblock {AlabOS}: A {Python}-based reconfigurable workflow management
  framework for autonomous laboratories.
\newblock \emph{Digital Discovery}, 3\penalty0 (11):\penalty0 2275--2288, 2024.
\newblock \doi{10.1039/D4DD00129J}.

\bibitem[Froger et~al.(2026)Froger, Andrews, Bettini, Budhiraja,
  Silveira~Cabral, Do, Garreau, Gaya, Lauren{\c{c}}on, Lecanu, Malkan, Mekala,
  M{\'e}nard, Moreno-Torres~Bertran, Piterbarg, Plekhanov, Rita, Rusakov,
  Vorotilov, Wang, Yu, Benhalloum, Mialon, and Scialom]{gaia2}
Romain Froger, Pierre Andrews, Matteo Bettini, Amar Budhiraja, Ricardo
  Silveira~Cabral, Virginie Do, Emilien Garreau, Jean-Baptiste Gaya, Hugo
  Lauren{\c{c}}on, Maxime Lecanu, Kunal Malkan, Dheeraj Mekala, Pierre
  M{\'e}nard, Gerard Moreno-Torres~Bertran, Ulyana Piterbarg, Mikhail
  Plekhanov, Mathieu Rita, Andrey Rusakov, Vladislav Vorotilov, Mengjue Wang,
  Ian Yu, Amine Benhalloum, Gr{\'e}goire Mialon, and Thomas Scialom.
\newblock {Gaia2}: Benchmarking {LLM} agents on dynamic and asynchronous
  environments.
\newblock In \emph{International Conference on Learning Representations
  (ICLR)}, 2026.
\newblock arXiv:2602.11964.

\bibitem[Gao et~al.(2025)Gao, Chang, Que, Xiong, Zhang, Qi, Liu, Wang, Ding,
  Li, Pan, Xie, Yan, Yan, and Zhang]{unilabos}
Jing Gao, Junhan Chang, Haohui Que, Yanfei Xiong, Shixiang Zhang, Xianwei Qi,
  Zhen Liu, Jun-Jie Wang, Qianjun Ding, Xinyu Li, Ziwei Pan, Qiming Xie, Zhuang
  Yan, Junchi Yan, and Linfeng Zhang.
\newblock {UniLabOS}: An {AI}-native operating system for autonomous
  laboratories.
\newblock arXiv:2512.21766, 2025.

\bibitem[Hochlehnert et~al.(2025)Hochlehnert, Bhatnagar, Udandarao, Albanie,
  Prabhu, and Bethge]{hochlehnert2025sober}
Andreas Hochlehnert, Hardik Bhatnagar, Vishaal Udandarao, Samuel Albanie, Ameya
  Prabhu, and Matthias Bethge.
\newblock A sober look at progress in language model reasoning: Pitfalls and
  paths to reproducibility.
\newblock In \emph{Conference on Language Modeling (COLM)}, 2025.
\newblock arXiv:2504.07086.

\bibitem[Itoh et~al.(2021)Itoh, Horinouchi, Uchida, Takahashi, and
  Ozaki]{itoh2021scheduling}
Takeshi~D. Itoh, Takaaki Horinouchi, Hiroki Uchida, Koichi Takahashi, and
  Haruka Ozaki.
\newblock Optimal scheduling for laboratory automation of life science
  experiments with time constraints.
\newblock \emph{{SLAS} Technology}, 26\penalty0 (6):\penalty0 650--659, 2021.
\newblock \doi{10.1177/24726303211021790}.

\bibitem[Kwon et~al.(2023)Kwon, Li, Zhuang, Sheng, Zheng, Yu, Gonzalez, Zhang,
  and Stoica]{vllm}
Woosuk Kwon, Zhuohan Li, Siyuan Zhuang, Ying Sheng, Lianmin Zheng, Cody~Hao Yu,
  Joseph~E. Gonzalez, Hao Zhang, and Ion Stoica.
\newblock Efficient memory management for large language model serving with
  {PagedAttention}.
\newblock In \emph{Proceedings of the 29th Symposium on Operating Systems
  Principles (SOSP)}, pages 611--626, 2023.
\newblock \doi{10.1145/3600006.3613165}.

\bibitem[Lin et~al.(2015)Lin, Kolobov, Kamar, and
  Horvitz]{lin2015metareasoning}
Christopher~H. Lin, Andrey Kolobov, Ece Kamar, and Eric Horvitz.
\newblock Metareasoning for planning under uncertainty.
\newblock In \emph{Proceedings of the Twenty-Fourth International Joint
  Conference on Artificial Intelligence ({IJCAI})}, pages 1601--1608, 2015.
\newblock URL \url{https://www.ijcai.org/proceedings/2015/229}.

\bibitem[Ling et~al.(2026)Ling, Chen, Khandelwal, and Zhong]{timelyllm}
Neiwen Ling, Guojun Chen, Anurag Khandelwal, and Lin Zhong.
\newblock {TimelyLLM}: Time-sensitive {LLM} serving system for physical-{I/O}
  limited agents.
\newblock In \emph{Proceedings of the 24th Annual International Conference on
  Mobile Systems, Applications and Services ({MobiSys})}, pages 261--278, 2026.
\newblock \doi{10.1145/3745756.3809203}.

\bibitem[Luo et~al.(2026)Luo, Shi, Cai, Zhang, Wong, Wang, Wang, Huang, Chen,
  Gonzalez, and Stoica]{autellix}
Michael Luo, Xiaoxiang Shi, Colin Cai, Tianjun Zhang, Justin Wong, Yichuan
  Wang, Chi Wang, Yanping Huang, Zhifeng Chen, Joseph~E. Gonzalez, and Ion
  Stoica.
\newblock {Agentix}: An efficient serving engine for {LLM} agents as general
  programs.
\newblock In \emph{23rd {USENIX} Symposium on Networked Systems Design and
  Implementation ({NSDI} 26)}, pages 2443--2459. {USENIX} Association, 2026.
\newblock Preprint titled Autellix, arXiv:2502.13965.

\bibitem[Lyu et~al.(2026)Lyu, Zhang, Wu, Wei, and Chen]{coagent}
Hongtao Lyu, Dingyan Zhang, Mingyu Wu, Xingda Wei, and Haibo Chen.
\newblock {CoAgent}: Concurrency control for multi-agent systems.
\newblock arXiv:2606.15376, 2026.

\bibitem[Ma et~al.(2025)Ma, He, Snell, Griggs, Min, and
  Zaharia]{ma2025nothinking}
Wenjie Ma, Jingxuan He, Charlie Snell, Tyler Griggs, Sewon Min, and Matei
  Zaharia.
\newblock Reasoning models can be effective without thinking, 2025.
\newblock arXiv:2504.09858.

\bibitem[Ma et~al.(2026)Ma, Li, Chen, Li, Li, Guo, Lin, and Chen]{ma2026timely}
Yichuan Ma, Linyang Li, Yongkang Chen, Peiji Li, Xiaozhe Li, Qipeng Guo, Dahua
  Lin, and Kai Chen.
\newblock Timely machine: Awareness of time makes test-time scaling agentic.
\newblock In \emph{Proceedings of the 64th Annual Meeting of the {Association
  for Computational Linguistics} (Volume 1: Long Papers)}, pages 4619--4636.
  Association for Computational Linguistics, 2026.
\newblock \doi{10.18653/v1/2026.acl-long.211}.

\bibitem[Mohammadi et~al.(2026)Mohammadi, Potamitis, Klein, Arora, and
  Bindschaedler]{atomix}
Bardia Mohammadi, Nearchos Potamitis, Lars Klein, Akhil Arora, and Laurent
  Bindschaedler.
\newblock {Atomix}: Timely, transactional tool use for reliable agentic
  workflows.
\newblock arXiv:2602.14849, 2026.

\bibitem[Narasimhan et~al.(2025)Narasimhan, Jitkrittum, Rawat, Kim, Gupta,
  Menon, and Kumar]{narasimhan2025cascades}
Harikrishna Narasimhan, Wittawat Jitkrittum, Ankit~Singh Rawat, Seungyeon Kim,
  Neha Gupta, Aditya~Krishna Menon, and Sanjiv Kumar.
\newblock Faster cascades via speculative decoding.
\newblock In \emph{International Conference on Learning Representations
  (ICLR)}, 2025.
\newblock arXiv:2405.19261.

\bibitem[{OECD}(2023)]{oecd2023aiscience}
{OECD}.
\newblock \emph{Artificial Intelligence in Science: Challenges, Opportunities
  and the Future of Research}.
\newblock OECD Publishing, Paris, 2023.
\newblock \doi{10.1787/a8d820bd-en}.

\bibitem[\"{O}zt\"{u}rk and Artuner(2026)]{ozturk2026firm}
Muhammet~Ali \"{O}zt\"{u}rk and Harun Artuner.
\newblock Tool-augmented {LLM} serving under firm deadlines: A queueing-control
  approach, 2026.
\newblock SSRN preprint.

\bibitem[Panapitiya et~al.(2026)Panapitiya, Saldanha, Job, and
  Hess]{panapitiya2026autolabs}
Gihan Panapitiya, Emily Saldanha, Heather Job, and Olivia Hess.
\newblock {AutoLabs}: cognitive multi-agent systems with self-correction for
  autonomous chemical experimentation.
\newblock \emph{Scientific Reports}, 16\penalty0 (1):\penalty0 19554, 2026.
\newblock \doi{10.1038/s41598-026-45593-z}.

\bibitem[Ray(2026)]{constrainttax}
Jaideep Ray.
\newblock The constraint tax: Measuring validity-correctness tradeoffs in
  structured outputs for small language models, 2026.
\newblock arXiv:2605.26128.

\bibitem[Reddi et~al.(2020)Reddi, Cheng, Kanter, Mattson, Schmuelling, Wu,
  Anderson, Breughe, Charlebois, Chou, Chukka, Coleman, Davis, Deng, Diamos,
  Duke, Fick, Gardner, Hubara, Idgunji, Jablin, Jiao, {St. John}, Kanwar, Lee,
  Liao, Lokhmotov, Massa, Meng, Micikevicius, Osborne, Pekhimenko, Rajan,
  Sequeira, Sirasao, Sun, Tang, Thomson, Wei, Wu, Xu, Yamada, Yu, Yuan, Zhong,
  Zhang, and Zhou]{reddi2020mlperf}
Vijay~Janapa Reddi, Christine Cheng, David Kanter, Peter Mattson, Guenther
  Schmuelling, Carole-Jean Wu, Brian Anderson, Maximilien Breughe, Mark
  Charlebois, William Chou, Ramesh Chukka, Cody Coleman, Sam Davis, Pan Deng,
  Greg Diamos, Jared Duke, Dave Fick, J.~Scott Gardner, Itay Hubara, Sachin
  Idgunji, Thomas~B. Jablin, Jeff Jiao, Tom {St. John}, Pankaj Kanwar, David
  Lee, Jeffery Liao, Anton Lokhmotov, Francisco Massa, Peng Meng, Paulius
  Micikevicius, Colin Osborne, Gennady Pekhimenko, Arun Tejusve~Raghunath
  Rajan, Dilip Sequeira, Ashish Sirasao, Fei Sun, Hanlin Tang, Michael Thomson,
  Frank Wei, Ephrem Wu, Lingjie Xu, Koichi Yamada, Bing Yu, George Yuan, Aaron
  Zhong, Peizhao Zhang, and Yuchen Zhou.
\newblock {MLPerf} inference benchmark.
\newblock In \emph{Proceedings of the {ACM/IEEE} 47th Annual International
  Symposium on Computer Architecture ({ISCA})}, pages 446--459, 2020.
\newblock \doi{10.1109/ISCA45697.2020.00045}.

\bibitem[Russell and Wefald(1991)]{russell1991metareasoning}
Stuart Russell and Eric Wefald.
\newblock Principles of metareasoning.
\newblock \emph{Artificial Intelligence}, 49\penalty0 (1--3):\penalty0
  361--395, 1991.
\newblock \doi{10.1016/0004-3702(91)90015-C}.

\bibitem[Schmidgall et~al.(2025)Schmidgall, Su, Wang, Sun, Wu, Yu, Liu, Moor,
  Liu, and Barsoum]{schmidgall2025agentlaboratory}
Samuel Schmidgall, Yusheng Su, Ze~Wang, Ximeng Sun, Jialian Wu, Xiaodong Yu,
  Jiang Liu, Michael Moor, Zicheng Liu, and Emad Barsoum.
\newblock Agent laboratory: Using {LLM} agents as research assistants.
\newblock In \emph{Findings of the Association for Computational Linguistics:
  EMNLP 2025}, pages 5977--6043, 2025.
\newblock \doi{10.18653/v1/2025.findings-emnlp.320}.

\bibitem[Snell et~al.(2025)Snell, Lee, Xu, and Kumar]{snell2025scaling}
Charlie Snell, Jaehoon Lee, Kelvin Xu, and Aviral Kumar.
\newblock Scaling {LLM} test-time compute optimally can be more effective than
  scaling parameters for reasoning.
\newblock In \emph{International Conference on Learning Representations
  (ICLR)}, 2025.
\newblock URL \url{https://openreview.net/forum?id=4FWAwZtd2n}.

\bibitem[Su et~al.(2025)Su, Healey, Nakov, and Cardie]{su2025underthinking}
Jinyan Su, Jennifer Healey, Preslav Nakov, and Claire Cardie.
\newblock Between underthinking and overthinking: An empirical study of
  reasoning length and correctness in {LLMs}, 2025.
\newblock arXiv:2505.00127.

\bibitem[{The White House}(2025)]{aiactionplan2025}
{The White House}.
\newblock America's {AI} action plan.
\newblock
  \url{https://www.whitehouse.gov/wp-content/uploads/2025/07/Americas-AI-Action-Plan.pdf},
  July 2025.

\bibitem[{U.S. National Science Foundation}(2026)]{nsf2026cloudlabs}
{U.S. National Science Foundation}.
\newblock {NSF} announces \$400{M} investment in new national network of
  {AI}-programmable cloud laboratories in alignment with the {U.S.}
  government's {Genesis Mission}.
\newblock
  \url{https://www.nsf.gov/tip/updates/nsf-announces-400m-investment-new-national-network-ai},
  July 2026.

\bibitem[Wen et~al.(2026)Wen, Ye, Zhang, Yang, and Zhu]{rtgym}
Yule Wen, Yixin Ye, Yanzhe Zhang, Diyi Yang, and Hao Zhu.
\newblock Real-time reasoning agents in evolving environments.
\newblock In \emph{International Conference on Learning Representations
  (ICLR)}, 2026.
\newblock arXiv:2511.04898.

\bibitem[Wu et~al.(2026)Wu, Miroyan, Chan, Darrell, Norouzi, and
  Gonzalez]{wu2026interruptible}
Tsung-Han Wu, Mihran Miroyan, David~M. Chan, Trevor Darrell, Narges Norouzi,
  and Joseph~E. Gonzalez.
\newblock Are large reasoning models interruptible?
\newblock In \emph{Proceedings of the 43rd International Conference on Machine
  Learning (ICML)}, volume 306 of \emph{Proceedings of Machine Learning
  Research}, pages 136539--136581, 2026.

\bibitem[Xu et~al.(2026)Xu, Liu, Wang, and Gu]{xu2025lart}
Zhiyu Xu, Jia Liu, Yixin Wang, and Yuqi Gu.
\newblock Latency-response theory model: Evaluating large language models via
  response accuracy and chain-of-thought length.
\newblock \emph{The Annals of Applied Statistics}, 20\penalty0 (3):\penalty0
  2253--2273, 2026.
\newblock \doi{10.1214/26-aoas2236}.

\bibitem[Zhang et~al.(2026)Zhang, Wu, Mu, Ning, Liu, Sarda, Lee, and
  Lai]{zhang2025jitserve}
Wei Zhang, Zhiyu Wu, Yi~Mu, Rui Ning, Banruo Liu, Nikhil Sarda, Myungjin Lee,
  and Fan Lai.
\newblock {JITServe}: {SLO}-aware {LLM} serving with imprecise request
  information.
\newblock In \emph{23rd {USENIX} Symposium on Networked Systems Design and
  Implementation ({NSDI} 26)}, pages 825--848. {USENIX} Association, 2026.

\bibitem[Zhang(2026)]{zhang2026timeconstraints}
Xiaotian Zhang.
\newblock Intelligence under time constraints: Rethinking test-time compute,
  2026.
\newblock arXiv:2609.14995.

\bibitem[Zhong et~al.(2024)Zhong, Liu, Chen, Hu, Zhu, Liu, Jin, and
  Zhang]{zhong2024distserve}
Yinmin Zhong, Shengyu Liu, Junda Chen, Jianbo Hu, Yibo Zhu, Xuanzhe Liu, Xin
  Jin, and Hao Zhang.
\newblock {DistServe}: Disaggregating prefill and decoding for
  goodput-optimized large language model serving.
\newblock In \emph{18th {USENIX} Symposium on Operating Systems Design and
  Implementation ({OSDI})}, pages 193--210. {USENIX} Association, 2024.

\bibitem[Zhu et~al.(2026)Zhu, Gao, Chen, Zhu, and Huang]{lap}
Linwu Zhu, Liqiang Gao, Yan Chen, Dan Zhu, and Jian Huang.
\newblock {LAP}: An agent-to-instrument protocol for autonomous science.
\newblock arXiv:2606.03755, 2026.

\bibitem[Zilberstein(1996)]{zilberstein1996anytime}
Shlomo Zilberstein.
\newblock Using anytime algorithms in intelligent systems.
\newblock \emph{{AI} Magazine}, 17\penalty0 (3):\penalty0 73--83, 1996.
\newblock \doi{10.1609/aimag.v17i3.1232}.

\end{thebibliography}
\bibliographystyle{plainnat}

\section*{AI Usage Disclosure}
LLMs (Claude Sonnet and Opus 5.5) were used for proofreading, grammar checking, and refining readability in the main manuscript. For the supplementary material, LLMs were also used to assist in drafting and reformatting internal project reports into Appendix sections. The LLMs were not involved in defining research scope, experimental design, data collection, or interpreting results. All scientific claims and final text across the manuscript and supplement were thoroughly audited and verified by the authors, who take full responsibility for the work.

\clearpage
\section*{How the appendix is organized}
\addcontentsline{toc}{section}{How the appendix is organized}
Part~\ref{app:partA} reports the methods and results of the paper's three studies of hosted Haiku~5.5 on synthetic
sample-stopping decisions. Part~\ref{app:partB} reports additional studies and scope checks, including favorable, unfavorable
and inconclusive results. Section~7 summarizes the constructed-load study and the shared-serving test of Appendix~\ref{app:serving} and concurrent selection from Appendix~\ref{app:concurrent}. Section~8 summarizes the hosted cost and difficulty study of Appendix~\ref{app:cost}. The other studies in Part~\ref{app:partB} are not used to support the paper's claims. Part~\ref{app:partC} records the study sequence, costs, deviations and computation checks.
Table~\ref{tab:nav} maps the core claims and the Sonnet scope check to their evidence. The fault, local-model and GPU-load studies are in Appendices~\ref{app:faults}, \ref{app:local} and \ref{app:serving}, respectively. Study history, accounting, deviations and computation checks are in Appendices~\ref{app:sequence}, \ref{app:accounting}, \ref{app:deviations} and \ref{app:verification}, respectively.

\begin{table}[H]
\centering
\caption{Main claims, the studies behind them and where each is reported in full. A result is ``confirmatory'' when its test was fixed in a study plan with error control within its prespecified family of tests. Other results are descriptive.}
\label{tab:nav}
\footnotesize
\setlength{\tabcolsep}{3pt}
\begin{tabular}{@{}>{\raggedright\arraybackslash}p{0.2\textwidth}>{\raggedright\arraybackslash}p{0.15\textwidth}>{\raggedright\arraybackslash}p{0.14\textwidth}>{\raggedright\arraybackslash}p{0.2\textwidth}>{\raggedright\arraybackslash}p{0.13\textwidth}>{\raggedright\arraybackslash}p{0.08\textwidth}@{}}
\toprule
main claim & study & independent unit & primary endpoint & evidence status & location \\
\midrule
Low-effort thinking improves returned terminal decisions; higher effort adds no detected improvement & terminal decision quality (1{,}200 new instances) & instance & terminal decision regret, thinking minus fast; medium minus low; high minus medium & confirmatory (four-test and two-test families) & \ref{app:quality} \\
\addlinespace[3pt]
Deadlines limit how much of the improvement is realized in execution & dispatch-clock deadline execution (600 new instances) & instance & growth in thinking's realized-utility advantage from 2.1 to 60~s & confirmatory (four-test family) & \ref{app:deadline} \\
\addlinespace[3pt]
Thinking's advantage grows with the deadline when preparation time counts; both modes' utility is near fallback utility at 2.1~s & evidence-onset deadline execution (600 new instances) & instance & growth in thinking's realized-utility advantage from 2.1 to 60~s & confirmatory (four-test family) & \ref{app:onset} \\
\addlinespace[3pt]
Overprediction of thinking's advantage recurs on new instances & replay replication (300 new instances) & instance & $D$ at 2.1~s & confirmatory (single test); planned after the original analysis & \ref{app:followup} \\
\addlinespace[3pt]
Separate summaries overestimate thinking's delivered expected advantage over fast mode & prospective fixed-answer replay (1{,}200 new instances) & instance, within 12 groups & overprediction difference $D$ at 2.1~s & confirmatory (single test) & \ref{app:dependence} \\
\addlinespace[3pt]
Scope: a second hosted model & Sonnet~5.5 regret and dispatch-clock studies & instance & regret contrast; deadline interaction & confirmatory (two-test family); neither contrast detected & \ref{app:quality} \\
\addlinespace[3pt]
Shared serving load on one local GPU makes most thinking answers late & constructed load (12 allocations, Qwen3-8B on one A6000) & allocation & growth $W$ of thinking's executed-utility deficit from 8 to 56 requests & confirmatory (single test) & \ref{app:load} \\
\addlinespace[3pt]
Requesting both modes may recover some value lost to lateness & concurrent selection using recorded answers (600 instances) and live requests (600 new instances) & instance & selection minus better fixed mode in realized utility & confirmatory (two single tests); recorded-answer gain detected, live gain not detected & \ref{app:concurrent} \\
\bottomrule
\end{tabular}
\end{table}

\appendix
\section{Core decision and deadline studies}
\label{app:partA}

This part reports the three core studies in full. The studies examine terminal decision quality, execution before deadlines and prospective replay. They use different new instances and protocols, with no single paired set of instances across studies.

\subsection{Tasks, measurement and reference controllers}
\label{app:comparators}
Table~\ref{tab:glossary} defines terms used in this appendix. Sections~2, 3, 5 and 6 define deadlines, the dispatch clock, the evidence-onset clock, the interaction, observation stages, test families and pooling. The fault-diagnosis model, used
only by the additional studies, is specified in Appendix~\ref{app:faults}.
\begin{table}[H]
\centering
\caption{Terms used in this appendix.}
\label{tab:glossary}
\small
\setlength{\tabcolsep}{4pt}
\begin{tabular}{@{}>{\raggedright\arraybackslash}p{0.24\textwidth}>{\raggedright\arraybackslash}p{0.72\textwidth}@{}}
\toprule
term & meaning \\
\midrule
terminal observation & an observation requiring a final decision: there is not enough time to continue in the decision-quality study, and continuation is excluded from the available actions in the deadline and replay studies \\
fallback & the preset action executed when an answer is late or unusable: reject for stopping and scientific faults, quarantine for physical faults; in the separate output-constraint protocol, a rejected proposal leaves the process waiting \\
posterior expected utility & the exactly computed expected value of an action given the evidence \\
simulator-realized utility & the value an action produced in the simulated run \\
task family & one of the two stopping processes (optical-fiber draw or quantum-dot synthesis) \\
cell & one experimental condition, such as one mode at one deadline \\
block & one collection batch \\
group (stratum) & comparable instances with the same task family and observation stage, and also the same collection batch in the replay analysis; stratified resampling draws instances within each group \\
allocation & one separate GPU server reservation during which every condition of a local serving study is run \\
background requests & requests that only generate load \\
better fixed mode & the single mode with higher utility at a given deadline, chosen in hindsight from the same data \\
\bottomrule
\end{tabular}
\end{table}

\paragraph{Stopping model.}
The two stopping processes are an optical-fiber draw and a quantum-dot synthesis.
Before any measurement, every combination of the grid values and rates below is equally likely. We do not approximate the posterior with a continuous-parameter distribution. The initial position is zero.
\begin{table}[H]
\centering\small
\caption{Finite stopping model and observation likelihoods. Noise terms are independent.}
\label{tab:stopping-model}
\begin{tabular}{@{}>{\raggedright\arraybackslash}p{4cm}>{\raggedright\arraybackslash}p{12cm}@{}}
\toprule
quantity & definition \\
\midrule
optimum $z$, rate $r$ & $z\in\{-1+i/10:i=0,\ldots,20\}$; $r\in\{1.5,3.\}$; uniform over 21 optima $\times$ 2 rates \\
quality at position $x$ & $h(x;z,r)=0.98\exp[-r(x-z)^2]$; good iff $h\geq0.8$ \\
quality observation $y$ & $y=h(x;z,r)+\mathrm{Normal}(0,\sigma_y)$ \\
residual observation $d$ & $d=z-x+\mathrm{Normal}(0,\sigma_d)$ \\
optical-fiber measurement / next parameters & $(\sigma_y,\sigma_d)=(.05,.08)$ \\
quantum-dot measurement / low resolution / tune & $(\sigma_y,\sigma_d)=(.10,.25)$ \\
quantum-dot high resolution & $(\sigma_y,\sigma_d)=(.02,.04)$ \\
work penalty, reserve, removal & $10^{-5}$ per charged physical second; 2~s per decision reserve; 10~s removal \\
\bottomrule
\end{tabular}
\end{table}
Here $\sigma_y$ and $\sigma_d$ are noise standard deviations. For each model $(z,r)$ the exact discrete posterior weight is the normalized exponential of
\[
\ell_{z,r}=\log(1/42)-\frac12\sum_{j:\,e_j=e_{\rm current}}
 \left[\left(\frac{y_j-h(x_j;z,r)}{\sigma_{y,j}}\right)^2+
       \left(\frac{d_j-(z-x_j)}{\sigma_{d,j}}\right)^2\right].
\]
The sum includes history and current evidence from the current epoch, the period since the sample was last remade. Remaking a sample draws a new latent quality and starts a new epoch; no fiber or quantum-dot action does so. The implementation subtracts
$\max\ell$ before exponentiating and normalizing. The posterior probability of a good sample is the sum of the weights
whose quality at the current position meets the threshold. Terminal values are
$Q(\texttt{accept})=2p-1-10^{-5}\cdot10$ and $Q(\texttt{reject})=-10^{-5}\cdot10$.
\begin{table}[H]
\centering\small
\caption{Stopping actions, durations and transitions. State changes occur on physical completion.}
\begin{tabular}{@{}lllp{7.6cm}@{}}
\toprule
process & action & duration (s) & transition \\
\midrule
both & accept; reject & 0; 0 & Terminal disposition, with the same charged 10~s in-module removal \\
optical fiber & next\_parameters & 5 & Set position to parameter; preserve latent model and epoch \\
quantum dot & low\_resolution & 68 & Preserve quality, position and epoch; take another measurement \\
quantum dot & high\_resolution & 308 & Preserve quality, position and epoch; use high-resolution likelihood \\
quantum dot & tune & 103 & Set position to parameter; preserve latent model and epoch (35~s tuning then 68~s scan) \\
\bottomrule
\end{tabular}
\end{table}
An actuator parameter is a number
in $[-1,1]$; the reference's candidate set is the finite grid above. Each native action schema is an \texttt{anyOf}
of objects requiring \texttt{action}, and requiring \texttt{parameter} only for \texttt{next\_parameters} or
\texttt{tune}; extra properties are forbidden. The task's prompt field contains the following verbatim text.
\begin{quote}\small
Choose a schema action using the task, likelihood, costs, cutoff and history. Return JSON.
\end{quote}
The public observation includes this prompt, task and prior, action schema and costs, cutoff, evidence-availability
time, current evidence, history, action history and remaining rounds. Observation noise is drawn after any completed
transition; earlier observations are retained in order. Work already running may finish after the cutoff, and removal
is still charged, but no productive work may start or be committed after it. At terminal observations no
continuation fits with the required decision reserve and removal; a continuation request scores as rejection minus the
charged work, and no new measurement is taken. When time still permits
continuation, scores use an approximate reference that looks one action ahead (for a quantum-dot measurement, the measurement followed by at most one tuning step); the exact optimum over all remaining steps is not computed.

\paragraph{Instance generation and conditioning on stage.}
Isolated trajectories in the original task allow five rounds. Before the selected observation, a fixed precursor policy sets
successive positions to $(-.6,.4,-.2,.8)$, using \texttt{next\_parameters} or \texttt{tune}.
The first, middle and late stages use generations 0, 2 and 4, respectively. Their observed histories contain 1, 3 or 5 measurements (including current evidence). Each history starts at position zero and follows the corresponding prefix of that sequence, with strictly increasing timestamps. The remaining native rounds are $5-\text{generation}$. The precursor policy generates histories for the prescribed observation stage, independently of the model's preferred actions.
The original paired horizons use a cutoff at evidence availability plus 600~s (ample) or plus removal, reserve
and .1~s (terminal); the latter is checked against every continuation cost. The local evaluation's terminal panel
selected the first qualifying candidates from fixed pools using posterior bins with edges
$(.1,.3,.5,.7,.9,.98)$ and quotas $(3,3,3,3,2,2,2)$ per family; its ample panel uses ordered native draws.
The earlier hosted effort and prior-sensitivity panels instead cycle through stages using seeds in order. They do not select instances by posterior probability, proximity to a boundary or confidence. The two timing studies also use draws in order without rejecting any draws, alternating families and cycling through stages. Their scored instances are equally divided among groups defined by task family and observation stage. Preliminary-check and pilot instances are
separate. Each scored instance receives a separate response for every combination of deadline and mode.

\paragraph{Stopping prompts and schemas.}
The stopping system message is given verbatim below.
\begin{quote}\small
Use the public task and evidence. Return only a JSON object satisfying the supplied action schema.
\end{quote}
The user message is canonical JSON, with sorted keys and no separator whitespace, containing
\texttt{\{"observation":\ldots\}}. Raw prompts add no reference answer. The local evaluation also adds posterior information, guidance on utility and stopping, or both to the prompt. Posterior information gives the finite probability distribution, the current probability of a good sample and the probabilities of candidate grid positions. The guidance is given verbatim below.
\begin{quote}\small
Utility is valid output minus false acceptance minus damage minus 0.00001 times charged physical seconds. For a terminal scientific choice with current threshold probability p, accept has value 2*p-1 minus removal cost; reject has value minus removal cost. There is no universal 0.98 acceptance rule. A continuation uses one round. With remaining\_rounds \textless{}= 1, a continuation request aborts without a new measurement. Include native action cost, a decision reserve for each subsequent decision, and terminal removal in cutoff feasibility. Running operations and removal are charged even after cutoff; cleanup does not restore completion credit. Incident continue requires confirmed success; confirmed false or source is established failure, for which a feasible retry can recover. Flag/escalate contains and abandons, with zero recovery credit. Retry at confirmed success causes damage at physical completion.
\end{quote}
The sentences about incidents, flags and retries concern other task types in the same environment and do not apply to stopping decisions. The provider-compatible schema recursively
moves numeric, string and array bounds into a \texttt{Required bounds:} description (retaining \texttt{minItems}
when zero or one); local validation still enforces the original bounds.

\paragraph{Accept-or-reject requests under both timing protocols.}
We replace the task's original schema with the following schema, reproduced exactly.
{\small
\begin{verbatim}
{"type":"object","required":["action"],"additionalProperties":false,
 "properties":{"action":{"type":"string","enum":["accept","reject"]}}}
\end{verbatim}
}
We set remaining rounds to 1 and cutoff to evidence availability plus deadline plus removal. We append the following text exactly, including the leading space.
{\small
\begin{verbatim}
 Forced terminal contract: only accept or reject; no continuation is permitted.
\end{verbatim}
}
In the Haiku clock studies, transport error, refusal and \texttt{max\_tokens} map to reject with failure labels \texttt{transport},
\texttt{refusal} and \texttt{truncation}. JSON decode errors (including excessive nesting), non-objects, extra keys,
or an action outside accept/reject map to reject/\texttt{malformed}. A valid object maps to its action and no failure. Lateness is flagged separately from these failure labels.

The user message for the illustrative scored instance at 2.1~s is shown below. Ellipses mark fields omitted for display and were not part of the transmitted text. Line breaks are also for display.
{\footnotesize
\begin{verbatim}
{"observation":{"action_costs_s":{"accept":0.0,"next_parameters":5.0,
"reject":0.0},... ,"decision_type":"fibre","evidence":{"epoch":0,
"noise_sd":0.05,"observed_at_s":5.0,"position":0.0,
"position_residual":0.16731737999295462,"quality_estimate":0.9298116759694084,
"residual_sd":0.08,"sample_id":"probe-963026:sample"},
"evidence_available_at_s":5.0,"generation":0,"history":[],
"physical_cutoff_s":17.1,"prompt":"Choose a schema action using the task, likelihood,
costs, cutoff and history. Return JSON. Forced terminal contract: only accept or
reject; no continuation is permitted.","remaining_rounds":1,"task":{...},...}}
\end{verbatim}
}

\paragraph{Hosted model identifiers and requests.}
Haiku requests use \texttt{claude-haiku-5-5}. In the effort, prior-sensitivity and later Haiku studies,
\texttt{max\_tokens}=16000 and \texttt{thinking} is either \texttt{\{type: disabled\}} (fast) or
\texttt{\{type: adaptive, display: summarized\}} (adaptive). The earlier hosted stopping evaluation and the fault
challenge used two modes: a fast mode (\texttt{thinking=\{type: disabled\}}, effort low, \texttt{max\_tokens}=2048)
and a thinking mode that omitted the \texttt{thinking} field (provider default) at effort medium with
\texttt{max\_tokens}=16000; its recorded responses contain thinking blocks.
The \texttt{output\_config} contains the named effort and \texttt{format=\{type: json\_schema, schema: ...\}};
there are no sampling parameters. The prior-sensitivity panel's fast-mode output cap is also 2048. Sonnet studies use \texttt{claude-sonnet-5-5}, effort low and a 16000-token
cap, with \texttt{thinking=\{type: between\_tools\}} for their baseline and \texttt{\{type: adaptive\}} for
low-effort thinking, preserving the inherited JSON-schema request format. The Sonnet baseline is never disabled thinking.

\paragraph{Illustrative scored decision (dispatch clock).}
The recorded instance is a first-stage optical-fiber observation with no earlier history. At position 0, the quality estimate is 0.9298 (noise standard deviation .05), and the position residual is 0.1673 (standard deviation .08).
The permitted actions at this final decision are accept and reject. We compute the exact posterior by evaluating the normalized finite weight formula above on this observation. The posterior probability of a good sample is $p=(Q(\texttt{accept})+1+10^{-5}\cdot10)/2$. The recorded expected values are $Q(\texttt{accept})=0.99958$ and
$Q(\texttt{reject})=-0.0001$, so accept is optimal. The underlying simulated sample is good, as confirmed by
completion when accepted. All records below are from scored calls. At both deadlines, the response JSON is exactly \texttt{\{"action":"reject"\}} for fast mode and \texttt{\{"action":"accept"\}} for low-effort thinking.
\begin{table}[H]
\centering\footnotesize
\setlength{\tabcolsep}{3pt}
\caption{Illustrative recorded responses and execution (no new contrast). Latency is measured from dispatch to receipt on the dispatch clock. We retain late responses without executing them. ``Answer used'' records whether the returned answer was executed. Values are rounded for display. Each deadline uses a separate response.}
\label{tab:worked}
\begin{tabular}{@{}llrlrl@{}}
\toprule
deadline (s) & mode & latency (s) & executed action & realized utility & answer used / late / goal \\
\midrule
2.1 & fast & 0.929 & reject & $-0.0001$ & yes / no / no \\
2.1 & low & 3.652 & reject & $-0.0001$ & no / yes / no \\
60 & fast & 1.082 & reject & $-0.0001$ & yes / no / no \\
60 & low & 5.163 & accept & 0.9999 & yes / no / yes \\
\bottomrule
\end{tabular}
\end{table}
At the short deadline, the fallback replaces the better answer. At the long deadline, the better answer is executed. This case does not estimate how often that pattern occurs.

For terminal stopping and fault decisions, \emph{exact Bayes} chooses the action with the highest posterior expected value, using exact rational arithmetic for faults. Regret is the shortfall from the Bayes-optimal decision's expected value, so exact Bayes has zero regret. For stopping decisions, the prespecified \emph{practical} rules use the task's reference procedure (equal to Bayes at terminal observations), a fixed posterior threshold (accept only if $p\geq0.98$), a naive majority of public quality estimates, or rejection in every case. For faults, the rules use a prior-sensitive posterior-threshold rule (defined in Appendix~\ref{app:faults}), the original feature-counting heuristic, or the fallback. All of these rules are deterministic functions of public information. We treat their computation latency as zero.

\subsection{Core study configurations}
\label{app:hosted}
Table~\ref{tab:coverage} in Appendix~\ref{app:sequence} lists study sizes and selected costs. Appendix~\ref{app:accounting} explains the accounting. The paragraphs below describe the configurations of the hosted studies used in the main text. Each additional study's configuration is given with that study in Part~\ref{app:partB}.

\paragraph{Hosted requests.} For Claude Haiku 5.5 (\texttt{claude-haiku-5-5}), fast mode disables thinking and uses low effort. Adaptive thinking uses low, medium or high effort, abbreviated low, medium and high in tables. The main hosted comparisons use a 16{,}000-token output cap, a JSON output schema, no sampling parameters and zero automatic retries. Appendix~\ref{app:comparators} lists earlier fast-mode exceptions with a cap of 2{,}048 tokens. The terminal and fault comparisons, medium- and high-effort comparisons, and both timing protocols send requests one at a time. The latency check and two live cohorts use prespecified schedules that dispatch requests in pairs with six-second spacing. A run stops after five consecutive transport failures.

\paragraph{New terminal and fixed-model fault instances.} The stopping comparison uses 1{,}200 new, unselected terminal observations generated from the stated priors and observation model. The observations are balanced over two families and three observation stages, with one response per mode. The fault comparison uses all 128 fault-family/sensor-vector combinations, with three
  new responses per mode.

\paragraph{Medium and high effort.} Medium-effort and high-effort thinking each return one response for each of the same 1{,}200 instances. The observations are identical byte for byte, and we interleave requests for the two effort levels. A prespecified rule would have dropped high-effort thinking if the pilot projected more than 20~h. The rule depended only on projected time. The pilot projected 5.4~h, so the rule did not apply.

\paragraph{Timing from evidence onset.} Both modes receive 600 new terminal instances generated from seeds not used in the initial sweep or the terminal comparison. Each instance is presented with deadlines of 2.1, 10 and 60~s. The simulator executes the returned action if the answer is valid and on time, and rejects the sample otherwise. Goal completion requires accepting a valid sample and completing its removal within the window.

\paragraph{Timing from provider dispatch.} Both modes receive another 600 new instances at 2.1, 4, 6, 10 and 60~s, using the same request contents, schema, execution and scoring as the evidence-onset protocol. The dispatch-clock protocol uses new instances and five deadlines instead of three. The dispatch clock starts when the provider is called, after input-token counting, pacing and admission. We hold the simulator at evidence onset until dispatch. The physical cutoff at evidence onset + deadline + 10~s in simulated time therefore falls at dispatch + deadline + 10~s in wall time.

\paragraph{Sonnet study design.} The study plan specified two Bonferroni tests at two-sided $\alpha=.025$ with 97.5\% intervals. Requests follow the terminal-regret and dispatch-clock protocols, with only the model and thinking fields changed. Fast mode uses low effort and permits no thinking before the answer (Sonnet~5.5 does not allow thinking to be disabled). Low-effort thinking uses adaptive thinking at low effort. The Sonnet regret comparison uses 1{,}200 new instances,
  one response per mode and terminal-regret scoring. The Sonnet deadline comparison uses 600 new instances at 2.1 and 60~s on the dispatch clock.

Inference accounting is described in Appendix~\ref{app:accounting}.

\subsection{Statistical procedures and analysis families}
\label{app:statistics}
For each study, we fixed the inputs, request schedule, failure handling and analysis before collecting scored responses.
Follow-up studies were designed sequentially after earlier results were available. 
\paragraph{Error control and interval levels.} The terminal-regret and fault-utility comparisons and the two timing interactions form a family of four tests with Bonferroni correction. Each test uses $\alpha=.0125$ and 98.75\% confidence intervals (CIs). The medium-versus-low and high-versus-medium comparisons use two tests, each with $\alpha=.025$ and 97.5\% CIs. The earlier effort and prior-sensitivity panels share a family of four contrasts with 98.75\% CIs. The counterfactual concurrent-request contrast and larger live utility test each use a separate
$\alpha=.05$ family (95\% CIs). The prospective replay has one two-sided test at $\alpha=.05$. Its secondary results are descriptive. The study of configuration under load has its own single-test family at $\alpha=.05$,
using a paired $t$ test across 12 allocations (separate server reservations). Sonnet uses two tests, each at $\alpha=.025$, with 97.5\% CIs. We compare raw $p$ values with .025.
Concurrent-dispatch latency equivalence requires both modes' 90\% intervals for the geometric-mean latency ratio to
fall within the prespecified margin $[1/1.10,1.10]$.

\paragraph{Terminal regret and fixed-model fault utility.} The family has four Bonferroni tests at $\alpha=.0125$. For terminal regret, we draw bootstrap samples of instances within each task family. For fault utility, we permute mode labels within cases and draw bootstrap samples of repeated responses, keeping case probabilities fixed.

\paragraph{Effort above low.} For the two tests at $\alpha=.025$, we draw 10{,}000 bootstrap samples of paired instances within task families.

\paragraph{Evidence-onset interaction.} The primary contrast is
  (thinking $-$ fast at 60~s) $-$ (thinking $-$ fast at 2.1~s), with a 98.75\% CI from a 10{,}000-draw instance bootstrap.

\paragraph{Dispatch-clock interaction.} The fourth Bonferroni test ($\alpha=.0125$) uses the same primary interaction contrast as the evidence-onset protocol. We bootstrap instances within each of six groups, using 10{,}000 draws.

\paragraph{Earlier and local analyses.} The initial effort sweep and prior-sensitivity panel share one four-contrast Bonferroni family (two contrasts per panel). For the initial effort sweep, the percentile CI and the prespecified centered-bootstrap test need not agree. The prespecified test determines whether to reject the null hypothesis. The local stopping evaluation uses 10,000 bootstrap draws of episodes within each task family and a prespecified hierarchy of primary tests. Under that hierarchy, the comparison between modes is descriptive. The fault challenge uses 4,096 bootstrap resamples of paired cases and Holm correction over 12 regret contrasts at $\alpha=.05$ (95\% CIs). Its deadline contrasts are descriptive. The deadline panels with serial or shared serving and the output-constraint panels are descriptive by design.

\paragraph{Study sequence.} We designed the studies sequentially, choosing later questions, experimental settings and test families after earlier results were available. Appendix~\ref{app:sequence} gives the full sequence and its timing. Error control within each study does not account for the selection of studies across the sequence.

\subsection{Returned decision quality}
\label{app:quality}
This study tests whether low-effort thinking returns better terminal stopping decisions than fast mode and
whether higher effort helps further. Appendix~\ref{app:hosted} describes the design. Each of the 1{,}200 new terminal instances received one response per mode, scored by exact terminal decision regret.

\begin{table}[H]
\centering
\caption{Prespecified Haiku 5.5 comparisons (first mode minus second). Stopping rows give regret, which ignores timing (lower is better). The fault row gives expected utility across deadlines with fallback and 5~s inspection (higher is better).}
\label{tab:confirm-supp}
\small
\setlength{\tabcolsep}{3pt}
\begin{tabular}{@{}>{\raggedright\arraybackslash}p{0.3\textwidth}llrrl@{}}
\toprule
cases & contrast & effect & CI & adjusted $p$ & result \\
\midrule
stopping: 1,200 new, unselected instances & low $-$ fast & $-0.210$ & $[-0.242,-0.180]$ & .002 & detected \\
faults: fixed 128-case model, 3 responses each & low $-$ fast & $+3.65$ & $[2.43, 4.86]$ & .002 & detected \\
stopping: the same 1,200 instances & medium $-$ low$^\dagger$ & $-0.004$ & $[-0.017, 0.008]$ & .88 & not detected \\
stopping: the same 1,200 instances & high $-$ medium & $+0.006$ & $[-0.007, 0.020]$ & .60 & not detected \\
\bottomrule
\end{tabular}
\par
\parbox{\textwidth}{\footnotesize All rows are confirmatory. Fast disables thinking. Low, medium and high are thinking effort levels. We compare Bonferroni-adjusted $p$ values with .05. The raw thresholds are .0125 for the four-test family covering terminal decision regret, fault utility and the two timing interactions, and .025 for the two-test higher-effort family. The corresponding percentile bootstrap CIs are 98.75\% and 97.5\%, respectively. Fault responses are repeated for a fixed 128-case model, so inference applies to that model and not to new scenarios. $^\dagger$ We collected medium responses about 13~h after the paired low-effort responses, on the same day. Any change in the provider's service between the two collection periods cannot be separated from the effort effect.
Earlier contrasts and their qualifications are in Table~\ref{tab:historical}.}
\end{table}

\paragraph{Primary results and supporting statistics.} Section~4 reports the primary results. Table~\ref{tab:confirm-supp} lists all four prespecified Haiku comparisons with corrections for their test families, including the fault-utility test reported in Appendix~\ref{app:faults}. All 3{,}168 scored terminal and fault calls returned
without failure, refusal, truncation or malformed output. Mean cost per decision was \$0.00031 in fast mode, \$0.00055
at low effort and \$0.00075 at high effort. The 99th-percentile provider-call latency was 8.6~s at low, 9.9~s at medium and 14.2~s at high effort. The share of answers arriving within 10~s fell from 99.8\% at low effort to 91.1\% at high effort.

\paragraph{Regret by returned action (post hoc, descriptive).} Most fast-mode regret came from continuing when accepting was optimal (0.189 of 0.268) or accepting when rejecting was optimal (0.078) (Table~\ref{tab:decomp}). Thinking reduced regret from continuing when accepting was optimal to 0.043. Thinking returned the optimal final action in 668 of 1{,}200 instances, compared with 74 for fast mode. The prompt offered continuation without labeling the decision as final, so the model had to infer from the action durations and cutoff that no further work fit. Continuing cost at most 0.0031 more than rejecting, so its regret mainly reflects lost acceptance value.

\begin{table}[H]
\centering
\caption{Terminal decision regret by returned action (post hoc, descriptive; counts in parentheses). Components are means per decision and sum to the total before rounding for display. ``Optimal'' refers to the exact Bayes controller.}
\label{tab:decomp}
\small
\setlength{\tabcolsep}{3pt}
\begin{tabular}{lrr}
\toprule
component & fast & low \\
\midrule
continued; accepting was optimal & 0.189 (265) & 0.043 (56) \\
accepted; rejecting was optimal & 0.078 (94) & 0.005 (13) \\
rejected; accepting was optimal & 0.000 (0) & 0.009 (22) \\
continued; rejecting was optimal & $<$0.001 (767) & $<$0.001 (441) \\
\midrule
total regret & 0.268 & 0.057 \\
optimal terminal action returned & 74 & 668 \\
\bottomrule
\end{tabular}
\end{table}

\paragraph{Earlier effort studies.} \begin{table}[H]
\centering
\caption{Earlier effort and prior-sensitivity contrasts for Haiku~5.5. All rows are confirmatory
within the earlier four-contrast effort/prior-sensitivity family. Stopping rows give exact terminal decision regret (negative is better). Fault rows give deadline-adjusted utility with 5~s inspection (positive is better). All displayed $p$ values are Bonferroni-adjusted and compared with .05, equivalent to comparing raw $p$ values with per-test $\alpha=.0125$. Percentile bootstrap CIs are 98.75\%. Fault rows report repeated responses on the fixed 128-case model, so inference applies to that model and not to new scenarios.}
\label{tab:historical}
\small
\setlength{\tabcolsep}{3pt}
\resizebox{\textwidth}{!}{%
\begin{tabular}{lllrrrl}
\toprule
study & task (instances) & contrast & effect & CI & Bonferroni $p$ & result \\
\midrule
initial effort sweep & fault (128 cases $\times$ 3) & low $-$ fast & $+3.39$ & $[3.28, 3.49]$ & .002 & detected \\
initial effort sweep & stopping (36 terminal $\times$ 3) & low $-$ fast & $-0.128$ & $[-0.285,-0.009]$ & .070 & not established \\
prior-sensitivity panel & fault, no-fault prior .99 & medium $-$ fast & $+4.48$ & $[3.63, 6.12]$ & .002 & detected \\
prior-sensitivity panel & stopping (240 new, unselected) & medium $-$ fast & $-0.171$ & $[-0.233,-0.109]$ & .002 & detected \\
\bottomrule
\end{tabular}}
\par\smallskip\footnotesize
For the initial stopping sweep, the percentile CI and prespecified centered-bootstrap test use different procedures and need not agree. The prespecified test determines whether to reject the null hypothesis and did not reject it.
\end{table}

The fault rows of Table~\ref{tab:historical} belong to the same four-contrast family as the stopping rows and are
discussed with the other fault results in Appendix~\ref{app:faults}.

\paragraph{Sonnet results.} Both Sonnet comparisons use 97.5\% CIs and raw $p$ values compared with per-test $\alpha=.025$ in the separate two-test Sonnet family. Neither prespecified contrast was detected. Both intervals include zero and effects of a few hundredths. The terminal-regret difference, low-effort thinking minus fast, was $-0.0143$ (97.5\% CI $[-0.0301,0.0012]$, raw $p=.041$). Regret was 0.062 for fast mode and 0.047 for thinking. On the dispatch clock, thinking's realized-utility advantage grew by $+0.0150$ from 2.1 to 60~s (interval $[-0.0133,0.0450]$, raw $p=.27$). The advantage was $-0.010$ at 2.1~s and $+0.005$ at 60~s. At 2.1~s, 92.7\% of fast answers and 87.2\% of thinking answers arrived in time.

\subsection{Deadline effects on executed value}
\label{app:deadline}
These studies measure how much of thinking's improvement in returned decisions survives deadline enforcement.
Both timing protocols present new terminal instances with only accept and reject available and score the executed action in the simulator. The dispatch clock uses 600 instances at 2.1, 4, 6, 10 and 60~s. The evidence-onset clock uses 600 instances at 2.1, 10 and 60~s. Each condition receives a new response, so answer changes cannot be separated from latency or deadline
enforcement.

\paragraph{Primary results.} On the dispatch clock, thinking's realized-utility advantage grew by 0.118 from 2.1 to 60~s (98.75\% CI $[0.065,0.172]$, Bonferroni-adjusted $p=.0004$). On the evidence-onset clock, thinking's realized-utility advantage grew from zero to 0.143 (98.75\% CI $[0.103,0.183]$, Bonferroni-adjusted $p=.0004$). Preparation and pacing caused both modes to use the fallback in at least 96.5\% of cases at 2.1~s on this clock. The interaction therefore numerically equals the advantage at 60~s. Results at individual deadlines are descriptive.

\paragraph{Dispatch collection.} The dispatch-clock study made 6{,}140 calls, including preliminary-check and pilot calls, at a total cost of \$1.81. All calls completed without failure, and 6{,}000 were scored. Median latency was 0.88~s for fast mode and 2.06~s for thinking. The corresponding 99th-percentile latencies were 2.41~s and 6.99~s. Thinking minus fast realized utility was $+0.005$, $+0.062$, $+0.105$, $+0.098$
and $+0.123$ at 2.1, 4, 6, 10 and 60~s (descriptive).

\paragraph{Fixed-answer sweep (post hoc, descriptive).} For the dispatch-clock evaluation's 600 instances, we evaluated both modes' answers generated with the 60~s prompt against deadlines of 2.1, 4, 6, 10 and 60~s. We used the recorded dispatch-clock delays and executed each answer if it was valid and arrived by the deadline. Otherwise, we executed reject. For these fixed answers, thinking's realized-utility advantage over fast mode was $-0.008$, $+0.073$, $+0.113$, $+0.123$ and $+0.123$ at the five deadlines, respectively. The advantage grew by $+0.132$ from 2.1 to 60~s (95\% CI $[0.105,0.160]$, 2{,}000-draw instance bootstrap within groups). This descriptive replay is conditional on the recorded answers and delays. Answers might differ if the prompt stated a different deadline.

\paragraph{Timing from evidence onset.}
\label{app:onset}
The evidence-onset clock runs from evidence onset to receipt and includes input-token counting and the client's
serial pacing and admission delay. The pacing and admission delay combines two waits that keep successive requests at least 6~s apart, one before input-token counting and one just before the provider call. Across scored calls, the median delay was 4.6~s in both modes. Across all 1{,}800 scored calls per mode, 4.4\% of fast answers and 1.7\% of thinking answers arrived within 2.1~s on the evidence-onset clock. In the 2.1~s condition, 21 fast answers and 7 thinking answers out of 600 arrived in time, leaving 579 and 593 late, respectively. Measured from the start of the provider call to its finish, 99.3\% of fast answers and 43.9\% of thinking answers would have arrived within 2.1~s. Adding only the second wait, just before the provider call, gives 99.1\% and 42.8\%. At 2.1~s, both modes therefore mainly use the fallback because of pacing, leaving utility at a floor. The prespecified interaction numerically equals the advantage at 60~s. The confirmatory result is that thinking's advantage is larger at 60~s than at 2.1~s. The equal mean executed utilities at 2.1~s are descriptive. The analysis does not attribute the 2.1~s result to thinking latency. Evidence-onset latency quantiles are not comparable with the provider-call latencies
reported for the terminal-regret and effort comparisons. The dispatch-clock protocol starts the dispatch clock at the provider call. Under that protocol, 98.8\% of fast answers and 54.5\% of thinking answers arrived within 2.1~s. The prespecified interaction was $+0.118$ with $\Delta J(2.1)=+0.005$.

\paragraph{Evidence-onset answers (descriptive).} The realized-utility difference, thinking minus fast, was 0.000, $+0.085$ and $+0.143$ at 2.1, 10 and 60~s. At 10~s, 4.3\% of thinking answers were too late to execute. Answer regret at 2.1, 10 and 60~s was
0.069, 0.079 and 0.070 for thinking and 0.218, 0.186 and 0.222 for fast mode. Accept was optimal in 170 of 600 instances. Fast mode generated 61 to 72 accepts, and thinking generated 122 to 125. At 60~s, 78 instances completed the goal only
with thinking and 3 only with fast mode.

\begin{table}[H]
\centering
\caption{Realized outcomes after execution with timing from evidence onset (600 instances per condition). Low denotes low-effort thinking. Fallback is the share of answers that arrived late and were replaced by rejection. Occupancy is resource-seconds per decision. Means for individual conditions are descriptive. Only the change in thinking's advantage between 2.1 and 60~s is confirmatory (Bonferroni-adjusted $p=.0004$). Fast mode has lower utility at 60~s than at 10~s because it accepts fewer samples (61 compared with 72). Neither condition uses the fallback.}
\label{tab:onset}
\small
\setlength{\tabcolsep}{3pt}
\begin{tabular}{llrrrr}
\toprule
deadline & mode & utility & goal completion & fallback & occupancy \\
\midrule
2.1 s & fast & 0.0016 & 0.3\% & 96.5\% & 12.07 \\
2.1 s & low & 0.0016 & 0.2\% & 98.8\% & 12.09 \\
10 s & fast & 0.070 & 9.5\% & 0.0\% & 15.15 \\
10 s & low & 0.155 & 16.7\% & 4.3\% & 16.64 \\
60 s & fast & 0.032 & 6.7\% & 0.0\% & 15.16 \\
60 s & low & 0.175 & 19.2\% & 0.0\% & 16.66 \\
\bottomrule
\end{tabular}
\end{table}

\begin{figure}[H]
\centering
\includegraphics[width=0.65\textwidth]{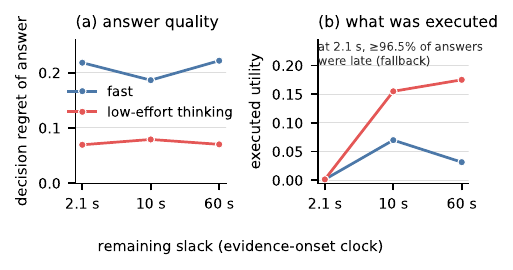}
\caption{Answer quality and execution for Claude Haiku 5.5 timed from evidence onset (600 new instances per condition). (a) Exact decision regret of the generated answers, including answers too late to execute. Lower is better, and the results are descriptive. (b) Mean realized utility of the executed action after deadline enforcement and fallback. Utility is $+1$ for accepting a good sample, $-1$ for a false accept and $0$ for a reject, each minus a small penalty for sample removal. Only the change in thinking's advantage between 2.1 and 60~s ($+0.143$, 98.75\% CI $[0.103,0.183]$) is confirmatory.}
\label{fig:onset-answers}
\end{figure}

\subsection{Prediction error from separate summaries}
\label{app:dependence}

We explored existing dispatch-clock records before designing this prospective confirmatory study on new instances. The exploratory estimates are not confirmatory evidence.
All values in the expected-value analyses below are posterior expected utility gains over rejection, $v=Q(a)-Q(\mathrm{reject})$, computed from the exact action values. We report simulator-realized utility separately.
The separate-summary estimate $G_{\mathrm{sep}}$ multiplies the proportion of answers that are valid and timely, $q$, by the mean value of valid answers. The joint estimate $G_{\mathrm{joint}}$ assigns each answer its value if it was valid and timely, and zero otherwise, then averages across answers. Their difference, $B=G_{\mathrm{sep}}-G_{\mathrm{joint}}$, measures overprediction. Positive $B$ means separate summaries predict too much value. The primary contrast $D$ is thinking's $B$ minus fast mode's $B$.
Within each group, $G_{\mathrm{sep}}=q\overline v_{\mathrm{valid}}$ and $G_{\mathrm{joint}}=\overline{v\mathbb{I}\{\mathrm{valid},L\le s\}}$. We average the group estimates with fixed equal weights.
When pooling, we compute separate summaries over the whole cohort. The pooled estimate need not equal the average of the group estimates.
The separate-summary estimate omits the association between value and timeliness. We measure the resulting error in this setting. We do not assume that serving systems generally use this approximation.

\paragraph{Design and study plan.}
The confirmatory study used 1{,}200 new terminal instances in two successive blocks of 600. Each block contained 100 instances per family and stage, giving 12 family, stage and block groups in all.
The scored seeds were 981014 to 982213. The preceding preliminary checks and technical pilot used separate instances and were excluded from every estimate.
For each instance, we obtained one response in Haiku~5.5 fast mode and one in low-effort thinking mode. We used the same accept-or-reject 60~s prompt and request settings as the dispatch-clock evaluation.
We shuffled instances within each block and placed each instance's two modes next to each other in randomized order. We finished the first block before starting the second.
The simulator ran at the prompted deadline. We replayed shorter deadlines using the recorded latency from dispatch to receipt, excluding counting, pacing and admission.
The replay cannot say how the model would answer if told the deadline was 2.1~s.
The protocol assigned failures a reject fallback with zero value and allowed no retries. No failures occurred.
The confirmatory test required every planned scored response. We made no outcome comparisons during collection and did not extend collection based on outcomes.
We made 2{,}428 attempts, comprising 4 preliminary-check calls, 24 pilot calls and 2{,}400 scored calls. All calls used the dispatch clock and cost \$0.722 in total.

Appendix~\ref{app:deviations} records amendments to the replay study plan.

\paragraph{Primary contrast and uncertainty.}
We used a single two-sided confirmatory test in a separate testing family at $\alpha=.05$. The test evaluated $D=B_{\mathrm{think}}(2.1)-B_{\mathrm{fast}}(2.1)$ with equal weights over the 12 groups.
For 10{,}000 paired-bootstrap draws (seed 983777), we resampled both modes of each instance together within groups, preserving group sizes. Intervals are percentile 95\% CIs.
The approximate two-sided $p$ is $\min(1,2\min(P^*(D^*\le0),P^*(D^*\ge0)))$, with ties counted in both tails. Just before summarization, we set statistics within the prespecified tolerance of zero ($|x|<10^{-12}$, absorbing floating-point ties) to zero.
At 2.1~s, separate summaries overestimated thinking's advantage over fast mode by $D=+0.0241$ posterior expected utility units, averaged equally over groups. The result is confirmatory (95\% CI $[0.0165,0.0317]$, approximate bootstrap $p<0.0001$). There were zero crossings in the 10{,}000 bootstrap draws. The displayed $p$ therefore reflects that Monte Carlo resolution and is not a rigorous bound on the tail probability. The positive estimate agreed with the exploratory result. The point estimate and most of the interval lie below the 0.03 planning target, which was not a success criterion.
All remaining contrasts and diagnostics are descriptive, with unadjusted intervals. The block contrast was not designed to test stability across blocks, and ranking was not a success requirement.

\begin{table}[H]
\centering\footnotesize
\caption{Overprediction of thinking's advantage over fast mode ($D$), in posterior expected utility units. Entries give estimates [95\% CI]. Positive values mean separate summaries overstate thinking's advantage. Only the within-group result at 2.1~s is confirmatory. Pooled results and results at other deadlines are descriptive.}\label{tab:dep-deadlines}
\setlength{\tabcolsep}{3pt}
\begin{tabular}{@{}lrr@{}}
\toprule
deadline (s) & within groups & pooled \\
\midrule
2.1 & $0.0241\;[0.0165,0.0317]$ & $0.0498\;[0.0397,0.0598]$ \\
4.0 & $0.0531\;[0.0425,0.0628]$ & $0.0668\;[0.0550,0.0787]$ \\
6.0 & $0.0166\;[0.0081,0.0250]$ & $0.0207\;[0.0118,0.0299]$ \\
10.0 & $0.0002\;[-0.0011,0.0020]$ & $0.0003\;[-0.0008,0.0019]$ \\
60.0 & $0.0000\;[0.0000,0.0000]$ & $0.0000\;[0.0000,0.0000]$ \\
\bottomrule
\end{tabular}
\end{table}

\begin{table}[H]
\centering\footnotesize
\caption{Overprediction of thinking's advantage ($D$) in each collection block and the difference between blocks (block 2 minus block 1). Estimates and intervals are descriptive and unadjusted. The block contrast was not designed to test stability.}\label{tab:dep-blocks}
\setlength{\tabcolsep}{3pt}
\begin{tabular}{@{}lrrrr@{}}
\toprule
deadline (s) & block 1 [95\% CI] & block 2 [95\% CI] & difference [95\% CI] & approx. $p$ \\
\midrule
2.1 & $0.0224\;[0.0112,0.0335]$ & $0.0258\;[0.0151,0.0365]$ & $0.0034\;[-0.0124,0.0190]$ & 0.667 \\
4.0 & $0.0536\;[0.0383,0.0681]$ & $0.0525\;[0.0382,0.0662]$ & $-0.0011\;[-0.0215,0.0198]$ & 0.910 \\
6.0 & $0.0146\;[0.0037,0.0260]$ & $0.0186\;[0.0062,0.0313]$ & $0.0041\;[-0.0126,0.0209]$ & 0.640 \\
10.0 & $-0.0003\;[-0.0011,0.0000]$ & $0.0007\;[-0.0019,0.0043]$ & $0.0010\;[-0.0017,0.0046]$ & 0.670 \\
60.0 & $0.0000\;[0.0000,0.0000]$ & $0.0000\;[0.0000,0.0000]$ & $0.0000\;[0.0000,0.0000]$ & 1.000 \\
\bottomrule
\end{tabular}
\end{table}
\paragraph{Expected-value summaries and secondary ranking.}

The joint estimate of thinking's advantage is the same when pooled or averaged over equally sized groups. Estimates from separate summaries need not be the same.
At 2.1~s, keeping each answer's value paired with its arrival time favored fast mode within groups. Thinking minus fast was $-0.025$ ($[-0.043,-0.008]$). Separate summaries gave $-0.001$, rating the modes nearly equal.
The pooled estimate from separate summaries was $+0.0244$, whereas the joint estimate was negative. This secondary ranking reversal was not the confirmatory endpoint.

{\footnotesize
\setlength{\tabcolsep}{3pt}
\begin{longtable}{@{}lrrrr@{}}
\caption{Posterior expected gain over rejection, averaged over groups. Entries give estimates [95\% CI]. $G_{\mathrm{sep}}$ is the gain predicted from separate summaries. $G_{\mathrm{joint}}$ is the gain computed with each answer's value and arrival time kept paired. $B=G_{\mathrm{sep}}-G_{\mathrm{joint}}$ is the overprediction. We use posterior expected values without simulator outcomes. Without grouping, the prediction for thinking at 2.1~s is 0.0726. All mode-level entries are descriptive.}\label{tab:dep-G}\\
\toprule
mode & deadline (s) & $G_{\mathrm{sep}}$ & $G_{\mathrm{joint}}$ & $B$ \\
\midrule\endfirsthead
\toprule
mode & deadline (s) & $G_{\mathrm{sep}}$ & $G_{\mathrm{joint}}$ & $B$ \\
\midrule\endhead
fast & 2.1 & $0.0483\;[0.0327,0.0640]$ & $0.0485\;[0.0324,0.0644]$ & $-0.0001\;[-0.0021,0.0024]$ \\
fast & 4.0 & $0.0494\;[0.0336,0.0654]$ & $0.0493\;[0.0331,0.0654]$ & $0.0001\;[-0.0011,0.0020]$ \\
fast & 6.0 & $0.0500\;[0.0339,0.0662]$ & $0.0501\;[0.0339,0.0663]$ & $-0.0001\;[-0.0003,0.0000]$ \\
fast & 10.0 & $0.0501\;[0.0339,0.0663]$ & $0.0501\;[0.0339,0.0663]$ & $0.0000\;[0.0000,0.0000]$ \\
fast & 60.0 & $0.0501\;[0.0339,0.0663]$ & $0.0501\;[0.0339,0.0663]$ & $0.0000\;[0.0000,0.0000]$ \\
low & 2.1 & $0.0470\;[0.0371,0.0569]$ & $0.0230\;[0.0139,0.0325]$ & $0.0240\;[0.0166,0.0313]$ \\
low & 4.0 & $0.0994\;[0.0858,0.1124]$ & $0.0461\;[0.0335,0.0587]$ & $0.0532\;[0.0427,0.0629]$ \\
low & 6.0 & $0.1399\;[0.1218,0.1581]$ & $0.1234\;[0.1049,0.1424]$ & $0.0165\;[0.0080,0.0248]$ \\
low & 10.0 & $0.1599\;[0.1391,0.1809]$ & $0.1597\;[0.1390,0.1806]$ & $0.0002\;[-0.0011,0.0020]$ \\
low & 60.0 & $0.1605\;[0.1397,0.1817]$ & $0.1605\;[0.1397,0.1817]$ & $0.0000\;[0.0000,0.0000]$ \\
\bottomrule
\end{longtable}
}

\begin{table}[H]
\centering\footnotesize
\caption{Difference in posterior expected gain over rejection, thinking minus fast. Entries give estimates [95\% CI]. Positive values favor thinking. ``Separate'' uses separate value and timing summaries. ``Joint'' keeps each answer's value and arrival time paired. Ranking comparisons are secondary and descriptive. The joint column also equals the pooled joint contrast. The signs-agree column compares the signs of the within-group separate and joint estimates.}\label{tab:dep-ranking}
\setlength{\tabcolsep}{3pt}
\begin{tabular}{@{}lrrrr@{}}
\toprule
deadline (s) & within-group separate & joint & pooled separate & signs agree \\
\midrule
2.1 & $-0.0013\;[-0.0186,0.0158]$ & $-0.0254\;[-0.0433,-0.0076]$ & $0.0244\;[0.0104,0.0387]$ & yes \\
4.0 & $0.0499\;[0.0319,0.0670]$ & $-0.0032\;[-0.0215,0.0151]$ & $0.0636\;[0.0476,0.0795]$ & no \\
6.0 & $0.0899\;[0.0714,0.1087]$ & $0.0733\;[0.0536,0.0932]$ & $0.0940\;[0.0758,0.1126]$ & yes \\
10.0 & $0.1097\;[0.0903,0.1293]$ & $0.1096\;[0.0902,0.1291]$ & $0.1099\;[0.0904,0.1295]$ & yes \\
60.0 & $0.1104\;[0.0909,0.1301]$ & $0.1104\;[0.0909,0.1301]$ & $0.1104\;[0.0909,0.1301]$ & yes \\
\bottomrule
\end{tabular}
\end{table}
\paragraph{Simulator-realized replay.}
We replayed each recorded answer in a fresh simulator run. For timely valid answers, we executed the recorded action at the recorded time. For other records, we executed reject at $\min(L,s)$. We found zero mismatches with the fixed-answer identity, which gives the recorded realized utility if timely and $-0.0001$ otherwise.
Replay agreement does not validate the simulator's physical model.

\begin{table}[H]
\centering\footnotesize
\caption{Simulator-realized fixed-answer utility $J$ and the thinking-minus-fast difference $\Delta J$. Entries give descriptive estimates [95\% CI], distinct from posterior expected value.}\label{tab:dep-realized}
\setlength{\tabcolsep}{3pt}
\begin{tabular}{@{}lrrr@{}}
\toprule
deadline (s) & fast $J$ & low $J$ & $\Delta J$ \\
\midrule
2.1 & $0.0449\;[0.0282,0.0616]$ & $0.0241\;[0.0132,0.0349]$ & $-0.0208\;[-0.0400,-0.0017]$ \\
4.0 & $0.0457\;[0.0291,0.0624]$ & $0.0457\;[0.0324,0.0599]$ & $0.0000\;[-0.0192,0.0192]$ \\
6.0 & $0.0466\;[0.0299,0.0632]$ & $0.1232\;[0.1032,0.1432]$ & $0.0767\;[0.0558,0.0983]$ \\
10.0 & $0.0466\;[0.0299,0.0632]$ & $0.1582\;[0.1357,0.1807]$ & $0.1117\;[0.0908,0.1333]$ \\
60.0 & $0.0466\;[0.0299,0.0632]$ & $0.1591\;[0.1357,0.1816]$ & $0.1125\;[0.0908,0.1342]$ \\
\bottomrule
\end{tabular}
\end{table}

\paragraph{Prediction on new instances.}
We fixed separate and joint predictors from the exploratory study before collection and tested them on the new instances. For predicting thinking's advantage over fast mode on the new instances, the joint predictor had the smaller absolute error at the three shortest deadlines (0.0149 against 0.0523 at 2.1~s) and the same error at 10 and 60~s. This comparison is descriptive.

\paragraph{Response and token diagnostics.}
The diagnostics in Table~\ref{tab:dep-counts} are observational, so they cannot show whether thinking tokens or difficulty cause the differences.
At the primary deadline, 543 of 1{,}200 low-effort thinking answers were timely. Of these timely answers, 96.1\% reported zero thinking tokens. Late answers had a median of 225 thinking tokens. Pooling all instances, mean value was 0.0509 for timely answers and 0.2511 for late answers. Of 231 accepts, 45 were timely.
All timing began at dispatch. Latency had a mean of 1.11~s and a median of 0.98~s for fast mode, and a mean of 2.96~s and a median of 2.50~s for thinking. Thinking answers used a mean of 160 thinking tokens and a median of 85.

\begin{table}[H]
\centering\footnotesize
\caption{Counts and posterior expected values by mode and deadline. Accept counts are shown as timely/late. Every answer was valid. Mean value over all answers is 0.0501 (95\% CI $[0.0339,0.0663]$) for fast mode and 0.1605 ($[0.1397,0.1817]$) for thinking, at every deadline. Mean values are computed within each indicated arrival group. Empty groups are marked n/a.}\label{tab:dep-counts}
\setlength{\tabcolsep}{3pt}
\begin{tabular}{@{}lrrrrrrr@{}}
\toprule
mode & deadline (s) & timely & late & invalid & accepts, timely/late & mean $v$, timely & mean $v$, late \\
\midrule
fast & 2.1 & 1155 & 45 & 0 & 110/2 & 0.0504 & 0.0444 \\
fast & 4.0 & 1183 & 17 & 0 & 111/1 & 0.0500 & 0.0588 \\
fast & 6.0 & 1196 & 4 & 0 & 112/0 & 0.0503 & 0.0000 \\
fast & 10.0 & 1200 & 0 & 0 & 112/0 & 0.0501 & n/a \\
fast & 60.0 & 1200 & 0 & 0 & 112/0 & 0.0501 & n/a \\
low & 2.1 & 543 & 657 & 0 & 45/186 & 0.0509 & 0.2511 \\
low & 4.0 & 845 & 355 & 0 & 75/156 & 0.0655 & 0.3866 \\
low & 6.0 & 1076 & 124 & 0 & 174/57 & 0.1376 & 0.3592 \\
low & 10.0 & 1196 & 4 & 0 & 230/1 & 0.1602 & 0.2500 \\
low & 60.0 & 1200 & 0 & 0 & 231/0 & 0.1605 & n/a \\
\bottomrule
\end{tabular}
\end{table}

\paragraph{Post hoc exploratory origin.}
We analyzed the dispatch-clock study's existing records for 600 scored instances, holding each mode's answers to the 60~s prompt fixed. All answers were valid. We used 10{,}000 bootstrap draws, resampling paired instances within family and stage groups (seed 20261019). All estimates and intervals are exploratory. At 2.1~s, $q$ was 0.988 for fast mode and 0.518 for thinking. Pooled separate summaries gave thinking minus fast as $+0.0525$, while the joint estimate gave $-0.0105$. Overprediction $D$ within family and stage groups was $+0.0374$ ($[0.026,0.048]$) at 2.1~s and $+0.0276$ ($[0.014,0.041]$) at 4~s.

Of the low-effort thinking mode's 100 accepts, 13 were timely at 2.1~s. Overall, 311 answers were timely, including 276 with zero reported thinking tokens. Late answers had a median of 226 thinking tokens.
Mean value was 0.025 for timely answers and 0.277 for late answers.
The simulator-realized fixed-answer replay gave $\Delta J(2.1)=-0.0083$ ($[-0.032,0.015]$), with the same timely shares. This realized-utility contrast is distinct from the expected-value joint contrast $-0.0105$.

\paragraph{Post hoc Sonnet replay.}
A descriptive analysis of existing Sonnet dispatch-clock records used fixed answers to the 60~s prompt and 4{,}000 bootstrap draws.
The Sonnet baseline used low effort and permitted no thinking before the answer. This setting differs from Haiku's disabled thinking.
At 2.1~s, $G_{\mathrm{sep}}/G_{\mathrm{joint}}$ was 0.2155/0.2143 for that baseline and 0.2123/0.2022 for low-effort thinking.
Within family and stage groups, overprediction for low-effort thinking minus overprediction for the baseline was $+0.0082$ ($[-0.0085,0.0250]$) at 2.1~s and $+0.0158$ ($[0.0042,0.0287]$) at 4~s.
Of 600 low-effort thinking answers, 515 were timely at 2.1~s. Of these timely answers, 514 had zero thinking tokens. Late answers had a median of 160 thinking tokens.
Mean value was 0.236 for timely and 0.318 for late answers.
These descriptive estimates are consistent with a weaker association between value and timeliness when fast answers already have high value. We did not test for a difference between models, and the estimates are observational.

Appendix~\ref{app:verification} reports a separate computation check on these records.

\subsection{Follow-up checks of the replay}
\label{app:followup}

This section reports four checks made after the main analysis of the replay, none of them in the original study plan.
We fixed the replication's design and test before collecting its data, so its test of $D$ at 2.1~s is confirmatory. All other results here are descriptive.

\paragraph{Replication on new instances.} We repeated the replay on 300 new instances, 50 for each task family and
observation stage, with seeds not used in any earlier study. The prompts, modes, 60~s limit, random mode order and
analysis code were unchanged. We sent requests one at a time, spaced 3~s apart, half the original spacing, so the run took 35 minutes. All 600 scored responses were valid, and the run cost \$0.19. The replication reproduced the primary contrast (Table~\ref{tab:replication}), with $D=+0.0191$ (95\% CI $[0.0039,0.0341]$, approximate $p=.015$) compared with $+0.0241$ in the original run. Late thinking answers again had higher expected gains than on-time answers. The ranking of the two modes did not replicate. Fast mode gained little on these instances, so the joint estimate favored thinking, with an interval
that includes zero.

\begin{table}[H]
\centering
\caption{The replay at 2.1~s in the original run and in the replication (posterior expected gain over rejecting).
On-time and late gains are pooled means over answers. The estimates of $B$, $D$ and the joint difference are averaged equally over groups of task family and stage (and block in the original run). Only $D$ at 2.1~s within groups is confirmatory in each run. Intervals are
95\% bootstrap intervals over instances.}
\label{tab:replication}
\small
\setlength{\tabcolsep}{4pt}
\begin{tabular}{@{}lrr@{}}
\toprule
quantity at 2.1~s & \hd{original\\(1{,}200 instances)} & \hd{replication\\(300 instances)} \\
\midrule
share of thinking answers on time & 0.4525 & 0.4367 \\
mean gain of on-time thinking answers & 0.0509 & 0.0657 \\
mean gain of late thinking answers & 0.2511 & 0.2064 \\
on-time thinking answers with no thinking tokens & 96.1\% & 92.4\% \\
overprediction $B$, fast mode & $-0.0001$ & $-0.0027$ \\
overprediction $B$, thinking & $+0.0240$ & $+0.0164$ \\
$D$, thinking minus fast & $+0.0241\;[0.0165,0.0317]$ & $+0.0191\;[0.0039,0.0341]$ \\
$D$ without grouping & $+0.0498\;[0.0397,0.0598]$ & $+0.0380\;[0.0192,0.0570]$ \\
joint estimate, thinking minus fast & $-0.0254\;[-0.0433,-0.0076]$ & $+0.0245\;[-0.0099,0.0589]$ \\
\bottomrule
\end{tabular}
\end{table}

\paragraph{Changing the fallback.} In the replay, a late answer led to rejecting the sample, which is the safe
choice. We scored the same recorded answers again as if a late answer led to accepting the sample. The prompt never stated which action follows a late answer, so rescoring changed the fallback without changing the recorded answers. With acceptance as the fallback, an
answer adds value only by rejecting a sample that is likely bad. Separate summaries then understated thinking's advantage at 4 and 6~s in both runs (Table~\ref{tab:fallback}). At 2.1~s, the runs disagreed, with $D$ near zero in the original run and positive in the replication. Using acceptance as the fallback also made late answers far more costly. At 2.1~s, the difference in delivered value, thinking minus fast mode, fell from $-0.025$ to $-0.222$ in the original run and from $+0.025$ to $-0.265$ in the replication. The direction and size of the error therefore depend on the fallback action.

\begin{table}[H]
\centering
\caption{Overprediction of thinking's advantage over fast mode ($D$) when a late answer leads to rejecting or to
accepting the sample, with 95\% CIs. Each run uses the same recorded answers under both fallbacks.}
\label{tab:fallback}
\small
\setlength{\tabcolsep}{3pt}
\begin{tabular}{@{}lcccc@{}}
\toprule
& \multicolumn{2}{c}{original run} & \multicolumn{2}{c}{replication} \\
deadline & reject & accept & reject & accept \\
\midrule
2.1~s & \hd{$+0.0241$\\$[0.0165,0.0317]$} & \hd{$-0.0004$\\$[-0.0162,0.0159]$} & \hd{$+0.0191$\\$[0.0039,0.0341]$} & \hd{$+0.0320$\\$[0.0024,0.0625]$} \\
\addlinespace[2pt]
4~s & \hd{$+0.0531$\\$[0.0425,0.0628]$} & \hd{$-0.0588$\\$[-0.0719,-0.0443]$} & \hd{$+0.0411$\\$[0.0217,0.0591]$} & \hd{$-0.0387$\\$[-0.0584,-0.0170]$} \\
\addlinespace[2pt]
6~s & \hd{$+0.0166$\\$[0.0081,0.0250]$} & \hd{$-0.0272$\\$[-0.0376,-0.0164]$} & \hd{$+0.0079$\\$[-0.0016,0.0186]$} & \hd{$-0.0121$\\$[-0.0246,-0.0011]$} \\
\addlinespace[2pt]
10~s & \hd{$+0.0002$\\$[-0.0011,0.0020]$} & \hd{$+0.0004$\\$[-0.0009,0.0017]$} & $0.0000$ & $0.0000$ \\
\bottomrule
\end{tabular}
\end{table}

\paragraph{Which thinking answers arrive late.} We split valid answers by the action they returned. Thinking answers
that accepted the sample arrived later than those that rejected it, while fast answers showed no such gap
(Table~\ref{tab:timing}). Late thinking answers more often accepted the sample and had higher mean expected gain than on-time answers. To check whether decision difficulty explains the higher gains of late thinking answers, we grouped instances into four bands of $|2p-1|$, a measure of how close the decision was. Averaging within-band differences with weights proportional to band size, late thinking answers still gained more than on-time answers, by 0.194 in the original run and 0.137 in the replication. In some small bands, with 60 or fewer answers, on-time answers gained more. These comparisons are observational and do not show that thinking causes the delay.

\paragraph{Gain by amount of thinking.} We also grouped thinking answers by the number of thinking tokens they used
(Table~\ref{tab:tokens}). Gain rose with the amount of thinking, except for a dip at 1 to 150 tokens, where almost all
answers rejected the sample. No answer with more than 150 thinking tokens arrived within 2.1~s. Arrival time and thinking tokens were closely linked, with rank correlations of 0.89 in the original run and 0.92 in the replication. Arrival time and gain had rank correlations of 0.36 in the original run and 0.32 in the replication. Harder decisions may require more thinking and offer larger
gains, so this pattern does not show that thinking caused the higher gains.

\begin{table}[H]
\centering
\caption{Thinking answers grouped by the number of thinking tokens (posterior expected gain over rejecting, pooled
over instances).}
\label{tab:tokens}
\small
\setlength{\tabcolsep}{4pt}
\begin{tabular}{@{}lrrrrrrrr@{}}
\toprule
& \multicolumn{4}{c}{original run} & \multicolumn{4}{c}{replication} \\
thinking tokens & answers & gain & \hd{on time\\at 2.1~s} & \hd{share\\accepting} & answers & gain & \hd{on time\\at 2.1~s} & \hd{share\\accepting} \\
\midrule
0 & 540 & 0.049 & 96.7\% & 9\% & 125 & 0.069 & 96.8\% & 7\% \\
1 to 150 & 206 & 0.015 & 10.2\% & 1\% & 58 & 0.000 & 17.2\% & 0\% \\
151 to 400 & 306 & 0.278 & 0\% & 29\% & 80 & 0.171 & 0\% & 18\% \\
more than 400 & 148 & 0.526 & 0\% & 63\% & 37 & 0.573 & 0\% & 70\% \\
\bottomrule
\end{tabular}
\end{table}

\begin{table}[H]
\centering
\caption{Arrival time by returned action at the 2.1~s deadline.}
\label{tab:timing}
\small
\begin{tabular}{@{}llrrrr@{}}
\toprule
run & mode & action & answers & median arrival (s) & share on time \\
\midrule
original & fast & accept & 112 & 1.02 & 0.982 \\
original & fast & reject & 1{,}088 & 0.98 & 0.960 \\
original & thinking & accept & 231 & 4.61 & 0.195 \\
original & thinking & reject & 969 & 1.99 & 0.514 \\
\addlinespace[2pt]
replication & fast & accept & 28 & 0.98 & 0.964 \\
replication & fast & reject & 272 & 0.98 & 0.971 \\
replication & thinking & accept & 49 & 4.23 & 0.184 \\
replication & thinking & reject & 251 & 2.17 & 0.486 \\
\bottomrule
\end{tabular}
\end{table}

\section{Additional studies and scope checks}
\label{app:partB}

This part reports the favorable, unfavorable and inconclusive results of the additional studies.
Appendices~\ref{app:serving}, \ref{app:concurrent} and \ref{app:cost} give the full serving, concurrent-selection and hosted cost studies summarized in the main text. The other studies are not used to support its claims.

\subsection{Fault diagnosis, prior sensitivity and image input}
\label{app:faults}
This section tests whether low-effort thinking improves fault-diagnosis decisions and compares model decisions with practical rules when faults are rare. For each fault-diagnosis decision, the model receives six yes/no sensor readings that depend on which of six hidden states is present. In the physical setting, the model may continue, reseat, quarantine or inspect once. In the scientific setting, it may accept, reject or inspect. Deadline-adjusted utility averages the executed action's utility over assumed external deadlines of 10, 30, 60, 300 and 1{,}800~s, weighted .01, .04, .10, .35 and .50. An inspection takes 5~s. The fallback is quarantine for a physical fault and reject for a scientific fault.

\paragraph{Fault probability model.}
State order below is $N$=\texttt{no\_fault}, $P$=\texttt{plate\_unseated}, $S$=\texttt{spill},
$C$=\texttt{missing\_cap}, $O$=\texttt{wrong\_orientation}, $T$=\texttt{tray\_jam}.
Feature order is \texttt{plate\_unseated}, \texttt{spill}, \texttt{missing\_cap}, \texttt{wrong\_orientation},
\texttt{tray\_jam}, \texttt{control\_anomaly}. Features are independent conditional on state.
\begin{table}[H]
\centering\small
\caption{Prior probability of each state and the probability that each binary feature is true (percent). Rows represent hidden states.}
\label{tab:fault-likelihood}
\begin{tabular}{@{}lrrrrrrr@{}}
\toprule
state & \hd{prior\\(percent)} & \hd{plate\\unseated} & spill & \hd{missing\\cap} & \hd{wrong\\orientation} & \hd{tray\\jam} & \hd{control\\anomaly} \\
\midrule
$N$ & 55 & 12 & 8 & 15 & 18 & 7 & 8 \\
$P$ & 12 & 82 & 8 & 15 & 18 & 7 & 35 \\
$S$ & 9 & 72 & 64 & 15 & 18 & 7 & 82 \\
$C$ & 8 & 12 & 8 & 70 & 18 & 7 & 58 \\
$O$ & 10 & 12 & 8 & 15 & 62 & 7 & 22 \\
$T$ & 6 & 12 & 8 & 15 & 18 & 77 & 67 \\
\bottomrule
\end{tabular}
\end{table}
For binary readings $b_f$, state $i$ has prior probability $\pi_i$, and $L_{if}$ is the probability that feature $f$ is true in that state; its posterior weight is $\pi_i\prod_f L_{if}^{b_f}(1-L_{if})^{1-b_f}$, normalized over the states;
these computations use exact rational arithmetic. A sensitivity no-fault prior $p_0$ replaces $\pi_N$ by $p_0$
and each remaining mass by $(1-p_0)\pi_i/(1-\pi_N)$; the likelihood is unchanged.
\begin{table}[H]
\centering\small
\caption{Terminal fault utilities in state order. Inspection's utility cost is $3/2$.}
\begin{tabular}{@{}llrrrrrr@{}}
\toprule
family & action & $N$ & $P$ & $S$ & $C$ & $O$ & $T$ \\
\midrule
physical & continue & 8 & $-14$ & $-30$ & $-18$ & $-14$ & $-22$ \\
physical & reseat & 5 & 7 & $-33$ & $-21$ & 7 & 6 \\
physical & quarantine & $-3$ & $-3$ & $-3$ & $-3$ & $-3$ & $-3$ \\
scientific & accept & 10 & $-12$ & $-24$ & $-8$ & 4 & $-5$ \\
scientific & reject & 0 & 0 & 0 & 0 & 0 & 0 \\
\bottomrule
\end{tabular}
\end{table}
\begin{table}[H]
\centering\small
\caption{Inspection signal probabilities (percent). Rows represent hidden states. Columns represent signals in the same order.}
\begin{tabular}{@{}lrrrrrr@{}}
\toprule
state & $N$ & $P$ & $S$ & $C$ & $O$ & $T$ \\
\midrule
$N$ & 72 & 10 & 4 & 6 & 5 & 3 \\
$P$ & 10 & 68 & 8 & 4 & 7 & 3 \\
$S$ & 6 & 16 & 66 & 6 & 3 & 3 \\
$C$ & 10 & 5 & 6 & 69 & 6 & 4 \\
$O$ & 12 & 12 & 3 & 5 & 64 & 4 \\
$T$ & 8 & 8 & 4 & 4 & 6 & 70 \\
\bottomrule
\end{tabular}
\end{table}
For each possible inspection signal, Bayes' rule updates the state belief and chooses the highest-value terminal
action, without another model call or inspection. The inspection value is this signal-probability-weighted value
minus $3/2$. The reference maximizes over terminal actions and inspection, breaking ties in published terminal order,
then inspection. The counting rule counts the first five flags: scientific accept requires zero flags and no control
anomaly; otherwise reject. Physical zero flags means continue unless the control anomaly calls for inspection;
one flag means quarantine for spill or missing cap, otherwise reseat; two or more flags means inspect.
The threshold rule uses the exact posterior. In the physical setting it continues if the total fault probability is below 0.4, otherwise reseats if the combined probability of an unseated plate, wrong orientation or tray jam is at least 0.6, and otherwise quarantines. In the scientific setting it accepts if the no-fault probability is at least 0.6 and otherwise rejects. It never inspects.
The main fault panel includes every combination of binary sensor readings for both settings (128 combinations), each weighted by its probability under the finite model. Repeated responses do not create new scenarios. The sensitivity panels use no-fault prior probabilities of $.999$, $.99$ and $.95$. The separate challenge panel concentrates on cases near decision boundaries rather than enumerating this population.
\begin{table}[H]
\centering\small
\caption{Prespecified deadline distributions, in seconds. We assumed these probabilities and did not measure laboratory traffic.}
\begin{tabular}{@{}lrrrrr@{}}
\toprule
mixture & 10 & 30 & 60 & 300 & 1,800 \\
\midrule
main comparison & .01 & .04 & .10 & .35 & .50 \\
pause tolerant & 0 & 0 & .05 & .25 & .70 \\
rapid monitoring & .10 & .20 & .30 & .30 & .10 \\
\bottomrule
\end{tabular}
\end{table}
Inspection-time sensitivities are 0, 5 and 30~s, with 5~s primary. Terminal execution is instantaneous in this
abstract fault model. A late or invalid response uses quarantine (physical) or reject (scientific). If inspection
starts on time but cannot finish by the deadline, fallback executes and its value is reduced by the inspection
cost already paid. Otherwise the returned action's posterior expected value is credited. Fault scores use expected utility, which is distinct from simulator-realized stopping utility.

\paragraph{Fault prompts and schemas.}
We used the following fault system message verbatim.
\begin{quote}\small
Choose the action maximizing expected utility using only the public information. Identify a most probable fault
(any top-tied state is correct). Return only JSON satisfying the supplied schema. An inspect action commits to
the published optimal follow-up policy; you need not return its branches.
\end{quote}
The schema requires exactly \texttt{action} and \texttt{fault}: action is a terminal action for the family or
\texttt{inspect}, and fault is one of the six states. The user message for the challenge supplies the public model, task family, schematic legend and perceptual settings. The text and image-plus-text conditions include the sensor report. In the image conditions, the message includes a base64-encoded PNG image followed by the standard text. In the image-only condition, that text omits feature values. When we supply the computed posterior, we provide it only as text. The later fault study uses raw text input. It replaces the scope statement with ``Synthetic fault model; finite evidence enumeration; no equipment is operated.'' It also replaces the prior with the rescaled prior, removes the schematic legend and perceptual settings, and adds an inspection duration of 5 seconds. The following text gives the exact replacement deadline rule.
\begin{quote}\small
If response latency exceeds deadline or output fails validation, use quarantine (physical) or reject (scientific).
Latency excludes pacing. Inspection takes 5 seconds after inference; credit its follow-up only if both finish by the
deadline. If inspection starts but finishes late, pay its utility cost and use safe fallback. Terminal execution takes
zero seconds in this abstract model.
\end{quote}

\paragraph{Primary result.} The prespecified test used repeated responses to the fixed 128-case fault model. Low-effort thinking increased deadline-adjusted utility relative to fast mode by 3.65 fault-utility units (98.75\% CI $[2.43,4.86]$). The cases were fixed, so new fault scenarios might give a different result. This fault-diagnosis test is one of the four tests in the family in Table~\ref{tab:confirm-supp}.

\paragraph{Comparison with practical rules.} An earlier offline challenge used 120 fault cases concentrated near decision boundaries. With text input, thinking had much lower decision regret than the mode without thinking, at 0.054 against 3.08 fault-utility units (Holm-adjusted $p=.003$). This challenge used medium effort with the provider's default thinking setting. Thinking answers took 21~s on average, and fast answers took 1.1~s. Thinking's advantage in utility was $+3.03$ (95\% CI $[2.42,3.69]$) with a 60~s deadline and $+0.30$ ($[-0.35,1.02]$) with a 10~s deadline. Table~\ref{tab:rules} compares practical rules. When the model read the same evidence from schematic images, decision regret with thinking rose by 0.55.

In sensitivity scenarios where the default action was almost always correct, simple rules were near-optimal, with regret of 0 to 0.112 for the threshold procedure. Haiku in fast mode was too cautious, with fault decision regret of 3.61 to 4.57. With thinking, fault decision regret was at most 0.015. Published task-exception rates~\cite{alabos} motivated the assumed no-fault probabilities, 0.95 to 0.999. Those rates count reported exceptions, which may miss hidden faults. On the earlier 240-instance panel testing sensitivity to the prior, terminal stopping regret was higher with medium-effort thinking than with the $p\geq0.98$ rule, at 0.054 against 0.037. Table~\ref{tab:rules} reports these descriptive comparisons, and Table~\ref{tab:historical} gives the earlier prespecified fault contrasts.

\begin{table}[H]
\centering
\caption{Decision regret of deployable rules and Claude Haiku 5.5 by evaluation panel. Lower is better, and the exact Bayesian rule has regret 0 by construction. Panels differ in instances, prompts and utility units, so comparisons should be made within a row only. Fault rows use fault-utility units. The stopping row uses the $\pm1$ scale. For the prior-0.99 and stopping rows, ``thinking'' means medium-effort thinking in the panel testing sensitivity to the prior. For the boundary panel, it means the thinking mode used in the fault challenge study.}
\label{tab:rules}
\small
\setlength{\tabcolsep}{2.5pt}
\resizebox{\columnwidth}{!}{%
\begin{tabular}{llrr}
\toprule
panel & practical rules & without thinking & thinking \\
\midrule
fault, boundary-enriched (120) & counting 1.63; threshold 1.89 & 3.08 & 0.054 \\
fault, no-fault prior 0.99 & posterior threshold 0.005; counting 3.68 & 4.57 & 0.015 \\
stopping, terminal (240 new) & $p\!\geq\!0.98$ rule 0.037; majority 0.152; reject 0.202 & 0.225 & 0.054 \\
\bottomrule
\end{tabular}}
\end{table}

\subsection{Local open-weight inference}
\label{app:local}
This study asks whether thinking also benefits a locally run model with publicly available weights and whether that benefit remains under the physical deadline. We run Qwen3-8B using vLLM inference software on one A6000 GPU. Without thinking, the configuration uses temperature 0.7, top-$p$ 0.8 and a 1{,}024-token cap. With thinking, it uses temperature 0.6, top-$p$ 0.95 and an 8{,}192-token cap. Unlike the hosted effort levels, local
comparisons change thinking, sampling and token cap together.

\paragraph{Result.} On a separate local stopping panel, the Qwen3-8B thinking configuration returned better terminal decisions, but every terminal thinking answer arrived too late and triggered the fallback. Terminal regret with thinking was 0.24 lower than without thinking (95\% CI $[0.12,0.37]$, descriptive under the prespecified test order). Responses with thinking took 12.5 to 90.9~s, while only about 2~s remained before the physical deadline. Mean terminal regret was 0.277 without thinking and 0.037 with thinking. Adding the computed posterior together with explicit utility and stopping guidance to the prompt
did not detectably reduce terminal regret. Regret without this support minus regret with it, averaged over both configurations, was $+0.042$ (95\% CI $[-0.011,0.093]$, $p=.11$).

\subsection{GPU capacity and serving load}
\label{app:serving}
This section measures how sharing one local GPU among several requests changes the comparison between modes. These experiments are separate from the hosted replay of answer value and arrival time.

\paragraph{Capacity scan.} A prespecified capacity scan found no sustained saturation through $K=56$ simulated laboratories running at once (about 0.5 requests/s, with requests waiting in at most 2\% of samples). The natural Qwen3-8B workload without thinking left spare capacity on one A6000 and no lasting queue. We therefore constructed bursts of simultaneous requests. Interference may differ when requests arrive naturally in a laboratory.

\begin{figure}[H]
\centering
\includegraphics[width=0.50\textwidth]{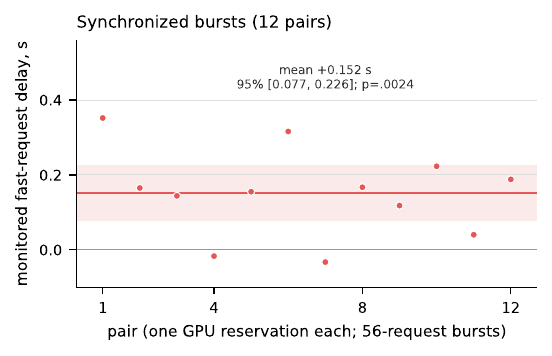}
\caption{Shared-serving interference in synchronized 56-request bursts (12 paired allocations). Points show the change in monitored fast requests' time to a valid response. The line and band show the mean and approximate 95\% paired-block interval. The exact randomization test gives $p=.0024$. The 5~s relevance threshold lies outside the displayed range. All monitored decisions remained timely and valid.}
\label{fig:burst_interference}
\end{figure}

\paragraph{Shared-serving interference.} A separate experiment held concurrency at 56 and switched 42 of the 50 background requests from fast to thinking, while six monitored requests stayed fast. In the prespecified comparison, the switch slowed the monitored requests by 0.152~s on average (approximate paired 95\% $t$ interval $[0.077,0.226]$). This increase was well below the prespecified 5~s practical threshold. The monitored requests' mean time to a valid response was 13.6~s. All 144 monitored decisions remained timely and valid at 20~s, with identical actions in both conditions (Figure~\ref{fig:burst_interference}).

\paragraph{Design safeguards.} We randomized condition order and separated conditions by warm-up, an idle interval and completion of outstanding requests. These safeguards reduce interference between successive conditions. Configurations change thinking, sampling and token cap together. Pair~0 required a second technical retry, beyond the prespecified maximum of one. We allowed this retry before observing any outcome (Appendix~\ref{app:deviations}).
\paragraph{Synchronized bursts.} The experiment used 12 randomized paired allocations and a fixed 36-case panel. Each allocation used one A6000 GPU running Qwen3-8B on vLLM (prefix caching disabled), with synchronized bursts of $K=56$ requests. In the control burst all 56 requests disable thinking. In
  the treatment burst 42 of the 50 background requests enable thinking, with different sampling settings and token caps. We measure the response times of six monitored fast requests. These requests occupy fixed positions and have the same case, seed and position in both conditions.
\paragraph{Interference contrast and test.} The primary contrast is the mean change in the monitored fast requests' capped time to a valid response. This capped time equals the receipt time for a valid answer received within 20~s and 20~s for a late, invalid or failed answer. We use an exact two-sided randomization test over $2^{12}$ sign
  assignments and an approximate paired-block $t$ interval, with a prespecified practical-relevance threshold of 5~s.

\paragraph{Constructed load.} This prespecified study ran Qwen3-8B on one A6000. Each burst contained six scored requests, all in the fast or all in the thinking configuration, plus background requests that all used thinking. Each burst had 8 requests at low load and 56 at high load. Thinking scored lower than fast mode at both loads (Table~\ref{tab:load-conditions}), and the gap grew with load. Raising the number of simultaneous requests from 8 to 56 widened thinking's deficit in the posterior expected utility of executed actions by 0.1970 (95\% $t$ CI $[0.1387,0.2553]$). This was the study's prespecified primary comparison. The test treats the 12 allocations as independent units. Requests queued during the decision window in all 12 allocations. A later analysis of server and client timings splits the extra delay into its parts (Appendix~\ref{app:load-breakdown}). Appendix~\ref{app:load} gives the design, the per-allocation differences and server measurements.

\paragraph{Serial and shared serving results.} This descriptive, prespecified panel used deadlines of 45~s and 240~s, with six blocks and 144 decisions. At 45~s, thinking had lower executed posterior expected utility despite lower answer regret. Late answers triggered rejection in 5/18 decisions under serial serving and 7/18 under shared serving. At 240~s, the difference favored thinking slightly under serial serving and favored the configuration without thinking under shared serving. Under shared serving, one thinking answer was empty at the token cap before the deadline. Without thinking, actions and regret were unchanged across serving modes. All four contrast intervals include zero. The two deadlines use different observations.

\subsubsection{Constructed-load study in detail}
\label{app:load}

The prespecified constructed-load study tested bursts of $K\in\{8,56\}$ concurrent requests. Each burst included six requests scored for decision utility, all using either fast mode (without thinking) or the thinking configuration. Each combination of load and configuration is a cell. All background requests generated load and used the thinking configuration. There were two background requests at the lower load and 50 at the higher load. We used one A6000, Qwen3-8B revision b968826d,
vLLM 0.30.0 and xgrammar 0.2.8, with prefix caching disabled. The two configurations change thinking, sampling and the token cap together (Appendix~\ref{app:local}). Both use top-$k$ 20 and min-$p$ 0.

The panel consists of the 36 original terminal observations, grouped into six blocks of six. Each block appears in
two allocations with fresh seeds, giving 12 scored allocations and 72 scored decisions per cell. Each allocation runs all four cells on the same server and GPU, using the same scored observations and seeds in every cell. All outstanding requests finish before the next cell begins. Seeded permutations assign cases to request positions. We balance condition order using a Latin square, so each condition occupies each position in the order three times. Background cases cycle through the
development observations in a fixed order, without selection on performance, and are identical across scored
configurations at each load.

Decisions use the original 45~s test setup with the unchanged accept/reject rules, parser and strict deadline. An answer received at or after 45~s is late. We let late inference requests finish rather than canceling them. Terminal removal is
charged 10~s. Late, invalid, empty, token-cap-truncated, refused or transport-failed answers trigger rejection. The executed posterior expected utility of a decision is $(2p-1)\mathbb{I}\{\text{committed accept}\}-0.0001$, where $p$ is the observation's exact posterior. This expected utility differs from simulator-realized utility. Before each cell's synchronized burst, we perform an idle check, development warm-ups, a second idle check and an idle interval. The next cell starts only after all post-burst requests finish and the isolation checks pass.

We fixed the design, assignment, endpoint and test before any model call, including those of the load check below.

\paragraph{Pre-collection load check.} Before collecting scored data, one development allocation had to show that the intended load occurred within the 45~s decision window. The check never examined utility. With thinking at high load, requests had to be waiting in at least 25\% of one-second samples, and at low load in at most 10\%. The observed shares were 35.6\% (peak 42 waiting) and 2.2\%. All 12 scored allocations then ran without errors, and none had to be replaced.

\paragraph{Executed-utility contrast.} For allocation $b$, let $A_{b,K}$ be the mean executed posterior expected utility with thinking minus that with fast mode over the six scored decisions at concurrency $K$. Let $W_b=A_{b,8}-A_{b,56}$. A positive $W_b$ means that higher load worsens thinking's result relative to fast mode. $A_8$ and $A_{56}$ denote the means of $A_{b,K}$ over allocations. The primary quantity $W$ is the mean of the 12 $W_b$. We test this mean with one two-sided paired $t$ test at $\alpha=.05$ (11 degrees of freedom) and report a 95\% $t$ interval. The inference is approximate and conditional on the fixed panel and on adequate removal of interference between cells. The six scored decisions within an allocation are not independent replications. All mechanism
and component results are descriptive.

Increasing concurrency from 8 to 56 widened thinking's deficit relative to fast mode in executed posterior expected utility by $W=+0.1970$ (95\% $t$ CI $[0.1387,0.2553]$, paired $t(11)=7.44$, approximate $p=1.3\times10^{-5}$).  The thinking configuration showed no executed-utility advantage at either load. The estimates were $A_8=-0.0867$ $[-0.1374,-0.0360]$ and $A_{56}=-0.2837$ $[-0.3127,-0.2547]$ (descriptive 95\% $t$ intervals across allocations). All 12 allocation contrasts were positive (Table~\ref{tab:load-allocations}).

\begin{table}[H]
\centering
\caption{Widening of thinking's deficit relative to fast mode when concurrency rises from 8 to 56, by allocation ($W_b$, three decimals). Positive values mean a larger deficit at higher load.}
\label{tab:load-allocations}
\small
\begin{tabular}{@{}lrrrrrrrrrrrr@{}}
\toprule
allocation $b$ & 0 & 1 & 2 & 3 & 4 & 5 & 6 & 7 & 8 & 9 & 10 & 11 \\
\midrule
$W_b$ & .093 & .196 & .285 & .061 & .147 & .167 & .305 & .196 & .148 & .120 & .314 & .333 \\
\bottomrule
\end{tabular}
\end{table}

\begin{table}[H]
\centering
\caption{Descriptive results by condition, with 72 scored decisions each. $K$ is the number of concurrent requests. Times are means in seconds. Submission and receipt times are measured from the trigger. Time to first token (TTFT) and completion time are measured from submission. Receipt time includes polling, and late requests are allowed to finish. Capped time equals receipt time for timely valid answers and 45~s otherwise. Decision regret uses available answers only, with the denominator in parentheses.}
\label{tab:load-conditions}
\scriptsize
\setlength{\tabcolsep}{4pt}
\begin{tabular}{@{}llrrrrrrrrrr@{}}
\toprule
$K$ & \hd{scored\\mode} & \hd{executed\\utility} & \hd{timely\\and valid} & submission & TTFT & completion & receipt & \hd{capped\\time} & late & \hd{token-cap\\truncation} & \hd{regret\\($n$)} \\
\midrule
8 & fast & .2811 & 100\% & .131 & 1.52 & 2.43 & 2.57 & 2.57 & 0 & 0 & .0345 (72) \\
8 & thinking & .1944 & 79.2\% & .128 & 1.95 & 39.24 & 39.37 & 31.51 & 14 & 1 & .0260 (71) \\
56 & fast & .2811 & 100\% & .711 & 8.42 & 13.47 & 14.18 & 14.18 & 0 & 0 & .0345 (72) \\
56 & thinking & $-.0026$ & 1.4\% & .711 & 9.33 & 86.69 & 87.41 & 44.91 & 71 & 0 & .0481 (72) \\
\bottomrule
\end{tabular}
\end{table}
These load comparisons are descriptive. Higher load sharply reduced the share of timely valid thinking answers, from about 79\% at 8 concurrent requests to 1.4\% at 56. Fast answers remained timely in every decision, while their mean receipt time rose from 2.6 to 14~s. Apart from the one token-cap truncation, no answer was invalid, refused, lost in transport or timed out at the client. At the lower load, thinking had slightly lower decision regret among available answers and no executed-utility advantage. At the higher load, executed utility was close to the fallback's utility.

\paragraph{Queueing.} Requests were waiting in about 2\% of one-second samples at low load and 34\% at high load, with a peak of 38 to 46 waiting. No preemptions occurred within the scored windows. Time to first token includes more than queue waiting, and these samples cannot attribute queueing delay to individual scored requests. Appendix~\ref{app:load-breakdown} splits the delay using the server's timing totals for each burst.

\paragraph{Record check and reproduction.} Before computing any comparison, we checked the raw records of the pre-collection check and all 12 scored allocations against the study's execution rules. All passed.

We checked a separately written implementation of the primary analysis beforehand on synthetic runs with positive, null and opposite-direction results (agreement within $3\times10^{-13}$). This implementation reproduced $W$, its interval and $t$, and all 12 $W_b$.
The reproduction checks computation and recorded execution only.

\subsubsection{Where the extra delay came from}
\label{app:load-breakdown}

We conducted this descriptive analysis after the study. For every request, the client recorded the sending time, first-token arrival time, completion time and time when the runtime received the answer. The client also recorded token counts. Times
are measured from the start of the burst. The server kept running totals of the time each request waited to be
scheduled, the time from scheduling to its first token and the time it spent generating its answer. These are elapsed
times, not time spent computing. The server also counted generated tokens and preemptions, which pause a running
request to free memory. Each burst began and ended with an idle server, so the change in these totals over a burst
covers exactly that burst's requests. In all 48 scored cells, the server counted $K$ requests and the same number of
generated tokens as the client, and it recorded no preemptions.

Table~\ref{tab:breakdown-focal} breaks down the response times of the scored answers. Increasing the load added 48.0~s to a thinking answer's response time. This consisted of 0.6~s before the request was sent, 7.4~s from sending to the first token and 40.1~s of generation. We divided the added generation time into contributions from longer answers and slower generation. We averaged estimates obtained by changing length first and by changing speed first. The contributions were 1.7~s from longer answers and 38.4~s from slower generation. Thinking answers differed between
loads even with the same seed. At high load, thinking answers had a median length of about 1{,}260 tokens. Fast answers had seven tokens at both loads. Their added delay of 11.6~s consisted of
0.6~s before sending, 6.9~s from sending to the first token and 4.1~s of generation.

\begin{table}[H]
\centering
\caption{Response times of the scored answers by load (means over 72 answers per cell, in seconds unless noted). The first three timing columns show consecutive parts of the total time until the runtime received the answer. Generation speed is not shown for fast answers, which have seven tokens.}
\label{tab:breakdown-focal}
\small
\setlength{\tabcolsep}{4pt}
\begin{tabular}{@{}llrrrrrrrr@{}}
\toprule
mode & $K$ & \hd{until\\sent} & \hd{to first\\token} & generation & total & tokens & tokens/s & \hd{valid and on\\time at 45~s} & \hd{90th-percentile\\arrival} \\
\midrule
fast & 8 & 0.13 & 1.52 & 0.91 & 2.57 & 7 & -- & 100\% & 2.7 \\
fast & 56 & 0.71 & 8.42 & 5.04 & 14.18 & 7 & -- & 100\% & 17.7 \\
thinking & 8 & 0.13 & 1.95 & 37.29 & 39.37 & 1{,}359 & 35.8 & 79.2\% & 69.5 \\
thinking & 56 & 0.71 & 9.33 & 77.37 & 87.41 & 1{,}430 & 17.4 & 1.4\% & 126.5 \\
\bottomrule
\end{tabular}
\end{table}

Across all requests in a burst, the wait to be scheduled grew from 0.1 to 5.8~s and the time from scheduling to the
first token from 1.3 to 2.3~s (Table~\ref{tab:breakdown-server}). In bursts whose scored requests used thinking, mean generation time rose from 36.5 to 74.6~s, and the time per token rose from 28 to 56~ms. The last request to produce a first token did so an average of 17.5~s after the burst began at high load, compared with 2.4~s at low load. Dividing the roughly 124{,}000 prompt tokens by the high-load interval gives about 7{,}100 tokens per second. Generation overlapped that interval. On average, fast answers finished 3.3~s before the interval ended. In the sampled intervals without prompt processing, estimated
generation speed per request was lower when more requests ran together, while total output was higher
(Table~\ref{tab:batch-speed}).

\begin{table}[H]
\centering
\caption{Server timing totals over all requests in a burst (means per request over 12 allocations, in seconds unless noted). Waiting is the time from arrival until the server schedules the request. The next column gives the time from scheduling until the first token. Burst output is the number of generated tokens per second from the start of the burst to its last token.}
\label{tab:breakdown-server}
\small
\begin{tabular}{@{}llrrrrrr@{}}
\toprule
\hd{scored\\mode} & $K$ & waiting & \hd{to first\\token} & generation & \hd{ms per\\token} & \hd{tokens per\\request} & \hd{burst output\\(tokens/s)} \\
\midrule
fast & 8 & 0.13 & 1.28 & 8.31 & 27.5 & 306 & 59 \\
thinking & 8 & 0.14 & 1.25 & 36.48 & 28.0 & 1{,}310 & 140 \\
fast & 56 & 5.81 & 2.26 & 65.49 & 53.8 & 1{,}220 & 396 \\
thinking & 56 & 5.81 & 2.25 & 74.56 & 55.6 & 1{,}342 & 438 \\
\bottomrule
\end{tabular}
\end{table}

\begin{table}[H]
\centering
\caption{Generation speed by the number of requests running together, from one-second server samples with no prompt
processing or waiting requests (all 48 scored cells). We estimate per-request speed by dividing server output by the midpoint of each band of running-request counts.}
\label{tab:batch-speed}
\small
\begin{tabular}{@{}lrrrrrrrrr@{}}
\toprule
running requests & 1 & 2 & 3--4 & 7--8 & 9--16 & 17--24 & 25--32 & 41--48 & 49--56 \\
\midrule
server output (tokens/s) & 39 & 77 & 126 & 261 & 350 & 483 & 606 & 799 & 938 \\
per request (tokens/s) & 39.4 & 38.3 & 35.9 & 34.8 & 28.0 & 23.6 & 21.3 & 18.0 & 17.9 \\
\bottomrule
\end{tabular}
\end{table}

The observed arrival times and generation speeds give a rough limit on answer length. Consider an answer of $n$ tokens generated at a steady $r$ tokens per second after its first token at time $t_0$. Ignoring the few milliseconds of polling, the answer arrives by deadline $s$ only if roughly $n\le r(s-t_0)$. With the observed mean values of $t_0$ and $r$, the limit at 45~s was about 1{,}540 tokens at low load and 610
at high load. Applying this limit to each answer's actual length gives 75\% of thinking answers on time at low load, close to the observed 80.6\% (79.2\% were also valid). At high load, the limit gives 11\%, compared with the observed 1.4\%, because start times and generation speeds varied across answers. The limit uses speeds measured afterward, so it describes the result and is not a forecast.

The same totals describe the two other serving results. In the shared-serving test described above, switching
background requests to thinking changed little before the first token. Mean waiting time before scheduling was 5.45 against 5.47~s. Mean time from scheduling to the first token was 2.22 against 2.17~s. In the capacity scan without thinking, mean waiting stayed at or below
0.13~s at every load from 8 to 56 laboratories. Over the same range, the time per token rose from 27--30~ms to 42--70~ms.
In the capacity scan, higher load increased generation time more than waiting time.

\paragraph{Scope.} These results concern one model revision on one A6000, a fixed 36-case synthetic terminal panel,
thinking-heavy background traffic and synchronized arrivals. These studies do not test the effect of thinking alone, behavior under natural laboratory demand, sustained capacity or a saturation point. They also do not test whether the results generalize across models or GPUs, or whether an LLM is needed instead of the inexpensive exact controller. Executed posterior expected utility remains distinct from
simulator-realized utility and from physical validation. The capacity scan, which used the natural workload without
thinking and found no sustained saturation, is a separate result.

\subsection{Concurrent selection, output constraints and an effort controller}
\label{app:concurrent}
These studies test whether requesting both modes at once recovers value lost to deadlines, and how executed decisions change when the model
may output only legal actions.

\paragraph{Rule and comparisons.} Concurrent selection requests both modes. It executes a timely valid thinking answer if one is available. If no thinking answer qualifies, it executes a timely valid fast answer, or the fallback if neither answer qualifies. Two comparisons estimate different
quantities. For the recorded-answer evaluation, we applied the rule after the fact to responses issued separately in the dispatch-clock evaluation. For each instance and deadline, we used the low-effort thinking outcome if its answer was valid and received in time. Otherwise, we used the fast outcome if its answer met both conditions, or rejection if neither answer qualified. Two live cohorts later ran
the rule with concurrent dispatch. After the recorded-answer evaluation and before the live cohorts, a latency check tested whether concurrent dispatch changes latency.

\paragraph{Latency check design.} We prespecified the study. It used new instances for 2 preliminary checks, 12 pilot trials and 300 scored trials, with seeds disjoint from all earlier hosted-study instances. Each request matched a request from the dispatch-clock evaluation, with a deadline of 4~s. We ran each instance under concurrent and separate dispatch, in randomized order, with at least 6~s between conditions. Under concurrent dispatch, both provider calls started together, with at most 0.62~ms between their observed start times. Under separate dispatch, the second request started after the first finished and at least 6~s after the first started.
  Every provider call used a fresh connection, so latency includes connection setup in both conditions.

\paragraph{Live cohort design.} Both studies were prespecified, and the larger live cohort was prespecified
  before any smaller-cohort outcome was inspected. The smaller live cohort used 162 new instances and the larger used 600, with disjoint seeds. We evaluated each instance at 2.1 and 4~s using the dispatch-clock protocol's requests, clock, scoring and simulator. We start both provider calls together, with at most 1.6~ms between their start times. We execute a valid low-effort thinking answer on arrival if it arrives by the deadline. Otherwise, we execute a valid fast answer at the deadline if it has arrived by then, or reject if neither answer qualifies. We allow late requests to finish without cancellation and never execute their answers. We scored 103 and 468 such late answers in the smaller and larger cohorts, respectively.

\paragraph{Analyses.} At each of the five deadlines, we subtract the better fixed mode's utility from the selection rule's utility. The recorded-answer contrast averages these differences using equal prespecified weights. We bootstrap instances with 10{,}000 draws, keeping all ten cells of each instance together. We recompute the maximum in every draw. The same draws give the 95\% percentile CI and the approximate two-sided bootstrap $p=2\min(P^*(C^*\le0),P^*(C^*\ge0))$ with the plus-one rule. Here, $C^*$ is the resampled selection contrast and $P^*$ denotes bootstrap probability. The comparison uses the larger of two estimated means, so small changes can switch the selected mode when the means are nearly equal. We did not separately calibrate the bootstrap for that case. Thus, $p$ is approximate and conditional on the prespecified procedure. This test has its own family at $\alpha=.05$. For each mode, the latency check's primary quantity is the paired geometric-mean ratio of concurrent to separate dispatch latency. We estimate it by bootstrapping instances with 10{,}000 draws. Equivalence requires the 90\% CI to lie within $[1/1.1,1.1]$. We declared latency equivalence only if both modes met this requirement. For the larger live cohort, we compare the live selection rule with a replay of the fixed mode that has the higher observed mean utility on the same concurrent answers. The primary utility contrast averages the differences over the two deadlines with equal weights. We bootstrap instances with 10{,}000 draws and recompute the maximum in each draw.

\begin{figure}[H]
\centering
\includegraphics[width=0.6\textwidth]{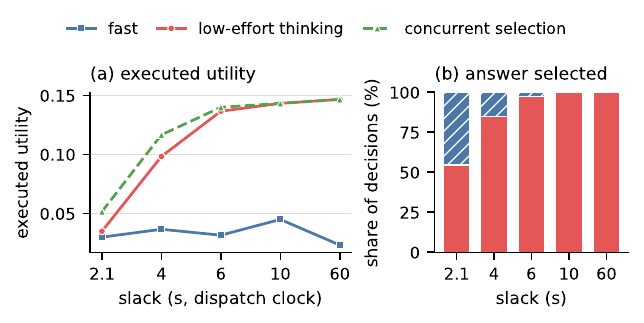}
\caption{Concurrent-request selection on recorded answers (the dispatch clock, 600 instances per cell). (a) Executed utility of each fixed mode and of the concurrent selection rule, evaluated on the same recorded answers. (b) The recorded answer the rule would select, either low-effort thinking (red), fast (blue, hatched) or the fallback (gray, at most 0.2\%). Only the averaged difference between selection and the best fixed mode is confirmatory. The best fixed mode is chosen in hindsight, and the comparison is counterfactual, using recorded answers. All comparisons at individual deadlines are descriptive.}
\label{fig:tiered}
\end{figure}

\paragraph{Recorded-answer results.} With equal weights across the five deadlines, the gain over the better fixed mode at each deadline was $+0.0077$ (95\% CI $[0.0013,0.0123]$, approximate $p=.017$). These deadlines form a grid chosen for the study, and real workloads may weight them differently. This approximate bootstrap is uncalibrated near ties between modes. Gains occurred where thinking missed the deadline. At 2.1~s, utility was 0.052 against the best fixed mode's 0.035, and the rule selected fast in 45\% of cases. At 4~s, utility was 0.117 against 0.098. There was no gain at 10 or 60~s. The gain from selection over thinking alone equals the probability that only the fast answer is timely, multiplied by the fast answer's gain over fallback conditional on that event. The numbers of decisions executing thinking/fast/fallback, out of 600, were 327/272/1, 510/90/0, 583/17/0, 599/1/0 and 600/0/0 at 2.1, 4, 6, 10 and 60~s, respectively. Cost per decision is the sum of the costs of both recorded calls. It is $1.78\times$ the cost of low-effort mode and $2.29\times$ the cost of fast mode. The same computation on the evidence-onset records is exploratory because their outcomes had already been seen. We detected no gain ($+0.002$, with an interval spanning zero). Figure~\ref{fig:tiered} shows these results.

\paragraph{Latency results.} The geometric-mean latency ratio for concurrent versus separate dispatch was 0.972 for fast mode (90\% CI $[0.944,0.999]$, 95\% CI $[0.939,1.004]$). For low-effort thinking, the ratio was 1.020 (90\% CI $[0.962,1.086]$, 95\% CI $[0.950,1.099]$). Both modes met the prespecified margin. For fast mode, median/99th-percentile latencies were 1.01/2.29~s under concurrent dispatch and 1.00/4.24~s under separate dispatch. For low-effort thinking, they were 2.36/7.08~s and 2.33/6.63~s, respectively.
  The shares of answers arriving in time under concurrent and separate dispatch differed by at most two percentage points (descriptive). All 1{,}256 calls succeeded, all answers were valid, and the calls cost \$0.37. This result covers one model and provider in one afternoon. It concerns geometric-mean latency and does not establish equivalence of tail latency, deadline success or utility.

\paragraph{Live results.} To check fidelity, we rerun the recorded-answer rule in a new simulator instance and require the live action and utility to match within $10^{-12}$. All 324 decisions of the smaller cohort and all 1{,}200 of the larger cohort met this criterion. This check assesses agreement with the rule separately from the utility comparison. In the larger live cohort, we did not detect the prespecified utility gain. The estimate was $+0.0125$ (95\% CI $[-0.0008,0.0233]$, $p=.073$). The gains at each deadline were $+0.0117$ at 2.1~s, where fast was the best fixed mode, and $+0.0133$ at 4~s, where low-effort thinking was best. The interval includes zero, so the live gain is inconclusive. It also includes the recorded-answer estimate of 0.0077. The thinking/fast/reject execution counts were 272/313/15 at 2.1~s and 485/115/0 at 4~s, each out of 600. Cost was 1.77$\times$ that of low-effort thinking alone, with cohort costs of \$0.21 and \$0.73, respectively. Live utility of concurrent selection was 0.012 and 0.092 in the
  smaller cohort (descriptive) and 0.047 and 0.100 in the larger cohort, at 2.1 and 4~s respectively.

\paragraph{Output constraints change executed decisions.} We compared two output grammars on the same decisions (Qwen3-8B, 880 responses, all present). These grammars specify which proposals the model may output. One allows any well-formed proposal. If the validator rejects a proposal, the process keeps waiting. The other allows only actions legal in the observed state. Both allow waiting. On balanced snapshots from a recorded simulator run without constructed decision boundaries, constraints prevented the tested class of undeclared actions. They still allowed all 100 valid recorded responses. We also constructed a panel of coordination snapshots near decision boundaries around equipment outages and resource holds. We captured the snapshots 0, 15 or 40~s after the event, with deadlines 10 or 160~s away, so that the best action depends on timing. On this panel, restricting outputs to legal actions increased executed regret relative to allowing any well-formed proposal. The increase was 0.55 without thinking and 0.17 with thinking at zero added delay, and 0.55 and 0.11 with measured delay. Constraints replaced some proposals that would have triggered validator rejection and waiting with legal but harmful actions. Because the cases were constructed and the result is descriptive, we draw no general conclusion about constrained decoding~\cite{constrainttax}. The coordination snapshots come from our earlier work with the same simulator. 

\paragraph{Offline effort controller.} In grouped cross-validation on earlier responses, an offline controller that selected effort based on the deadline showed no advantage on held-out responses over the best fixed mode (estimated difference $-0.012$). We therefore did not test it live.

\subsection{Hosted value and cost study}
\label{app:cost}
We planned this one-hour study (Section~8) before collection. After a pilot, we increased its sample sizes based only on cost and duration. Every result is descriptive, and the intervals are pointwise.

\paragraph{Decisions and difficulty.} Before any call, we labeled a pool of 12{,}000 new terminal instances. These were ordered draws from new seeds, balanced over family and stage. The latest-reading rule accepts if the latest quality reading is at least 0.8. The mean-reading rule accepts if the mean of all quality readings is at least 0.8. The exact Bayes action accepts if $p>0.5$.
Both rules are correct on easy decisions, one is correct on conflicting decisions, and neither is correct on hard decisions. The pool frequencies are 0.796, 0.179 and 0.025, respectively. We selected the first 160 decisions at each difficulty level in pool order. Accepting was optimal in 12.5\%, 92.5\% and
73.1\% of easy, conflicting and hard decisions, and the exact Bayes value per decision was 0.112, 0.894 and 0.422.

\paragraph{Configurations and collection.} Requests reuse the replay's 60~s payload (system message, observation and JSON schema). Only model, thinking, effort and speed differ. Haiku fast disables thinking, and Haiku thinking uses adaptive thinking. Sonnet fast uses the setting that permits no thinking before the answer. Opus and Fable cannot disable thinking and run at low effort. Opus with accelerated output uses the provider's paid faster-output option. Effort is low throughout,
with a 16{,}000-token cap, no sampling parameters, no retries and no refusal fallback. Prices per million input/output tokens are \$0.10/\$0.50 for Haiku, \$2/\$10 for Sonnet, \$4/\$20 for Opus, \$8/\$40 for Opus with accelerated output and \$10/\$50 for Fable.
Configurations ran concurrently in one shared interleaved order, with up to four unfinished requests per configuration.
The deadline-aware prompts for Haiku thinking and Opus stated deadlines of 2.1~s or 10~s. These prompts and two live pairs used the first 80 decisions of each difficulty level. The live pairs were Haiku fast with Opus and Sonnet fast with Fable. Both requests in each pair started together. In a load phase, we sent requests to Haiku fast and thinking at random times according to a Poisson process, at rates of 0.5 and 5 requests/s. Each condition used three 60~s blocks in a fixed random order, with at most 64 unfinished requests. We measured deadlines from scheduled arrival. A pilot of 408 calls (\$1.53) checked every setting. The main run made
6{,}724 calls (\$82.62), all valid. After the pilot, based only on cost and duration, we extended Fable from 240 to all 480 decisions. We also extended the deadline-aware prompts and live pairs from 120 to 240.

\begin{table}[H]
\centering\footnotesize
\caption{Hosted configurations on the 480-decision panel. Value is the posterior expected gain over rejection. Here, ``all'' uses every answer regardless of arrival time. The column ``hard accuracy'' gives the share of hard decisions answered with the exact Bayes action. $G(s)$ is the gain delivered by deadline $s$ ($G_{\mathrm{joint}}$ of Section~2). Cost is the mean list-price cost per call.}
\label{tab:cost-configs}
\setlength{\tabcolsep}{3pt}
\begin{tabular}{@{}lrrrrrrrrr@{}}
\toprule
configuration & \hd{value\\(all)} & \hd{hard\\accuracy} & \hd{median\\latency (s)} & \hd{90th-percentile\\latency (s)} & \$/call & $G(2.1)$ & $G(4)$ & $G(10)$ & $G(60)$ \\
\midrule
Haiku fast & 0.099 & 0.28 & 0.92 & 1.14 & 0.00026 & 0.097 & 0.099 & 0.099 & 0.099 \\
Haiku thinking & 0.258 & 0.28 & 1.57 & 3.66 & 0.00037 & 0.062 & 0.221 & 0.258 & 0.258 \\
Sonnet fast & 0.307 & 0.17 & 1.37 & 2.05 & 0.00523 & 0.284 & 0.300 & 0.307 & 0.307 \\
Opus & 0.454 & 0.88 & 4.68 & 10.01 & 0.01851 & 0.047 & 0.161 & 0.439 & 0.454 \\
Opus accelerated & 0.461 & 0.91 & 2.08 & 4.34 & 0.03814 & 0.217 & 0.438 & 0.457 & 0.461 \\
Fable & 0.420 & 0.60 & 3.28 & 20.56 & 0.05202 & 0.177 & 0.204 & 0.282 & 0.420 \\
\bottomrule
\end{tabular}
\end{table}

\begin{table}[H]
\centering\footnotesize
\caption{Results by difficulty. Entries for each difficulty level give value delivered by 2.1~s / by 10~s. The weighted value uses the pool frequencies and a deadline of 2.1~s. Arrival times are medians in seconds. Exact Bayes values are 0.112 for easy decisions, 0.894 for conflicting decisions and 0.422 for hard decisions. Entries of $-0.00$ are small negative values.}
\label{tab:cost-levels}
\setlength{\tabcolsep}{3pt}
\begin{tabular}{@{}lrrrrrrr@{}}
\toprule
configuration & easy & conflicting & hard & \hd{weighted\\value} & \hd{easy\\time (s)} & \hd{conflicting\\time (s)} & \hd{hard\\time (s)} \\
\midrule
Haiku fast & $-$0.00/$-$0.00 & 0.29/0.29 & 0.01/0.01 & 0.048 & 0.9 & 0.9 & 0.9 \\
Haiku thinking & 0.01/0.08 & 0.13/0.65 & 0.04/0.05 & 0.034 & 1.5 & 2.5 & 1.3 \\
Sonnet fast & 0.11/0.11 & 0.79/0.84 & $-$0.04/$-$0.03 & 0.224 & 1.4 & 1.4 & 1.4 \\
Opus & 0.02/0.10 & 0.12/0.88 & 0.00/0.34 & 0.037 & 2.8 & 4.3 & 7.8 \\
Opus accelerated & 0.08/0.11 & 0.57/0.88 & 0.01/0.38 & 0.163 & 1.3 & 1.9 & 3.2 \\
Fable & 0.10/0.10 & 0.43/0.76 & 0.00/$-$0.02 & 0.156 & 1.3 & 2.8 & 10.2 \\
\bottomrule
\end{tabular}
\end{table}

\begin{table}[H]
\centering\footnotesize
\caption{Break-even value of a good sample at a 2.1~s deadline, with 95\% bootstrap intervals. At this value, $V^*$, the more expensive configuration's extra delivered value pays for its extra cost, $V^*=\Delta c/\Delta G$. Pairs whose value difference is negative are not listed. The comparison of Haiku thinking with Opus is unresolved ($\Delta G$ $-0.015$, interval $[-0.043,0.012]$).}
\label{tab:cost-breakeven}
\setlength{\tabcolsep}{4pt}
\begin{tabular}{@{}llrr@{}}
\toprule
cheaper & more expensive & $\Delta G$ [95\% CI] & $V^*$ (\$) [95\% CI] \\
\midrule
Haiku fast & Sonnet fast & 0.187 [0.150, 0.224] & 0.027 [0.022, 0.033] \\
Haiku fast & Opus accelerated & 0.120 [0.084, 0.157] & 0.315 [0.239, 0.454] \\
Haiku fast & Fable & 0.080 [0.044, 0.118] & 0.643 [0.435, 1.185] \\
Haiku thinking & Sonnet fast & 0.222 [0.187, 0.256] & 0.022 [0.019, 0.026] \\
Haiku thinking & Opus accelerated & 0.155 [0.123, 0.187] & 0.244 [0.200, 0.308] \\
Haiku thinking & Fable & 0.115 [0.083, 0.147] & 0.448 [0.344, 0.635] \\
Opus & Opus accelerated & 0.170 [0.139, 0.200] & 0.116 [0.097, 0.142] \\
Opus & Fable & 0.130 [0.102, 0.159] & 0.258 [0.204, 0.335] \\
\bottomrule
\end{tabular}
\end{table}

\paragraph{Combinations.} Concurrent pairs request two configurations together. They execute the answer from the configuration designated as stronger if that answer is timely and valid. Otherwise, they execute the other answer if it is timely and valid, or reject if neither qualifies. A conditional call requests the stronger configuration after the first answer arrives, and only when a run-time signal calls for it, such as disagreement with the latest-reading rule. The second answer must still arrive by the original deadline, so its latency adds to that of the first call. If it is late, the first answer executes if it was timely and valid, and otherwise the sample is rejected. We evaluate both strategies by replaying separately issued calls and charge for every call. Table~\ref{tab:cost-pareto} lists the options on the
frontier of value against cost at 2.1~s. We compared median latencies between live pairs and single calls descriptively. They were 0.93 against 0.93~s for Haiku fast, 4.72 against 4.77~s for Opus, 1.35 against 1.37~s for Sonnet and 3.34 against 4.23~s for Fable. These comparisons do not establish that latency was equivalent. Answers did not always repeat. Of 240 Haiku fast answers, 23 differed between the live pair and the single calls. As a result, the live Haiku and Opus pair delivered 0.085 at 2.1~s against 0.141 in the replay. The Sonnet and Fable pair delivered 0.285 against 0.283.

\begin{table}[H]
\centering\footnotesize
\caption{Options on the frontier of value against cost at 2.1~s (all six single configurations, 15 concurrent pairs and 45 conditional-call
rules evaluated). Value is the delivered posterior expected gain per decision over rejection.}
\label{tab:cost-pareto}
\setlength{\tabcolsep}{4pt}
\begin{tabular}{@{}>{\raggedright\arraybackslash}p{0.5\textwidth}rrr@{}}
\toprule
option & \hd{cost per\\decision (\$)} & value & \hd{added value\\per added \$} \\
\midrule
reject without a model & 0 & 0 & none \\
Haiku fast & 0.00026 & 0.097 & 371 \\
Haiku thinking only when Haiku fast disagrees with the latest-reading rule & 0.00041 & 0.108 & 75 \\
concurrent pair, Haiku thinking, else Haiku fast & 0.00063 & 0.151 & 188 \\
Sonnet fast & 0.00523 & 0.284 & 29 \\
concurrent pair, Sonnet fast, else Haiku fast & 0.00549 & 0.289 & 20 \\
concurrent pair, Opus accelerated, else Sonnet fast & 0.04336 & 0.313 & 0.6 \\
\bottomrule
\end{tabular}
\end{table}

\paragraph{Separate and joint summaries.} At 2.1~s, the overprediction $B$ of separate summaries was $+0.074$ $[0.057,0.089]$ for Haiku thinking, $+0.001$ for Haiku fast and $-0.001$ for Sonnet fast. It was $+0.004$ $[-0.009,0.018]$ for Opus, $+0.022$ $[0.006,0.038]$ for Opus accelerated and $-0.026$ $[-0.043,-0.009]$ for Fable. The prespecified primary contrast was Haiku thinking's $B$ minus fast mode's, averaged over groups defined by task family and observation stage. This contrast was $+0.0733$ $[0.0560,0.0886]$. For Haiku thinking, computing $B$ within difficulty levels gave $+0.038$. Thus, differences between difficulty levels account for about half of the $+0.074$ averaged over task family and stage. Haiku thinking was slowest on conflicting decisions, which carry the most value. For Fable, the estimate of $-0.026$ averages over groups defined by task family and observation stage. Adding difficulty level to these groups changed the estimate to $+0.0006$ $[-0.011,0.009]$. This decomposition
does not identify a cause. When choosing which configuration to purchase, separate summaries and paired values selected different configurations in 14 of 252 combinations evaluated on the same data used to make the choice. These combinations covered seven deadlines, six sample values from \$0.01 to \$1{,}000 and six lab sizes. The mean loss across combinations was \$1.60 per hour, and the largest was \$160. We also used two-fold cross-fitting, choosing on one half of the data and evaluating on the other. This gave nonzero differences on held-out data in 4 of 42 combinations of deadline and sample value. The intervals for these differences all included zero.

\paragraph{Deadline-aware prompts.} We compared prompts that stated the deadline with the 60~s prompt replayed at the same deadline on the same 240 decisions. For Haiku thinking, stating the deadline changed delivered value by $+0.012$ $[-0.023,0.048]$ at 2.1~s and $+0.008$ $[-0.032,0.048]$ at 10~s. For Opus, the changes were $+0.019$ $[-0.004,0.043]$ at 2.1~s and $+0.005$ $[-0.014,0.025]$ at 10~s. Actions agreed with the 60~s-prompt answers on 85, 83, 97 and 97\% of decisions, respectively.

\paragraph{Hosted load.} Across three blocks per condition using unselected decisions, Haiku fast's median latency from scheduled arrival was 1.02~s at 0.5 requests/s and 0.97~s at 5 requests/s. Haiku thinking's medians were 1.53~s and 1.46~s, respectively. No request was dropped, and all 1{,}924 offered requests completed with valid answers. In one 5 requests/s block, Haiku fast's 90th-percentile latency was 4.6~s, against 1.4 to 1.9~s in the other blocks. The study does not establish provider capacity. For our provider account, median latency did not rise between the two request rates.

\paragraph{Limits.} The study covers one provider and one collection window. Its designed panel is weighted toward conflicting and hard decisions, with population-weighted values in Table~\ref{tab:cost-levels}. Costs use list prices rather than invoices. We replay answers from 60~s prompts at shorter deadlines and measure latencies under the stated concurrent collection procedure. The dollar scale $V$ is an assumption under which accepting a good sample gains $V$ and accepting a bad sample loses $V$.

\section{Study history and checks}
\label{app:partC}

This part records study planning, costs, deviations and analysis checks.

\subsection{Study sequence}
\label{app:sequence}
We ran the studies in sequence, and earlier results shaped later questions, settings and test families. Each study was planned before its own data collection. Error control within a study does not cover choices made across studies.

After the evidence-onset results, we fixed the concurrent-selection analysis plan and its analysis script before the first dispatch-clock call, so no outcome from that evaluation existed. We tested the script only on synthetic data. After the dispatch-clock result, we prespecified the concurrent-dispatch latency check and then the smaller live cohort, which tested whether live actions and utilities matched the rule. We prespecified the larger live cohort before the smaller one finished and before its analysis existed. The evaluation software kept the smaller cohort's outcomes hidden until it finished. The larger cohort added a primary utility contrast on 600 new instances at the two short deadlines, chosen from earlier evidence. We do not pool the two cohorts.

We then added the Sonnet follow-up. A later exploratory analysis of existing dispatch-clock records motivated the prospective replay. We chose its question and test family after that finding, so they were not original endpoints. After those results, we designed the prespecified study of configuration under load, with its own test family on a fixed local panel. Each addition had its own study plan, fixed before its data collection. We later added two more studies, each also planned before its collection. These were the hosted value and cost study (Appendix~\ref{app:cost}) and the replay repeat on 300 new instances (Appendix~\ref{app:followup}).

\begin{table}[H]
\centering
\caption{Experimental coverage. The reported numbers of scored calls or decisions exclude pilots and preliminary checks where specified. Local vLLM studies use
1 GPU. Hosted inference before the prospective replay cost \$18.62 for Haiku and \$28.71 for Sonnet. The replay added \$0.722.}
\label{tab:coverage}
\footnotesize
\setlength{\tabcolsep}{3pt}
\renewcommand{\arraystretch}{1.04}
\begin{tabular}{@{}>{\raggedright\arraybackslash}p{0.32\textwidth}>{\raggedright\arraybackslash}p{0.28\textwidth}>{\raggedright\arraybackslash}p{0.32\textwidth}@{}}
\toprule
comparison & model and inference setup & study size and cost \\
\midrule
terminal and fault decisions & Haiku 5.5, hosted & 3{,}168 scored calls, \$1.73 \\
medium and high effort & Haiku 5.5, hosted & 2{,}400 scored calls \\
evidence-onset evaluation & Haiku 5.5, simulator & 3{,}600 executed decisions, \$1.08, no failures \\
dispatch-clock evaluation & Haiku 5.5, simulator & 6{,}000 executed decisions \\
prospective replay & Haiku 5.5, recorded answers & 1{,}200 new instances, 2{,}400 scored calls \\
concurrent versus separate dispatch & Haiku 5.5, hosted & 1{,}200 scored calls \\
two live cohorts & Haiku 5.5, simulator & 324; 1{,}200 live decisions \\
Sonnet regret; deadline interaction & Sonnet 5.5, hosted; simulator & 2{,}400; 2{,}400 scored calls \\
initial effort; prior sensitivity & Haiku 5.5, hosted & 6{,}264 scored calls \\
fault challenge & Haiku 5.5, hosted & 120 cases, 1{,}154 calls \\
local stopping evaluation & Qwen3-8B, vLLM & 368 calls \\
synchronized-burst experiment & Qwen3-8B, vLLM & 12 paired allocations \\
serial/shared deadline panel & Qwen3-8B, vLLM & 24 allocations \\
output-constraint panel & Qwen3-8B & 880 responses \\
capacity scan & Qwen3-8B, vLLM & $K\in\{8,\dots,56\}$, 2 blocks \\
configuration under load & Qwen3-8B, vLLM & 12 allocations, 36 fixed cases \\
hosted value and cost study & six hosted configurations & 480 decisions, 6{,}724 main-run and 408 pilot calls, \$84 \\
replay repeat & Haiku 5.5, recorded answers & 300 new instances, 600 scored calls, \$0.19 \\
\bottomrule
\end{tabular}
\end{table}

\subsection{Inference cost}
\label{app:accounting}
We estimated inference costs from recorded token usage, at the prespecified rates of \$0.10 per million input tokens and \$0.50 per million output tokens for Haiku and at list prices for the other configurations. Interrupted calls were counted at their maximum possible cost. We did not check these amounts against invoices, and they cover inference only. All 4{,}884 Sonnet calls succeeded (\$28.71). The shared-serving experiment used 1.54 GPU-hours.

\subsection{Deviations and corrections}
\label{app:deviations}
\paragraph{Shared-serving retry.} The study plan allowed one technical retry, but allocation pair~0 needed two. Both failures happened before any scored burst. The first was a missing import in the warm-up code, and the second a job-launch timing fault. We allowed the second retry before any outcome existed. That retry and pairs 1 to 11 completed, and the analysis uses only the successful attempt for pair~0.

\paragraph{Technical failures.}
\label{app:incidents}
Automatic checks stopped execution seven times, including the two failures above, each time before any scored output. All seven were software or job-scheduling faults in code paths that had not been tested end to end.

\paragraph{Replay study-plan amendments.} Both amendments were made before the study's first provider call, when no replay outcome existed. The first raised the study's runtime limit and changed how the technical tests simulated late responses. The second specified how results would be reported. It required estimates for both blocks and their contrast, pooled and within-group estimates, an approximate label for the bootstrap $p$, a rule for ties, and a comparison of the lower confidence bound with the planning target. The estimator, instances, predictors, schedule and requests did not change.

\paragraph{Numerical corrections.} The primary analyses for concurrent selection and the larger live cohort compared bootstrap contrasts with zero in floating-point arithmetic. Rounding could therefore exclude contrasts equal to zero, which the prespecified test counts in its tails. Recounting with integer outcome sums on the same draws gives $p=.017$ and $p=.073$, the values reported here. Estimates, intervals and decisions did not change.

\subsection{Computation checks}
\label{app:verification}
For each study, we froze the instances, prompts, request formats, schedule, failure handling and analysis before collecting scored data. After collection, separate implementations recomputed the analyses in Table~\ref{tab:checks}. These checks confirm the computations and agreement with the fixed simulator. They do not validate the simulator's physical model. We retain the code, configurations, raw records and analysis scripts for the studies in this paper.

\begin{table}[H]
\centering
\caption{Computation checks by a separate implementation. The last column gives the largest numerical difference. ``Post hoc'' marks descriptive analyses.}
\label{tab:checks}
\footnotesize
\setlength{\tabcolsep}{3pt}
\begin{tabular}{@{}>{\raggedright\arraybackslash}p{0.3\textwidth}>{\raggedright\arraybackslash}p{0.5\textwidth}>{\raggedright\arraybackslash}p{0.16\textwidth}@{}}
\toprule
study & what was reproduced & difference \\
\midrule
terminal and fault comparisons & estimates, intervals, $p$ and all 3{,}236 records & $1.3\times10^{-15}$ \\
medium and high effort & estimates and intervals, on the same instances & 0 \\
evidence-onset evaluation & all 3{,}756 records against the simulator & 0 \\
dispatch-clock evaluation & interval, $p$ and all 6{,}000 scored records & $<3\times10^{-17}$ \\
concurrent selection & rule, selected records, interval and $p$ & 0 \\
latency check & intervals, schedule, 1{,}256 requests and cost & 0 \\
smaller live cohort & all 324 decisions re-executed; schedule, deadlines and cost & $<3\times10^{-20}$ \\
larger live cohort & all 1{,}200 decisions re-executed; estimate, interval and $p$ & $<3\times10^{-20}$ \\
Sonnet regret comparison & all 2{,}428 score records and the bootstrap & 0 \\
Sonnet deadline comparison & all executed actions and the bootstrap & 0 \\
shared-serving experiment & estimate, exact $p$ (10/4{,}096) and 1{,}344 records & $2.9\times10^{-13}$ \\
regret by returned action (post hoc) & all categories, from raw records & exact at display precision \\
evidence-onset answer regret (post hoc) & reference values and all 3{,}600 requests & $<3\times10^{-20}$ \\
fixed-answer sweep (post hoc) & all 6{,}000 replayed records re-executed & 0 \\
\bottomrule
\end{tabular}
\end{table}

A separately written implementation also reproduced the replay's $+0.0241$ overestimate from the raw records, with a 95\% CI of $[0.0163,0.0315]$ from independent bootstrap draws. This check reuses the same records and adds no confirmatory test. Separate implementations recomputed the earlier analyses at the time, and a third implementation in exact arithmetic matched the fault challenge's scoring. Tests before scoring found three software defects, which we fixed and retested on preliminary cases before scoring.

\end{document}